%% file: main.tex
\documentclass[letterpaper]{article}
\usepackage[preprint]{kosmo}
\usepackage{multirow}
\usepackage{float}
\usepackage{xurl}
\usepackage{tabularx}
\usepackage{needspace}
\usepackage{tocloft}

\hypersetup{linktoc=all}

\pretocmd{\section}{\FloatBarrier\Needspace{100pt}}{}{}
\pretocmd{\subsection}{\FloatBarrier\Needspace{72pt}}{}{}
\pretocmd{\subsubsection}{\Needspace{60pt}}{}{}

\makeatletter
\patchcmd{\kosmo@maketitle}{after skip=0.48cm}{after skip=0.08cm}{}{\PackageError{layout}{Abstract spacing patch failed}{}}
\renewcommand{\maketitle}{%
  \par\begingroup
  \kosmo@maketitle
  \@thanks
  \endgroup
  \setcounter{footnote}{0}
  \gdef\@thanks{}\gdef\@author{}\gdef\@title{}
}
\makeatother

\newcommand{\norm}[1]{\left\lVert#1\right\rVert}
\title{Kosmo-GFM: A Production-Ready\\Geometry Foundation Model}
\author{Kosmo Research}
\affiliations{}
\correspondingauthor{}
\projectpage{\url{https://kosmoresearch.github.io/Kosmo-GFM/}}
\hypersetup{pdftitle={Kosmo-GFM: A Production-Ready Geometry Foundation Model},
  pdfauthor={Kosmo Research},pdfsubject={Technical report, single-column edition},
  pdfkeywords={geometry foundation model, reconstruction, PoseEngine, KosmoWorld}}
\kosmoabstract{
We present Kosmo-GFM, a scene reconstruction system that integrates a geometry foundation model
(GFM), a pose optimization engine (PoseEngine), and reconstruction-derived supervision from
KosmoWorld. GFM predicts camera parameters, depth, cross-view correspondences, and dynamic 3D points
from images and videos. PoseEngine combines these predictions with geometric optimization to recover
scene-level camera poses and geometry for downstream 3D and 4D Gaussian Splatting. KosmoWorld
generates images and geometric supervision from reconstructed real-world scenes to improve GFM's
geometric predictions. On the Sparse and Medium subsets of SpatialBench, GFM achieves the lowest
absolute relative depth error and the highest relative-pose AUC at 5$^\circ$ among the compared
methods. It also achieves the lowest mean endpoint errors for dynamic points and scene flow across
the five evaluated datasets. Across 34 scenes completed by all compared reconstruction systems,
PoseEngine achieves the highest scene-averaged pose AUC at thresholds of 1$^\circ$, 3$^\circ$, and
5$^\circ$, with a cumulative runtime of 14.07 hours, approximately 20\% of GLUEMAP's 69.67 hours.
PoseEngine Flash provides a further accuracy--efficiency trade-off, with a measured mixed-precision
runtime of 2.62 hours, approximately 4\% of GLUEMAP's runtime. Separate scaling experiments show
generally favorable trends with increased data, model computation, and training budgets. These
results support the use of learned geometric priors within a scene-level reconstruction pipeline.
}

\AtBeginEnvironment{thebibliography}{\setlength{\baselineskip}{11.5pt}\interlinepenalty=10000}

\makeatletter
\patchcmd{\paragraph}{3.25ex \@plus1ex \@minus.2ex}{1.75ex \@plus.5ex \@minus.2ex}{}{\PackageError{layout}{Paragraph spacing patch failed}{}}
\makeatother
\begin{document}
\onecolumn
\raggedbottom
\setlength{\parindent}{0pt}
\maketitle

\begin{figure}[H]
  \centering
  \includegraphics[width=.84\textwidth]{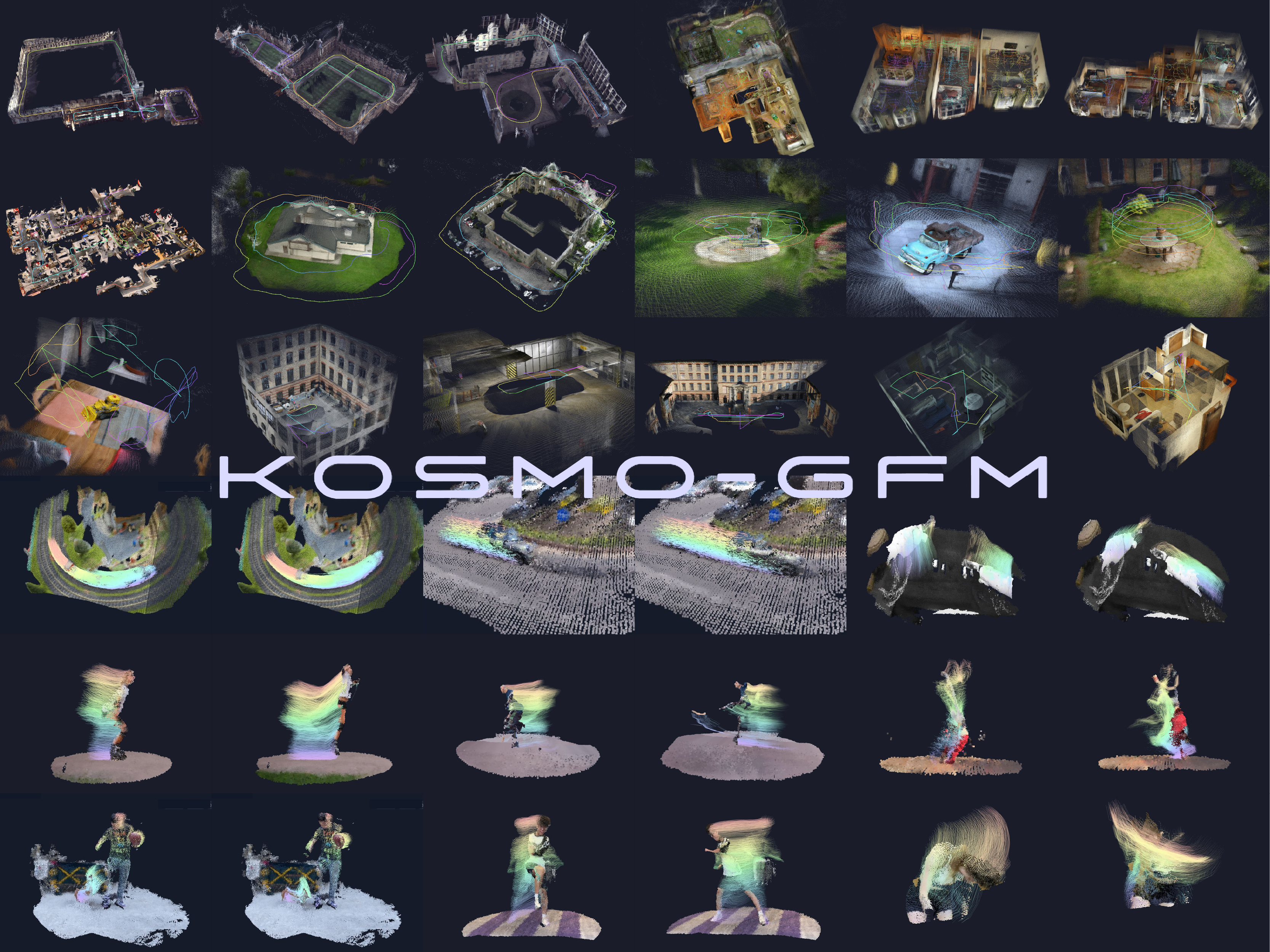}
  \caption{Kosmo-GFM. Top: static scene reconstructions with camera trajectories. Bottom: dynamic 3D points and motion trajectories.}
  \label{fig:web-concept}
\end{figure}

\clearpage
{\hypersetup{linkcolor=black}\tableofcontents}
\clearpage
\section{Introduction}
\label{sec:introduction}
Recovering real-world geometry and motion from images and videos is a central
problem in computer vision. Traditional reconstruction systems recover scenes
through feature matching, camera pose estimation, and geometric optimization.
While these methods benefit from explicit geometric constraints, they depend
on reliable visual correspondences and sufficient overlap between views.
Textureless regions, repetitive structures, wide baselines, and scene dynamics
introduce uncertainty that can compromise both camera estimation and geometry
recovery. As the number of input views increases, maintaining scene consistency can also incur
substantial computational cost. Balancing reconstruction quality, efficiency,
and robustness to complex scenes therefore remains an important challenge.

Geometry foundation models offer a new approach to these challenges. Through
large-scale pretraining, they learn transferable spatial priors and directly
predict cameras, depth, cross-view correspondences, and dynamic geometry from
visual observations. These predictions provide information where conventional
geometric constraints are insufficient. However, model predictions do not
inherently satisfy cross-view and temporal consistency across an entire scene,
particularly for long sequences, large environments, and dynamic observations.
Spatial priors and geometric optimization are therefore complementary: the
former provide useful predictions and initialization, while the latter
integrate them into coherent scene structure. The objective is not to
eliminate optimization altogether, but to exploit model capabilities to reduce
the optimization required for reliable reconstruction.

Following this principle, we build Kosmo-GFM, a scene reconstruction system
that combines a geometry foundation model, GFM, with a pose optimization
engine, PoseEngine. GFM estimates camera parameters, depth, cross-view
correspondences, and dynamic 3D points from images and videos, providing
spatial priors for both static geometry recovery and dynamic scene
understanding. PoseEngine leverages these outputs to efficiently recover
scene-level camera poses and geometry, supporting the subsequent construction
of 3D Gaussian Splatting (3DGS)~\citep{kerbl2023gaussians} representations for static scenes and 4D
Gaussian Splatting (4DGS)~\citep{wu2024fourdgs} representations for dynamic scenes. The system thus
connects geometric predictions from visual observations to scene-level
reconstruction and renderable 3D and 4D representations.

Beyond integrating the model and optimization engine, Kosmo-GFM emphasizes the
continued improvement of spatial priors. We use high-quality real-world
reconstruction data from KosmoWorld to train and improve GFM's understanding
of scene geometry. As GFM improves, more reliable predictions can reduce
uncertainty in subsequent geometry recovery and lessen the reliance on
explicit optimization, providing a basis for simpler pose optimization and
higher system efficiency. We therefore evaluate not only the accuracy of GFM's
predictions, but also their practical role in the complete reconstruction
system.

We evaluate GFM on static and dynamic geometry, and evaluate PoseEngine by pose accuracy, runtime, and the quality of the resulting 3DGS reconstructions. We also conduct a separate scaling study using models developed specifically for those experiments. Our main contributions and findings are:
\begin{itemize}
  \setlength{\itemsep}{1pt}\setlength{\parskip}{0pt}
  \item \textbf{GFM, a geometry foundation model with task-specific decoders.}
  From images and videos, GFM predicts cameras, depth, cross-view matches,
  and dynamic 3D points, and can optionally take camera parameters as input.
  It achieves strong relative-pose accuracy on both SpatialBench subsets
  and reduces average endpoint error for dynamic points and scene flow.
  \item \textbf{KosmoWorld, real-world reconstruction data for geometric
  priors.} Adding KosmoWorld improves depth and relative-pose estimation
  on both SpatialBench subsets, demonstrating the value of real-world supervision.
  \item \textbf{PoseEngine, a prior-driven reconstruction engine.}
  Across scenes completed by all compared methods, PoseEngine improves
  mean pose accuracy over GLUEMAP with substantially lower runtime.
  PoseEngine Flash offers a further accuracy--efficiency trade-off.
  \item \textbf{Scaling behavior of geometric reconstruction.} Scaling experiments vary data
  volume, model scale, and training compute. Depth and camera accuracy
  generally improve with more training data, larger models, and increased training compute.
\end{itemize}

\section{Related Work}
\label{sec:related}
\input{sections/related-work}

\section{Kosmo-GFM}
\label{sec:method}
\begin{figure}[!htbp]
\centering
\includegraphics[width=\textwidth]{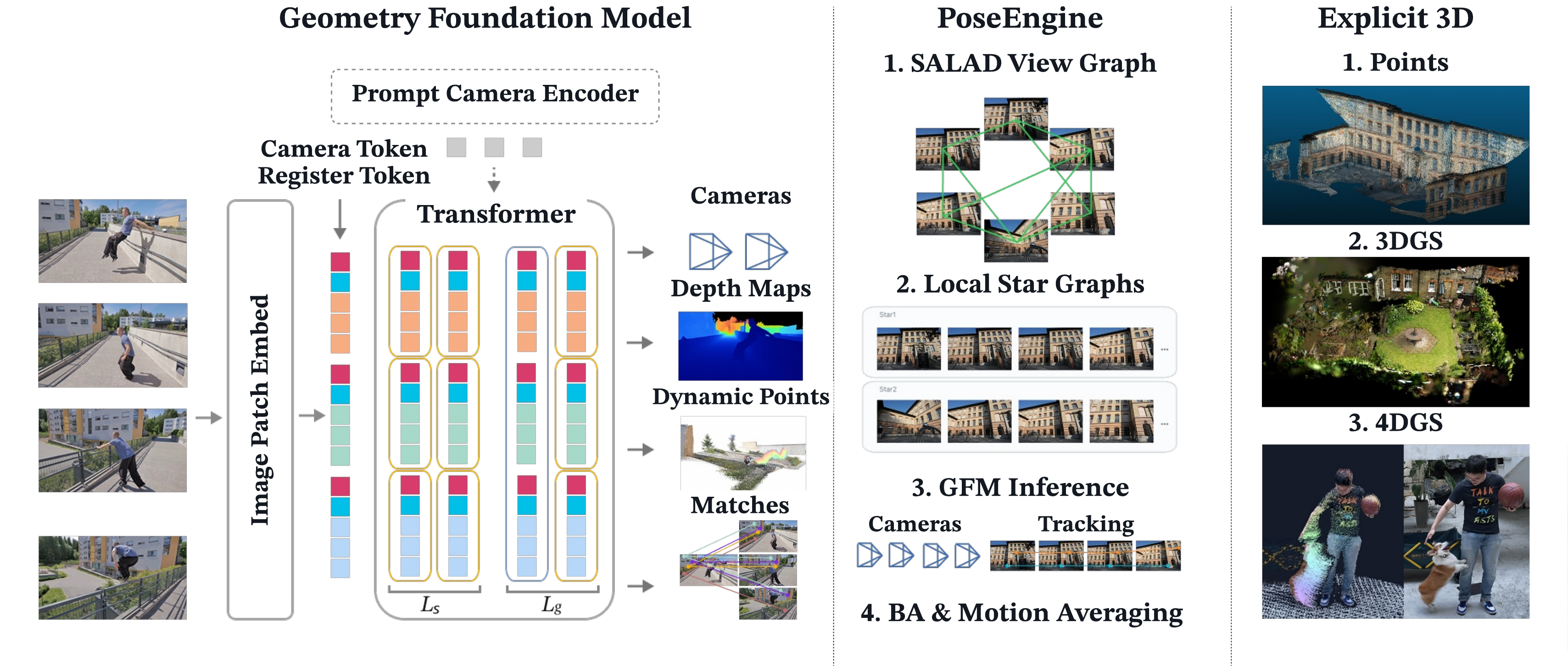}
\caption{Overview of Kosmo-GFM. GFM supplies geometric priors, and PoseEngine organizes cross-view constraints and refines scene cameras. The resulting cameras and geometry support point-cloud, 3DGS, and 4DGS reconstruction.}
\label{fig:web-pipeline}
\end{figure}

This section introduces the three complementary components of Kosmo-GFM and how they work together (Figure~\ref{fig:web-pipeline}). GFM uses shared multi-view features to predict cameras, depth, cross-view correspondences, and dynamic points, with optional camera conditioning. PoseEngine combines these priors with cross-view constraints to recover scene cameras for dense geometry prediction and subsequent rendering. KosmoWorld converts real-world reconstructions into training supervision; the reported ablation shows improvements in depth and relative-pose prediction. Together, these components connect multi-task geometric prediction, scene-level camera recovery, and training-data construction, supporting a range of static and dynamic reconstruction outputs.

\subsection{GFM}
Given images $\mathcal I=\{I_i\}_{i=1}^{V}$, GFM learns a shared multi-view representation from which task-specific decoders recover geometry and motion. The encoder follows the Depth Anything 3~\citep{lin2025da3} design: within-view attention captures image structure, while cross-view attention establishes geometric context across observations. Camera tokens and four register tokens complement the image tokens. The backbone is ViT-Giant with $14\times14$ patches. Features from blocks 19, 27, 33, and 39 are combined into 3,072-dimensional representations for depth decoding. Static geometry, motion, and matching are trained under task-specific objectives, as described below. Besides RGB-only inference, GFM can also take known camera parameters as an input prompt (Section~\ref{sec:camera-conditioned-gfm}). Section~\ref{sec:data} compares the two modes.

\subsubsection{Decoder Architecture}
\begin{figure}[!htbp]
\centering
\includegraphics[width=\textwidth]{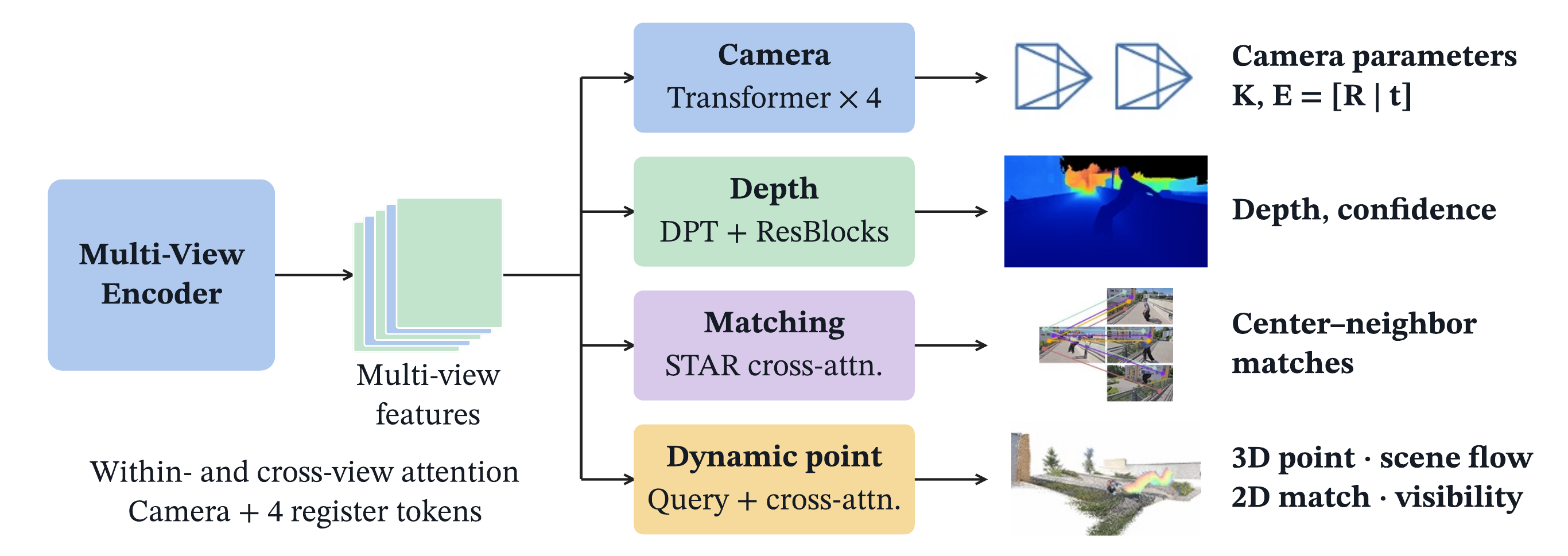}
\caption{Task-specific decoding from shared multi-view features. Camera and depth decoding recover static geometry, STAR matching establishes cross-view correspondences, and source--target queries predict geometry and motion at a selected time.}
\label{fig:architecture}
\end{figure}

The decoders address complementary aspects of reconstruction (Figure~\ref{fig:architecture}). Camera and depth prediction establish a common geometric frame. Cross-view matching links observations of the same scene point, and spatiotemporal queries extend this representation to moving points.

\paragraph{\textbf{Camera decoder.}}
A four-block transformer decodes each camera token into translation, quaternion rotation, and horizontal and vertical fields of view. The nine-dimensional estimate is refined over four iterations, with the previous estimate detached before conditioning the next update. Supervision at each iteration guides the progressive refinement of the intrinsic matrix $K_i$ and the extrinsics $E_i=[R_i\mid t_i]$. GFM uses the camera coordinate system of the first frame as the world coordinate system and predicts extrinsics in the world-to-camera convention: $E_i$ maps a world point $P$ to the camera frame of view $i$ as $R_iP+t_i$.

\paragraph{\textbf{Depth decoder.}}
The depth decoder follows a DPT-style coarse-to-fine design~\citep{ranftl2021dpt}. Four feature levels are projected to 256 channels, reshaped into spatial maps, and fused through successive upsampling and skip connections. After each upsampling stage of the DPT head, we add residual (ResNet) blocks~\citep{he2016resnet}, which reduce spurious 3D points near depth discontinuities. The final prediction contains depth and confidence, with an exponential activation ensuring positive depth. For a pinhole camera, the pixel $\bar x=(u,v,1)^\top$ is mapped to the common coordinate frame by

\begin{equation}
 \hat P_i^{\mathrm{local}}=\hat z_i K_i^{-1}\bar x,\qquad
 \hat P_i^{\mathrm{world}}=R_i^\top(\hat P_i^{\mathrm{local}}-t_i).
 \label{eq:unproject}
\end{equation}
\paragraph{\textbf{Dynamic-point decoder.}}
Following UniQuery4R~\citep{chen2026uniquery4r}, a query $q=(u,v,s,t)$ identifies a pixel $(u,v)$ in source frame $s$ and asks for its state at target frame $t$. Multi-scale features sampled at the source pixel are combined by gated fusion and attend to target-view features. The resulting representation predicts the target-time 3D point, scene flow, 2D correspondence, and visibility. A source-only path predicts local depth. All points are expressed in the first-view reference frame. Encoding a clip once allows its features to be reused across source pixels and target times, supporting both sparse tracking and dense prediction.

\Needspace{7\baselineskip}
\paragraph{\textbf{Matching decoder.}}
The matching decoder adopts the STAR organization of MVLoMa~\citep{lu2026mvloma}, connecting one center view to several neighboring views. For each view, descriptors are sampled from GFM feature maps at the keypoint locations. Cross-attention between the center and neighboring views then updates these descriptors using cross-view evidence. The resulting descriptors yield matching assignments for each center--neighbor pair, providing correspondences for multi-view track construction.

\subsubsection{Training Loss}
Training combines geometric regression, confidence estimation, visibility classification, and correspondence learning. For each task, angle brackets denote averaging over the valid samples. Task-specific masks exclude invalid geometric labels.

\paragraph{\textbf{Camera loss.}}
Camera supervision constrains relative pose across views and field of view within each image. For world-to-camera transformations $E_i$ and $E_j$, the relative transformation is

\begin{equation}
 T_{i\rightarrow j}=E_jE_i^{-1}.
 \label{eq:relative}
\end{equation}
At refinement iteration $k$, we average translation errors and rotation errors measured using quaternions over view pairs:

\begin{equation}
 \mathcal{L}_{\mathrm{pose}}^{(k)}=
 \left\langle\lambda_T\mathcal{L}_{T,i\rightarrow j}^{(k)}+
 \lambda_R\mathcal{L}_{R,i\rightarrow j}^{(k)}\right\rangle_{i,j}.
 \label{eq:pose}
\end{equation}
The overall camera objective accumulates pose and field-of-view supervision over the refinement iterations:

\begin{equation}
 \mathcal{L}_{\mathrm{camera}}=
 \sum_k\alpha_k\left[\mathcal{L}_{\mathrm{pose}}^{(k)}+
 \lambda_{\mathrm{FoV}}\left\langle\mathcal{L}_{\mathrm{FoV},j}^{(k)}\right\rangle_j\right].
 \label{eq:camera}
\end{equation}
We use L1 pose errors, with separate weights for translation, rotation, and field of view. Iteration weights discount earlier refinement steps, and dataset-level weights balance supervision across training sources.

\paragraph{\textbf{Depth loss.}}
Depth supervision compares predicted and reference local 3D points, $\hat P$ and $P$. Let $r=\norm{\hat P-P}_1$ and let $z$ be reference depth. Normalizing the geometric residual by depth reduces its dependence on scene scale. With predicted confidence $c$, the depth-normalized objective is

\begin{equation}
 \mathcal{L}_{\mathrm{depth}}=
 \lambda_{\mathrm{geom}}\left\langle\frac{r}{|z|+\epsilon}\right\rangle+
 \lambda_{\mathrm{conf}}\left\langle cr-\beta\log c\right\rangle.
 \label{eq:depth}
\end{equation}
Here $\epsilon$ stabilizes the geometric normalization, $c=1+\exp(a)$, and $\beta$ controls the penalty on trivial confidence reduction. Only the geometric term is divided by reference depth, whereas the confidence term uses the absolute 3D residual $r$. Validity masks and robust residual filtering suppress unreliable supervision.

\paragraph{\textbf{Dynamic-point and visibility loss.}}
Dynamic-point supervision follows the UniQuery4R formulation~\citep[Sec.~3 and Appendix~A]{chen2026uniquery4r}. Training geometry is normalized by the mean distance from the clip's points to the reference origin. For a target-time prediction $\hat P_i$ and reference point $P_i$, let $z_i$ denote the reference depth and $\mathcal V_P$ the valid-label set. Define

\begin{equation}
 e_i=\norm{\hat P_i-P_i}_1,\qquad
 r_i=\frac{e_i}{|z_i|+\epsilon},\qquad c_i=1+\exp(a_i^{P}),\qquad
 \mathcal R_P=\{i\in\mathcal V_P:r_i<\tau_P\}.
 \label{eq:dynamic-residual}
\end{equation}
Here $\tau_P$ is the geometric residual threshold. With per-query weights $w_i$, point supervision combines a filtered geometric residual with confidence-weighted regression:

\begin{equation}
 \mathcal L_{\mathrm{dynamic}}=
 \lambda_{\mathrm{geom}}\langle w_i r_i\rangle_{\mathcal R_P}
 +\lambda_{\mathrm{conf}}\langle w_i(c_i e_i-\beta\log c_i)\rangle_{\mathcal V_P}.
 \label{eq:dynamic}
\end{equation}
The confidence term uses the unnormalized position error $e_i$, whereas residual rejection applies only to the geometric term.

Alongside 3D position regression, the dynamic-point objective supervises visibility for queries visible in the source view. Let $y_i^t$ indicate target-view visibility, $a_i^v$ be its predicted logit, and $\mathcal V_s$ be the source-visible query set. The visibility component is binary cross-entropy:

\begin{equation}
 \mathcal L_{\mathrm{visibility}}=
 -\left\langle y_i^t\log\sigma(a_i^v)+(1-y_i^t)\log(1-\sigma(a_i^v))
 \right\rangle_{\mathcal V_s}.
 \label{eq:visibility}
\end{equation}
Target-occluded points remain negative examples, so the model learns both visibility and occlusion. Empty valid sets contribute zero. This term is weighted separately from the correspondence-related auxiliary visibility terms. Auxiliary 2D correspondence supervision uses a robust residual and retains occluded target tracks with reduced weight. Scene-flow supervision separates direction and magnitude and balances moving and static points.

\paragraph{\textbf{Matching loss.}}
Following MVLoMa~\citep[Sec.~3.3--3.4]{lu2026mvloma}, an $N$-view STAR comprises one center and $N-1$ neighbors. Supervision at each of the $L$ matcher layers combines positive assignment, matchability, and geometric-negative terms:

\begin{equation}
 \mathcal L_{\mathrm{matches}}=\frac{1}{L}\sum_{\ell=1}^{L}
 \left[\mathcal L_{\mathrm{glue}}^\ell+\mathcal L_{\mathrm{matchability}}^\ell
 +\frac{1}{N-1}\sum_{j=1}^{N-1}\mathcal L_{\mathrm{neg}}^{j,\ell}\right].
 \label{eq:matching}
\end{equation}
For center and neighbor descriptors $d_p^c$ and $d_q^j$, dual-softmax normalization produces an assignment confidence for each edge:

\begin{equation}
 S_{pq}^j=\langle d_p^c,d_q^j\rangle,\qquad
 P_{pq}^{j,\ell}=\operatorname{softmax}_p(S^{j,\ell}/\tau)_{pq}
 \operatorname{softmax}_q(S^{j,\ell}/\tau)_{pq}.
 \label{eq:assignment}
\end{equation}
Positive correspondences are obtained by depth-and-pose reprojection, subject to valid depth, image bounds, and target-depth and occlusion consistency. The matchability term predicts whether a valid counterpart exists. Geometric negatives are identified from high-confidence assignments that violate the ground-truth epipolar relation:

\begin{equation}
 F_j=K_j^{-\top}[t_j]_\times R_jK_c^{-1},\qquad
 e_{pq}^j=\frac{|(\bar x_q^j)^\top F_j\bar x_p^c|}
 {\norm{(F_j\bar x_p^c)_{1:2}}_2},\qquad
 \mathcal N_j=\{(p,q)\in\mathcal M_j:e_{pq}^j>\tau_{\mathrm{epi}}\}.
 \label{eq:epipolar-negatives}
\end{equation}
Here $\tau_{\mathrm{epi}}$ is the epipolar-error threshold in pixels, $(R_j,t_j)$ is the relative pose, $K_c,K_j$ are intrinsics, and $\bar x$ denotes homogeneous pixel coordinates. The candidate set $\mathcal M_j$ contains final-layer mutual maxima above threshold $\mu$. Coordinates and intrinsics are expressed at the prediction resolution. The same negative set supervises every layer:

\begin{equation}
 \mathcal L_{\mathrm{neg}}^{j,\ell}=
 -\frac{1}{|\mathcal N_j|}\sum_{(p,q)\in\mathcal N_j}
 \log(1-P_{pq}^{j,\ell}).
 \label{eq:negative}
\end{equation}
Empty negative sets contribute zero. Matching supervision trains the cross-view attention and assignment layers while keeping the GFM feature encoder frozen.

\input{sections/scene-scale}

\subsubsection{View Sampling}
\label{sec:training-data}
\begin{figure}[!htbp]
\centering
\includegraphics[width=\textwidth]{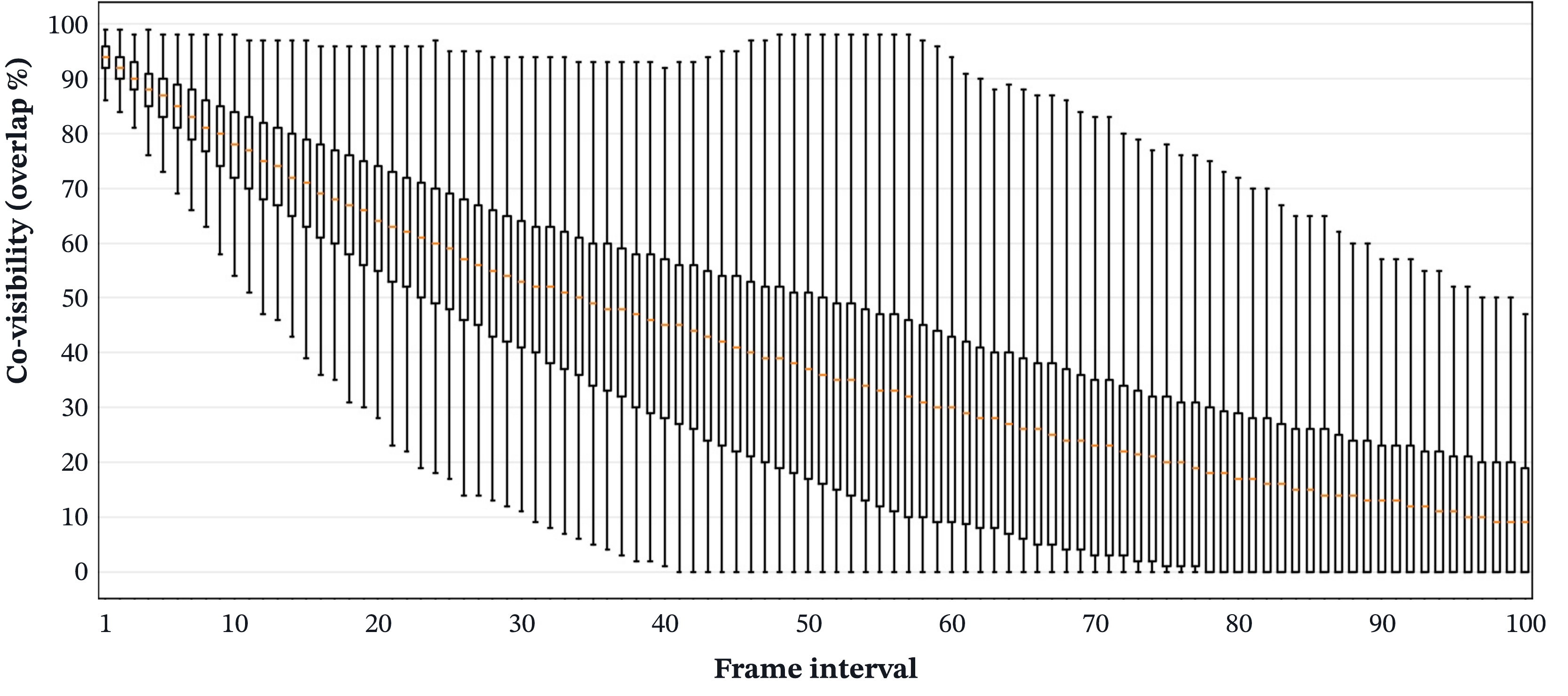}
\caption{Co-visibility versus temporal separation. A sampling step near the lower-bound zero crossing favors low-overlap training clips.}
\label{fig:sampling}
\end{figure}

We sample training clips with deliberately low inter-frame co-visibility, reducing redundant observations of the same scene content. Low-overlap observations encourage the model to learn more precise features for establishing correspondences across viewpoints. To select the temporal interval, we examine how the co-visibility distribution changes with frame separation and choose a step where its lower bound approaches zero. Figure~\ref{fig:sampling} illustrates this sampling criterion on ScanNet~\citep{dai2017scannet}. For $V$ input views, we form a temporal window of $V\times\mathrm{step}$ frames and resample $V$ views within it. The interval is chosen separately for each data source, and it is not a requirement that every sampled view pair have zero overlap.

\subsubsection{Camera-conditioned GFM}
\label{sec:camera-conditioned-gfm}
We obtain Camera-conditioned GFM by further training the RGB-only GFM. It takes images or video frames with known per-frame camera intrinsics and extrinsics and predicts a dense point cloud for each frame. A prompt camera encoder maps each frame's camera parameters to a prompt camera token.

Unlike Depth Anything 3~\citep{lin2025da3}, we replace the first register token with the prompt token instead of replacing the camera token. Directly replacing the camera token lets prompt information flow directly to the camera head. This can substantially improve camera accuracy, but risks an information shortcut: the camera head can reproduce the supplied parameters while patch tokens make limited use of them. Our design encourages both camera and patch tokens to integrate the prompt information for dense geometry prediction.

We also construct training samples with no inter-frame co-visibility, encouraging both token types to use camera prompts without visual correspondences. Table~\ref{tab:prompt} in Section~\ref{sec:data} reports the results.

\vspace{-8pt}
\subsubsection{AnyCamera}
\label{sec:anycamera-method}
\begin{figure}[!htbp]
\centering
\includegraphics[width=\textwidth,height=\dimexpr\textheight-165pt\relax,keepaspectratio]{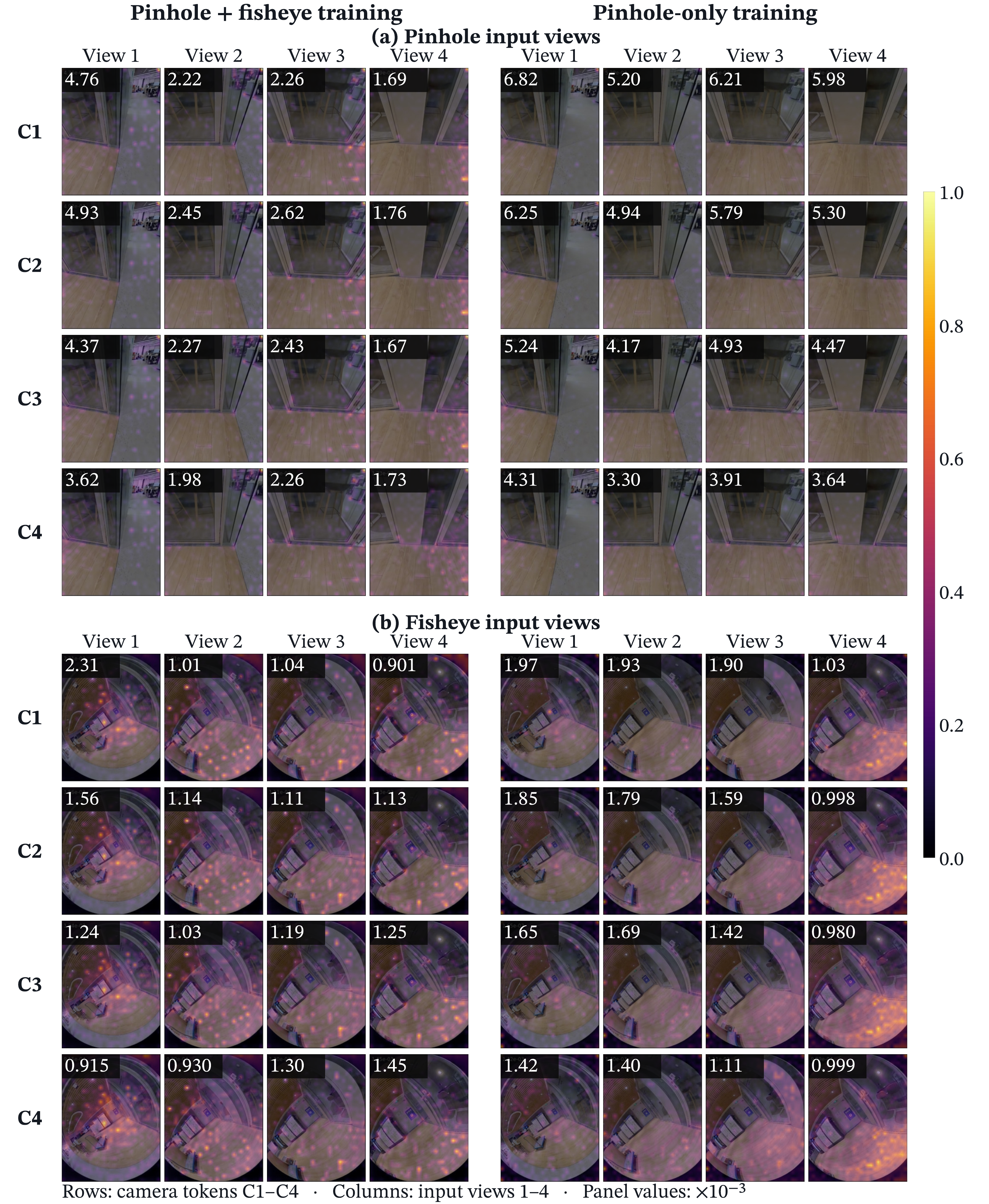}
\caption{Camera-token attention across views under mixed-camera and pinhole-only training. Each panel overlays the attention of every camera token (rows) on the input views (columns). Top: four pinhole input views. Bottom: four fisheye input views. Both models use the same image sequence in each example.}
\label{fig:anycamera-attention}
\end{figure}

\begin{figure}[!htbp]
\centering
\includegraphics[width=\textwidth,height=.80\textheight,keepaspectratio]{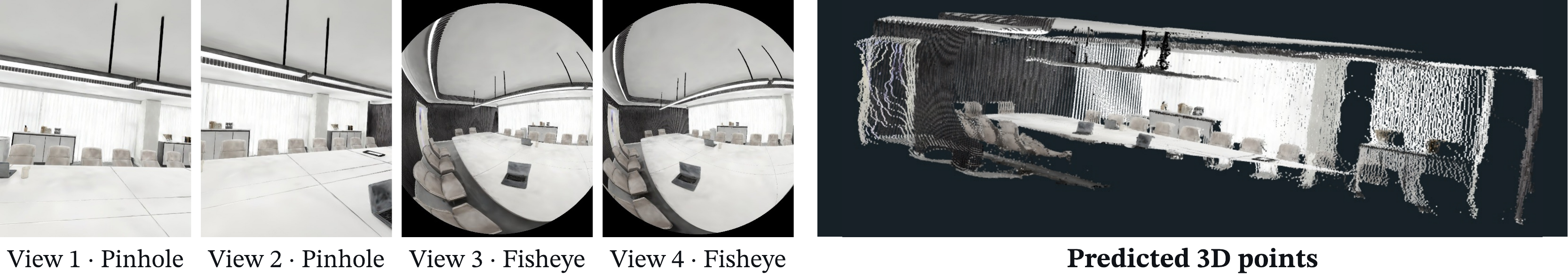}
\caption{Mixed pinhole--fisheye reconstruction. Left: two pinhole and two fisheye input views. Right: 3D points predicted by GFM from the four views.}
\label{fig:anycamera-fisheye}
\end{figure}

\begin{figure}[!htbp]
\centering
\includegraphics[width=\textwidth,height=.80\textheight,keepaspectratio]{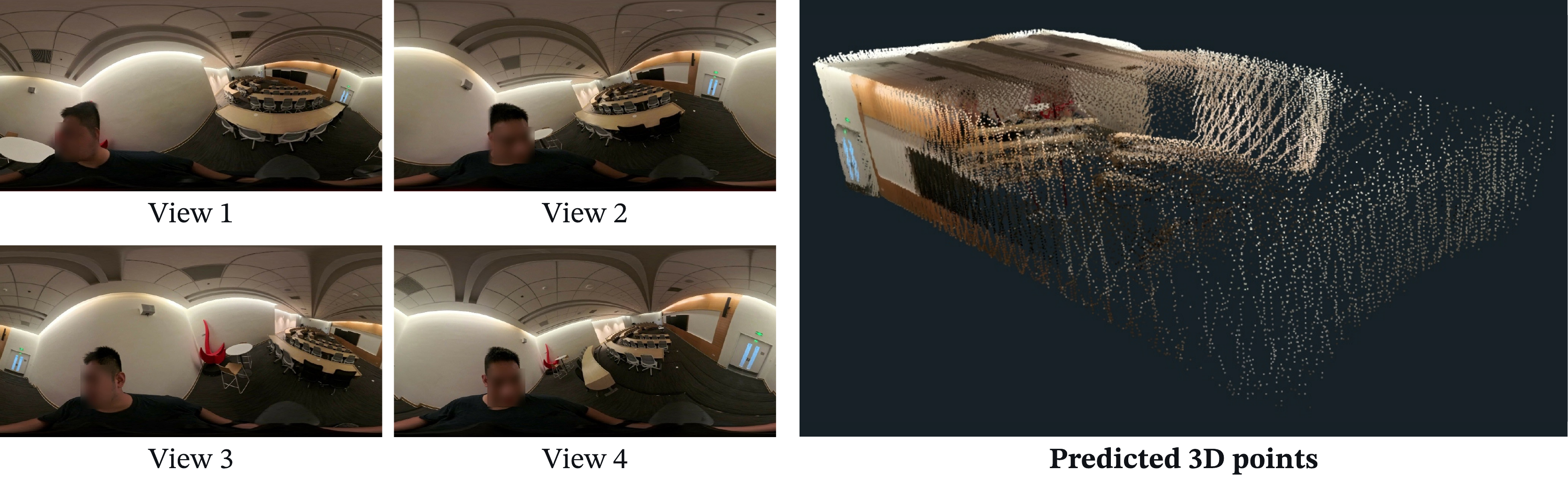}
\caption{Panoramic reconstruction. Left: four equirectangular input views. Right: 3D points predicted by GFM from the four panoramas.}
\label{fig:anycamera-panorama}
\end{figure}

To extend GFM beyond pinhole imagery, we incorporate fisheye and panoramic observations during training. Compared with conventional pinhole cameras, fisheye and panoramic cameras offer a wider field of view. This broader coverage can provide more multi-view constraints in latent space, enabling more accurate camera poses and more efficient reconstruction. Figure~\ref{fig:anycamera-attention} compares camera-token attention under pinhole-only and pinhole-plus-fisheye training. It provides a qualitative view of the change in cross-view feature aggregation. Figure~\ref{fig:anycamera-fisheye} shows reconstruction with mixed pinhole and fisheye cameras, and Figure~\ref{fig:anycamera-panorama} shows reconstruction with panoramic cameras.
\vspace{-6pt}

\subsection{PoseEngine}
\label{sec:poseengine}
Feed-forward predictions provide a strong geometric initialization, but scene reconstruction also requires consistency across many observations. Inspired by GLUEMAP~\citep{pan2026gluemap}, PoseEngine combines GFM priors with optimization and organizes reconstruction around STARs, each consisting of one center image and its neighbors. It provides two STAR construction schemes for different input types.

\paragraph{\textbf{STAR construction.}} For unordered image collections, PoseEngine follows the STAR construction scheme of GLUEMAP~\citep{pan2026gluemap}: it retrieves candidate neighbors with SALAD~\citep{izquierdo2024salad} global descriptors and verifies them with Doppelgangers++~\citep{xiangli2025doppelgangers}, forming STARs of 26 images. For ordered sequences, each STAR is built mainly from temporally adjacent frames, supplemented by a small number of SALAD-retrieved frames, again totaling 26 images. We compare both STAR policies in the matching study, while the main reconstruction benchmark uses the ordered policy, labeled \textbf{PoseEngine}.

\paragraph{\textbf{Matching.}} Within each STAR, PoseEngine matches the center image to every neighbor in a single pass using the MVLoMa decoder of GFM. For STAR inputs, a single GFM forward pass therefore provides both geometric priors and correspondences without requiring additional images or new image pairings. The evaluated implementation uses a LoMa Large fallback for inputs of at most 26 images.

\paragraph{\textbf{Design principle.}} The stronger GFM becomes, the simpler PoseEngine can be: more accurate camera and geometry priors require fewer bundle-adjustment rounds, which substantially reduces system runtime. Section~\ref{sec:system-results} supports this view, as PoseEngine achieves higher mean pose accuracy than GLUEMAP with substantially lower cumulative runtime.

\paragraph{\textbf{PoseEngine Flash.}} For large-scale scenes, we further provide PoseEngine Flash, which combines FP8/FP16 mixed-precision inference, a simple bundle-adjustment scheme, and system-level optimization of the overall pipeline to substantially reduce runtime and enable fast reconstruction of large scenes.

The refined cameras condition dense geometry prediction and initialize 3DGS and 4DGS reconstruction. During 3DGS and 4DGS fitting, joint optimization of camera poses and Gaussian parameters can compensate for part of the pose error introduced by faster reconstruction. This enables PoseEngine Flash to support high-quality 3DGS and 4DGS reconstruction while reducing pose-estimation cost.

\subsection{KosmoWorld}
\label{sec:kosmoworld}
\begin{figure}[!htbp]
\centering
\includegraphics[width=\textwidth]{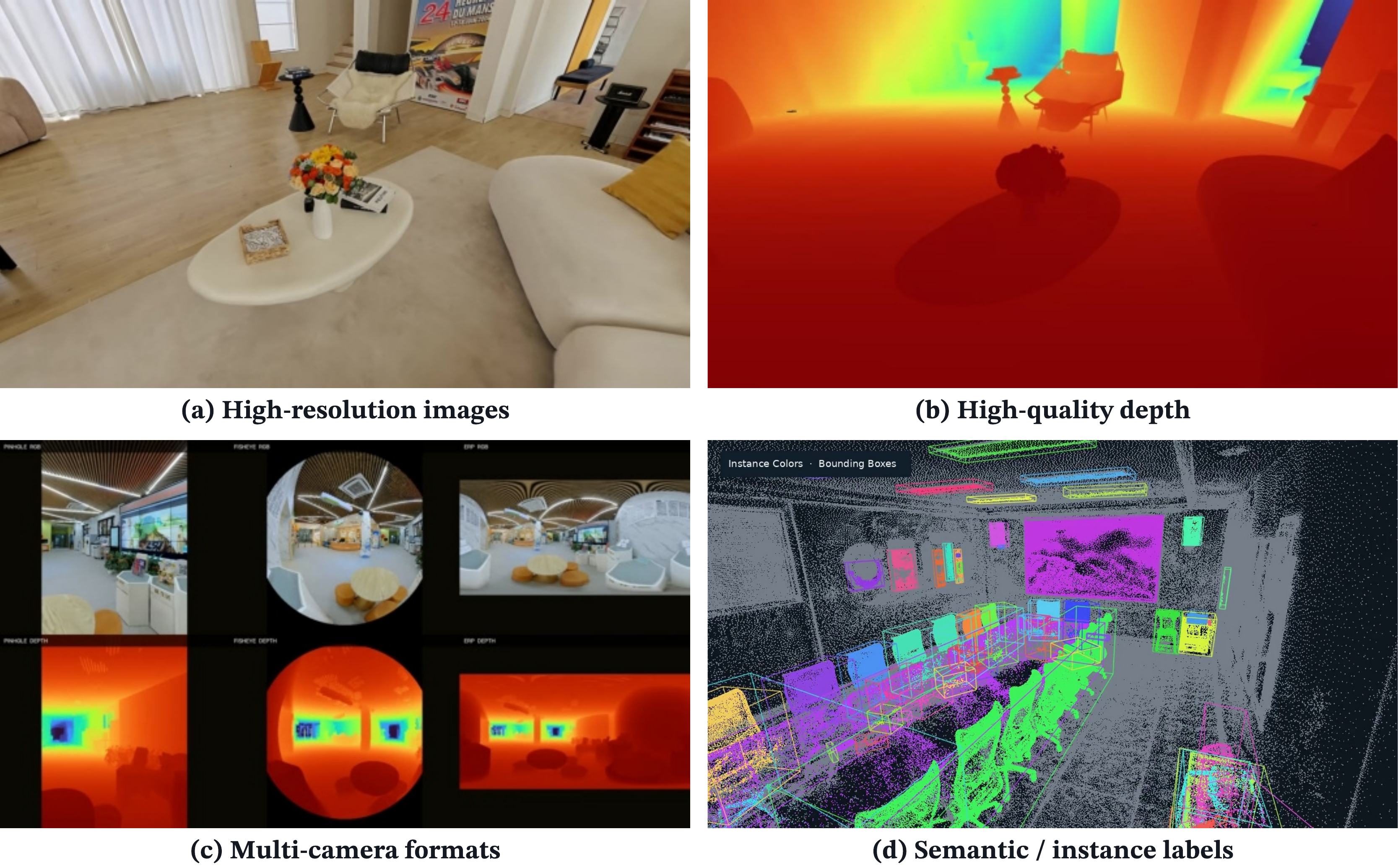}
\caption{KosmoWorld training data and supervision. The four panels show high-resolution images, depth, multi-camera formats, and semantic and instance annotations. Video panels show representative recorded frames.}
\label{fig:kosmoworld}
\end{figure}

KosmoWorld provides high-quality real-world reconstruction data for learning geometric priors. Its pipeline reconstructs real scenes, selects locations with reliable geometry and rendering quality, and generates images with corresponding depth and camera poses along configurable trajectories. The resulting data support diverse viewpoints and camera models while preserving the appearance and geometry of real scenes (Figure~\ref{fig:kosmoworld}).

\input{sections/kosmoworld-pipeline}

\subsubsection{Generated Data and Supervision}
\label{sec:kosmoworld-supervision}
Each RGB frame and its depth supervision share the corresponding camera pose and projection settings. The pipeline records coordinate systems, depth conventions, and valid-image masks so that image content and geometric supervision remain aligned. Depth is derived from the reconstructed scene, making region screening and trajectory validation integral to data quality.

Outputs are organized into sequences containing images, depth, camera intrinsics and extrinsics, and timestamps. Validation records retain the source scene, trajectory, and sample positions, allowing each training sample to be traced to its generation settings. These image, depth, and camera-pose sequences supplement GFM training.

\begin{figure}[!htbp]
\centering
\includegraphics[width=\textwidth]{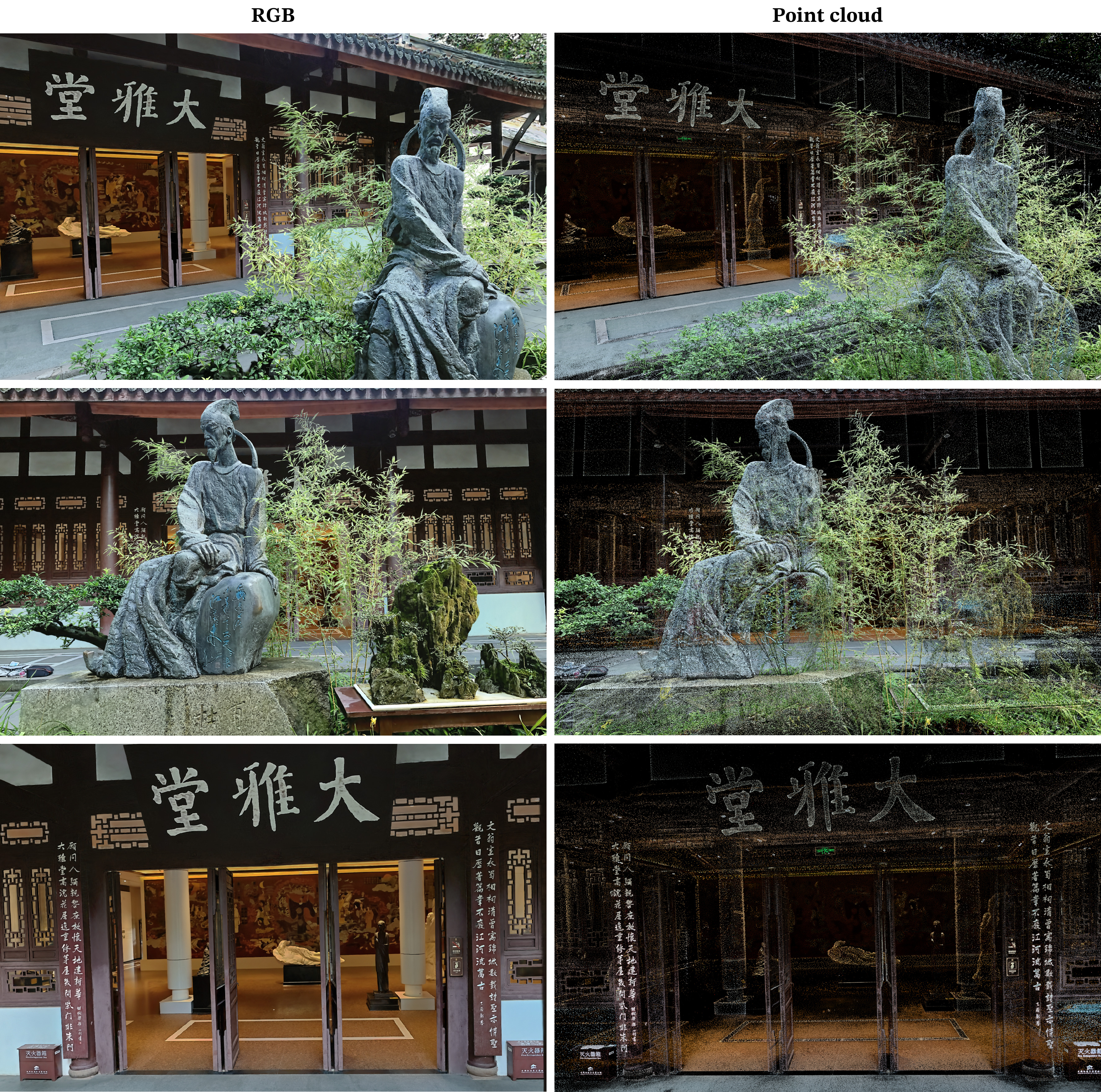}
\caption{KosmoWorld examples from three viewpoints of the same scene. Each row pairs an RGB image (left) with a point-cloud visualization (right).}
\label{fig:kosmoworld-rgb-points}
\end{figure}

Figure~\ref{fig:kosmoworld-rgb-points} shows three views of a KosmoWorld scene, pairing RGB images with point-cloud visualizations. The examples illustrate the appearance and geometric structure of the statue, vegetation, and building facade from different viewpoints.

\FloatBarrier
\subsubsection{Effect of KosmoWorld Data}
\label{sec:kosmoworld-ablation}
\begin{figure}[!htbp]
\centering
\includegraphics[width=\textwidth]{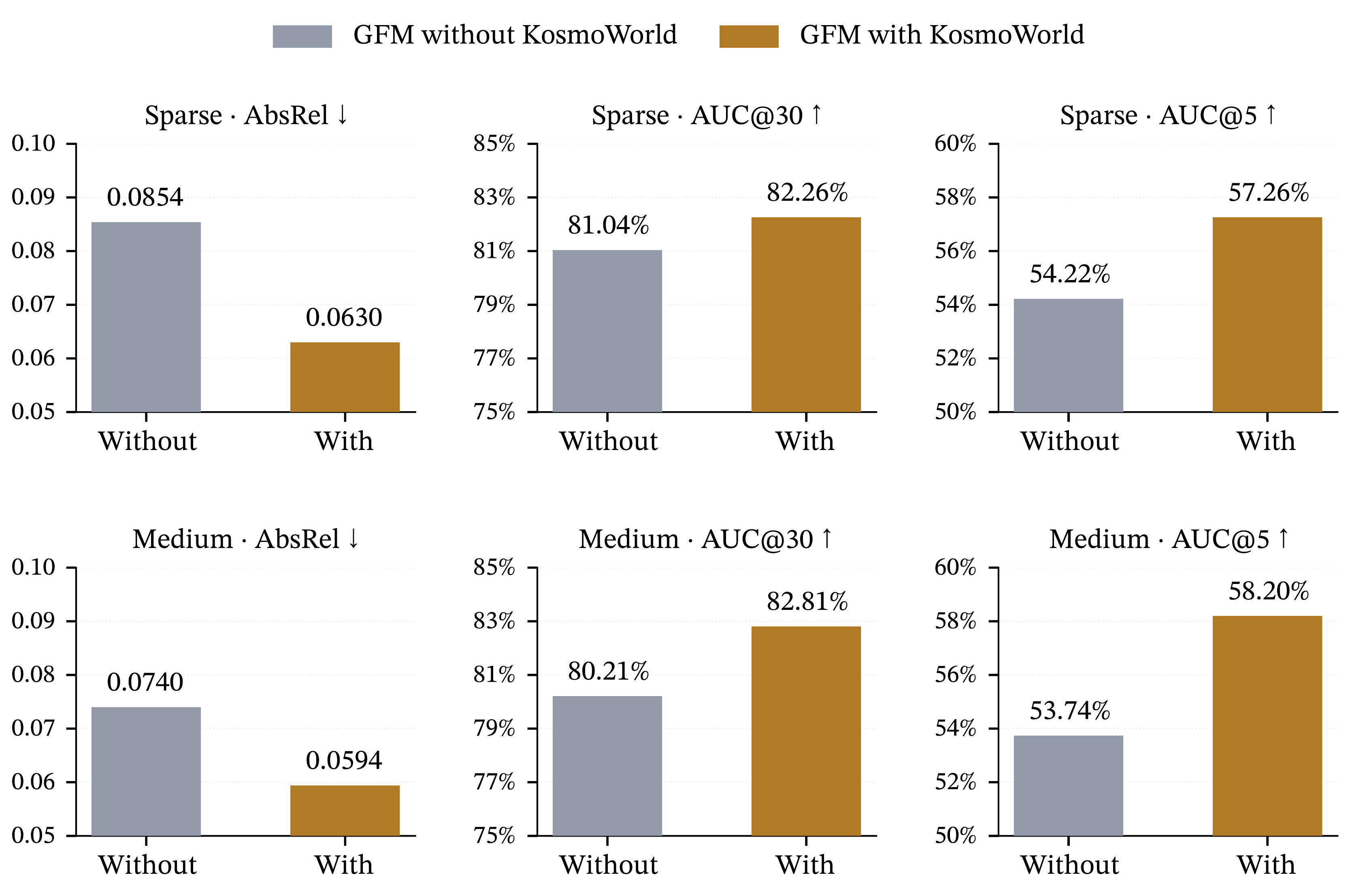}
\caption{Effect of KosmoWorld data on SpatialBench Sparse (top) and Medium (bottom). Each setting compares GFM trained with and without KosmoWorld using depth error and camera AUC. Full data-ablation metrics appear in Table~\ref{tab:ablation-full}.}
\label{fig:data-impact}
\end{figure}

We assess the contribution of KosmoWorld on the Sparse and Medium subsets of SpatialBench~\citep{peng2026spatialbench}, comparing GFM trained with and without KosmoWorld. Adding KosmoWorld consistently improves depth and relative-pose estimation in both settings (Figure~\ref{fig:data-impact} and Table~\ref{tab:ablation-full}). These results support the value of real-world reconstruction supervision for learning geometric priors.

\input{tables/ablation-full}

\section{Benchmark}
\label{sec:results}
This section evaluates Kosmo-GFM at two levels: GFM's geometric and motion prediction before scene-level optimization, and PoseEngine's camera reconstruction accuracy, efficiency, and 3DGS rendering quality. Among the compared methods, GFM achieves the lowest absolute relative depth error and leading relative-pose results on both SpatialBench Sparse and Medium, together with the lowest average endpoint errors for dynamic points and scene flow. On the 34 scenes completed by all evaluated systems, PoseEngine achieves the highest mean pose accuracy with lower reconstruction time than GLUEMAP. PoseEngine Flash delivers the fastest reconstruction among the compared systems while retaining complete image registration. These results demonstrate strong geometric prediction and offer complementary choices for reconstruction accuracy and efficiency.

\subsection{GFM}
\subsubsection{3D Reconstruction}
\label{sec:data}
\paragraph{\textbf{Evaluation setup.}}
We evaluate depth, relative camera pose, and camera trajectories on the Sparse and Medium subsets of SpatialBench~\citep{peng2026spatialbench}. Depth alignment and scale follow the benchmark protocol. For valid reference depths $z$ and predictions $\hat z$, the error metrics are

\begin{equation}
 \mathrm{AbsRel}=\left\langle\frac{|\hat z-z|}{z}\right\rangle,\quad
 \mathrm{SqRel}=\left\langle\frac{(\hat z-z)^2}{z}\right\rangle,\quad
 \mathrm{RMSE}=\sqrt{\langle(\hat z-z)^2\rangle}.
 \label{eq:depth-metrics}
\end{equation}

\input{tables/sparse-reconstruction}
We also report threshold accuracy $\delta_\tau=\langle\mathbf1[\max(\hat z/z,z/\hat z)<\tau]\rangle$, with $\tau\in\{1.03,1.05,1.10\}$. For relative pose, RAcc@$\tau$ and TAcc@$\tau$ are the fractions of image pairs whose relative rotation and translation-direction errors fall below $\tau$ degrees, and AUC@$\tau$ is the normalized area under the recall curve of the pairwise pose error, taken as the larger of the two angular errors, up to $\tau\in\{5,15,30\}$ degrees. All are reported as fractions. For trajectories, ATE is the root-mean-square error of camera centers after aligning the predicted trajectory to the reference, and RPE$_t$ and RPE$_r$ are the translational and rotational errors of relative motions between consecutive frames. Alignment and units follow the SpatialBench protocol.

\paragraph{\textbf{Depth and camera accuracy.}}
Tables~\ref{tab:sparse-reconstruction} and \ref{tab:medium-reconstruction} compare depth and camera accuracy for Sparse and Medium. GFM achieves the lowest absolute relative depth error and the highest relative-pose accuracy among the listed methods in both settings, demonstrating strong geometric prediction with both sparse and denser view inputs.

\input{tables/medium-reconstruction}
\paragraph{\textbf{Camera-conditioned reconstruction.}}
The results above use RGB-only input. Within Kosmo-GFM, refined PoseEngine cameras also condition dense geometry prediction (Section~\ref{sec:poseengine}) and the feed-forward 3DGS decoder (Section~\ref{sec:applications}). Table~\ref{tab:prompt} compares RGB-only GFM with Camera-conditioned GFM (Section~\ref{sec:camera-conditioned-gfm}), with and without per-frame scale prediction, on Sparse. The results show that camera conditioning effectively incorporates the input pose information. Predicting a per-frame scale alleviates the degradation in depth quality associated with direct normalization by the camera-pose scale. As described in Section~\ref{sec:scene-scale}, this design decouples the numerical range of per-frame depth from the camera-pose scale while preserving the input poses as a shared scale reference.

\input{tables/prompt}
\paragraph{\textbf{Performance across datasets.}}
Tables~\ref{tab:sparse-datasets} and \ref{tab:medium-datasets} report per-dataset results following the SpatialBench scene-selection and preprocessing protocol. GFM achieves strong results on robotic data and data with challenging viewpoints, with gains in both depth estimation and camera-pose prediction across diverse scene types.

\input{tables/sparse-datasets}

\input{tables/medium-datasets}

\FloatBarrier
\subsubsection{4D Reconstruction}
\paragraph{\textbf{Evaluation setup.}}
Dynamic geometry is evaluated on Panoptic Studio~\citep{joo2015panoptic}, PointOdyssey~\citep{zheng2023pointodyssey}, Dynamic Replica~\citep{karaev2023dynamicstereo}, Aria Digital Twin~\citep{pan2023adt}, and SynthVerse~\citep{zhao2026synthverse}. The first four datasets follow the WorldTrack evaluation~\citep{rw_st4rtrack} used by UniQuery4R~\citep{chen2026uniquery4r}. The reported average additionally includes SynthVerse. OpenD4RT is an unofficial implementation of D4RT~\citep{zhang2025d4rt,opend4rt2026implementation}, and the reported dynamic results use its 32CLIP variant. We measure dynamic-point APD (Average Percentage of Points within Delta), scene-flow accuracy, and 3D endpoint error (EPE), measured in meters. APD measures the percentage of evaluated dynamic points whose predicted 3D positions lie within a distance threshold of the reference positions, averaged over the evaluation thresholds. It is reported as a percentage, with higher values indicating more accurate point reconstruction. Point EPE measures position error, while flow EPE measures displacement error. Threshold-based accuracy and mean endpoint error capture different aspects of prediction quality. We report scene-flow accuracy (Acc.) as the percentage of predictions with a 3D Euclidean displacement error of at most 0.1~m.

\begin{figure}[!htbp]
\centering
\includegraphics[width=\textwidth]{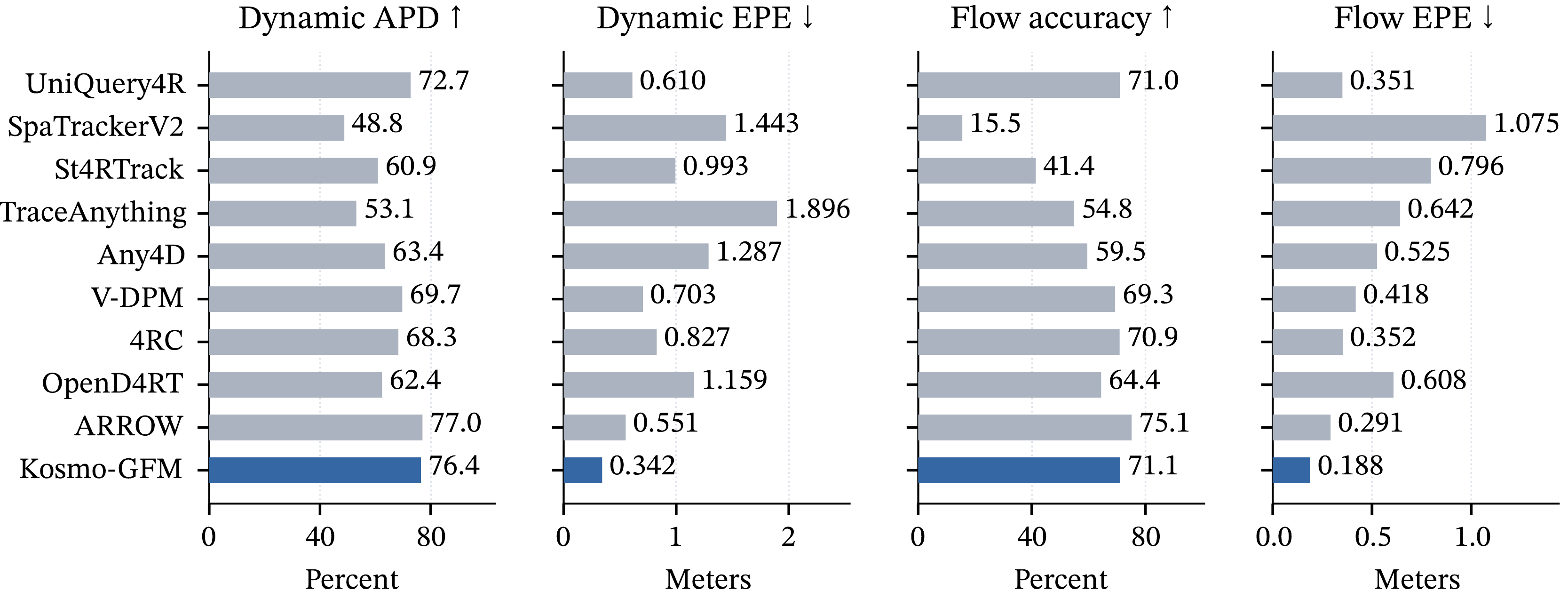}
\caption{Average dynamic-point and scene-flow performance. GFM achieves the lowest mean EPE for both tasks among the compared methods. Threshold-based accuracy and EPE characterize complementary aspects of motion prediction.}
\label{fig:motion}
\end{figure}

\paragraph{\textbf{Dynamic-point and scene-flow accuracy.}}
Compared with UniQuery4R, GFM reduces average endpoint error for both dynamic points and scene flow while retaining competitive threshold-based accuracy. Figure~\ref{fig:motion} and Table~\ref{tab:motion} summarize these complementary aspects of motion prediction.

\input{tables/motion}
GFM achieves the lowest average endpoint error for both dynamic points and scene flow. Tables~\ref{tab:motion-point-accuracy} and \ref{tab:motion-flow-accuracy} further show its strengths in dynamic geometry and motion prediction, including leading point reconstruction on PointOdyssey and SynthVerse and leading scene-flow accuracy and endpoint error on Aria Digital Twin and SynthVerse.

\input{tables/motion-point-accuracy}

\input{tables/motion-flow-accuracy}

\subsection{PoseEngine}
\label{sec:system-results}
We select 25 ETH3D~\citep{schoeps2017eth3d}, nine MipNeRF360~\citep{barron2022mipnerf360}, six Tanks \& Temples~\citep{knapitsch2017tanks}, and four SMERF~\citep{duckworth2024smerf} scenes for scene-level reconstruction and compare COLMAP~\citep{schoenberger2016sfm}, GLUEMAP~\citep{pan2026gluemap}, VidMap~\citep{pataki2026vidmap}, PoseEngine, and PoseEngine Flash. In this evaluation, VidMap fails on ten ETH3D scenes because its evaluated implementation cannot handle inconsistent image resolutions within a scene. The common comparison therefore uses the 34 scenes completed by all methods: 15 ETH3D, nine MipNeRF360, six Tanks \& Temples, and four SMERF scenes, with 11,336 input images in total.

\input{sections/pose-scenes}

\subsubsection{Efficiency and Metrics}

\paragraph{\textbf{Pose and trajectory metrics.}}
For each image pair, pose error is the maximum of rotation error and translation-direction error. AUC@$a$ integrates the recall curve up to $a$ degrees and normalizes by $a$, and we report AUC@1/3/5 as percentages. R@$a$ and T@$a$ threshold the angular components separately. Evaluation uses all ordered pairs, including self pairs, among each method's registered frames with ground truth, followed by equal-weight scene averaging. Registration coverage is reported because the evaluated frame sets can differ across methods.

Focal-length AbsRel is computed after rescaling predictions to the reference image resolution, averaging horizontal and vertical errors over valid cameras. Trajectory error uses camera centers aligned by Sim(3):

\begin{equation}
 \mathrm{ATE}_{\rm nRMSE}=
 \frac{\sqrt{\frac1N\sum_i\|\tilde C_i-C_i\|_2^2}}
 {\sqrt{\frac1N\sum_i\|C_i-\bar C\|_2^2}}.
 \label{eq:ate-metric}
\end{equation}
Here $\tilde C_i$ and $C_i$ are aligned predicted and reference centers, and $\bar C$ is the reference centroid. This normalized ATE is dimensionless and differs from the unnormalized ATE reported by SpatialBench.

\paragraph{\textbf{Timing protocol.}}
Runtime is the sum of per-scene reconstruction durations. COLMAP uses SIFT~\citep{lowe2004sift} with exhaustive matching and shared PINHOLE intrinsics. GLUEMAP uses $\pi^3$~\citep{rw_pi3} with shared SIMPLE\_PINHOLE intrinsics. VidMap uses its uncalibrated base setting with RoMa~\citep{edstedt2023roma} matching and DA3~\citep{lin2025da3} depth. All three use twelve CPU threads, and GLUEMAP and VidMap use four data-loading workers. VidMap timing includes model loading and compilation from a fresh cache.

PoseEngine Flash runtime is measured using FP8/FP16 mixed-precision inference under the same timing protocol.

\begin{figure}[!htbp]
\centering
\includegraphics[width=\textwidth]{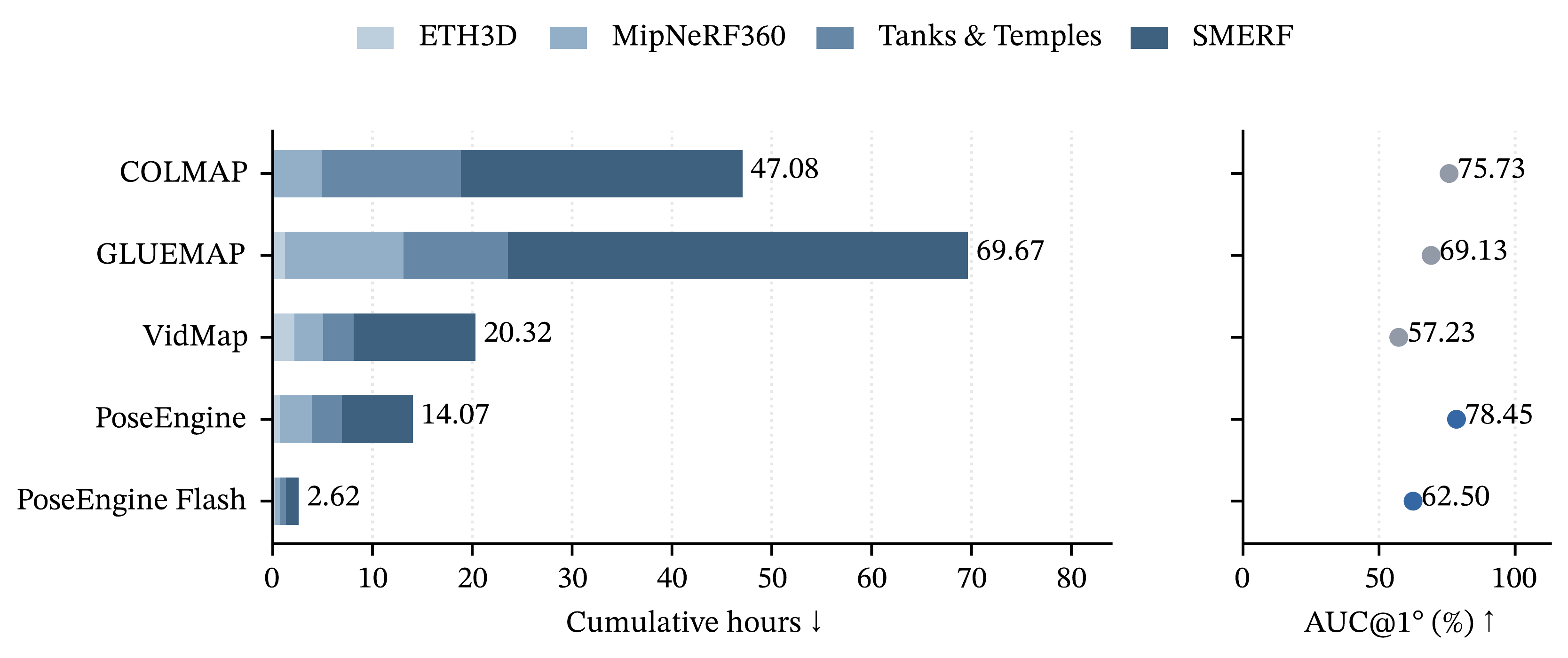}
\caption{Cumulative reconstruction time and pose accuracy on 34 scenes. Time is decomposed by dataset. Flash uses measured mixed-precision runtime. Pose accuracy is evaluated on registered frames.}
\label{fig:runtime}
\end{figure}

\paragraph{\textbf{Reconstruction accuracy and efficiency.}}
Table~\ref{tab:pose} summarizes reconstruction quality and coverage, and Figure~\ref{fig:runtime} breaks down cumulative runtime by dataset. PoseEngine achieves the highest mean pose accuracy among the compared methods while reducing reconstruction time relative to GLUEMAP. It also obtains the lowest focal-length and normalized trajectory errors. PoseEngine Flash delivers the fastest reconstruction while retaining complete image registration and low trajectory error.

\input{tables/pose-summary}

\subsubsection{Matching}
\paragraph{\textbf{Evaluation setup.}}
We compare MVLoMa, which shares GFM features, with LoMa Large~\citep{nordstrom2026loma} under unordered and ordered STAR construction (Table~\ref{tab:matching}). This study uses all 44 scenes selected from ETH3D, MipNeRF360, Tanks \& Temples, and SMERF, and adds four Oxford Spires~\citep{tao2024oxfordspires} sequences, bringing the total to 48. The additional sequences are Keble College (2024-03-12, sequence 04), Observatory Quarter (2024-03-13, sequence 01), Blenheim Palace (2024-03-14, sequence 05), and Christ Church (2024-03-18, sequence 02).

\paragraph{\textbf{Higher pose accuracy across datasets.}}
MVLoMa improves strict-threshold pose accuracy over LoMa Large on every evaluated dataset under both STAR policies. The larger average gain with ordered STARs highlights the benefit of using shared multi-view features for matching in sequential reconstruction.

\paragraph{\textbf{More consistent camera trajectories.}}
MVLoMa also reduces the mean trajectory error after Sim(3) alignment on every evaluated dataset under both policies. Together, these results show that its matching improves both relative-pose accuracy and scene-level trajectory consistency.

\input{tables/matching}

\subsubsection{3DGS}
\label{sec:rendering}
\paragraph{\textbf{Rendering protocol.}}
We evaluate 3DGS reconstruction on nine MipNeRF360 and six Tanks \& Temples scenes using LiteGS~\citep{liao2025litegs}. Each model is trained for 30,000 iterations, with every eighth image in filename order held out for evaluation. The reference-pose baseline keeps cameras fixed, whereas other initializations jointly optimize training poses and Gaussians. Before evaluation, each non-reference test pose is refined against its test image for twenty steps with Gaussians and intrinsics fixed. PSNR (dB) and SSIM~\citep{wang2004ssim} measure image fidelity, and VGG LPIPS~\citep{zhang2018lpips} measures perceptual distance. These scores evaluate rendering after image-conditioned pose refinement, not fixed-pose generalization. Tanks \& Temples uses trajectory-aligned PoseEngine initialization, while MipNeRF360 uses initialization from sparse camera chunks. Comparisons are made within each dataset.

\input{tables/rendering}
\paragraph{\textbf{Rendering quality.}}
PoseEngine cameras provide an effective initialization for downstream 3DGS reconstruction (Table~\ref{tab:rendering}). On Tanks \& Temples, joint camera and Gaussian optimization yields rendering quality close to the reference-pose baseline. The selected image comparisons also show similar overall scene appearance.

\begin{figure}[!htbp]
\centering
\includegraphics[width=\textwidth,height=.82\textheight,keepaspectratio]{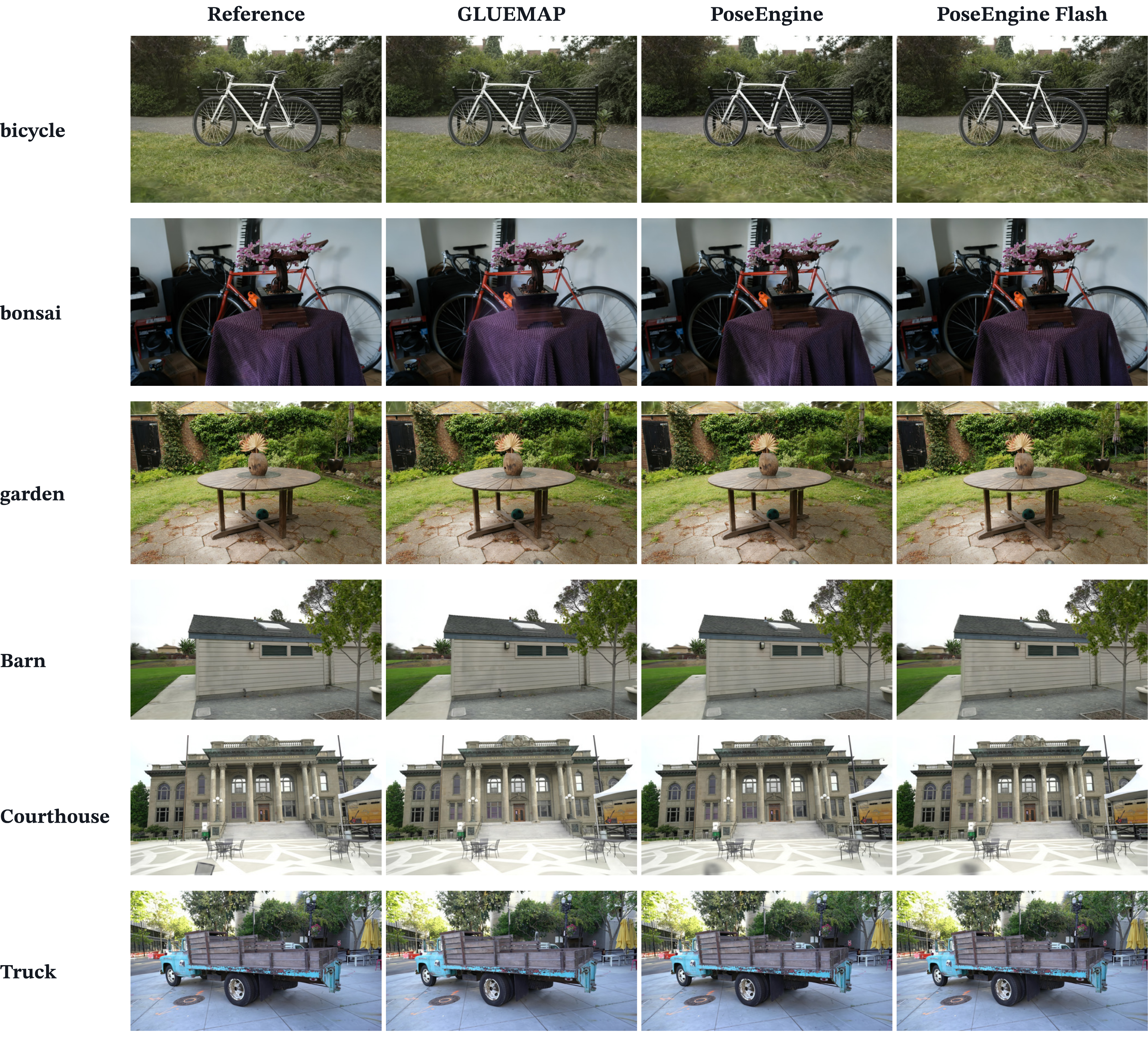}
\caption{3DGS rendering on three examples each from MipNeRF360 and Tanks \& Temples. Columns use reference cameras, GLUEMAP, PoseEngine, and PoseEngine Flash. Within each row, all renderings correspond to the same test image, and non-reference poses receive twenty-step image-conditioned refinement.}
\label{fig:render-comparison}
\end{figure}

Figure~\ref{fig:render-comparison} compares three examples each from MipNeRF360 and Tanks \& Temples using the same test image within each row. Vegetation and thin structures reveal appearance differences that are not fully captured by aggregate pose metrics. Quantitative rendering results aggregated over all fifteen scenes appear in Table~\ref{tab:rendering}.

\section{Scaling}
\label{sec:data-scaling}
\label{sec:scaling}
Given limited open-source 3D training data and our computational resource constraints, we do not establish a quantitative scaling law for GFM. Instead, we examine whether increasing training-data quantity, model scale, and training compute continues to improve depth and camera-pose accuracy within the budgets we can evaluate. After selecting an encoder initialization, we study these three axes using experimental models that are separate from the released GFM. The results show favorable scaling trends overall, with metric-level fluctuations and diminishing gains at larger recurrent depths. These observations motivate further scaling, but do not establish a power-law scaling relationship that can be extrapolated beyond the tested range.

\paragraph{\textbf{Scaling protocol.}}
The three sweeps report SpatialBench Sparse depth and camera-pose accuracy in the accompanying plots and tables. Values in the training-compute plots are rounded to three decimal places.

\subsection{Encoder Selection}
\label{sec:scaling-encoder}
GFM models rely on visual encoders initialized from pretrained models such as DINOv2~\citep{oquab2023dinov2}. The available pretrained checkpoints therefore constrain the model scales that can be studied directly.

We compare the Base variants of DINOv2, DINOv3~\citep{simeoni2025dinov3}, and LingBot-Vision~\citep{fu2026lingbotvision} as encoder initializations, training base-size reconstruction models for 100k iterations on Hypersim~\citep{roberts2021hypersim} with 2--24 input views. LingBot-Vision performs best among the three general-purpose encoders across the reported depth and pose measures (Table~\ref{tab:scaling-encoder}). We therefore use pretrained weights from the LingBot-Vision family to initialize the encoders in the subsequent scaling experiments.

\input{tables/scaling-encoder}

To study intermediate scales without separately pretraining larger visual encoders, we adopt a recurrent-depth Transformer approach that reuses the 1.7B model's weights across additional computational depth. This produces the 2.5B, 3.2B, and 4.8B variants evaluated below. For these recurrent variants, the reported scale counts parameter uses in the expanded computation graph, including repeated uses of shared weights; it is not the number of independent trainable parameters. These variants let us probe scaling through increased recurrent computation while retaining the available pretrained initialization.

\subsection{Data Scale}
\begin{figure}[!htbp]
  \centering
  \includegraphics[width=\textwidth]{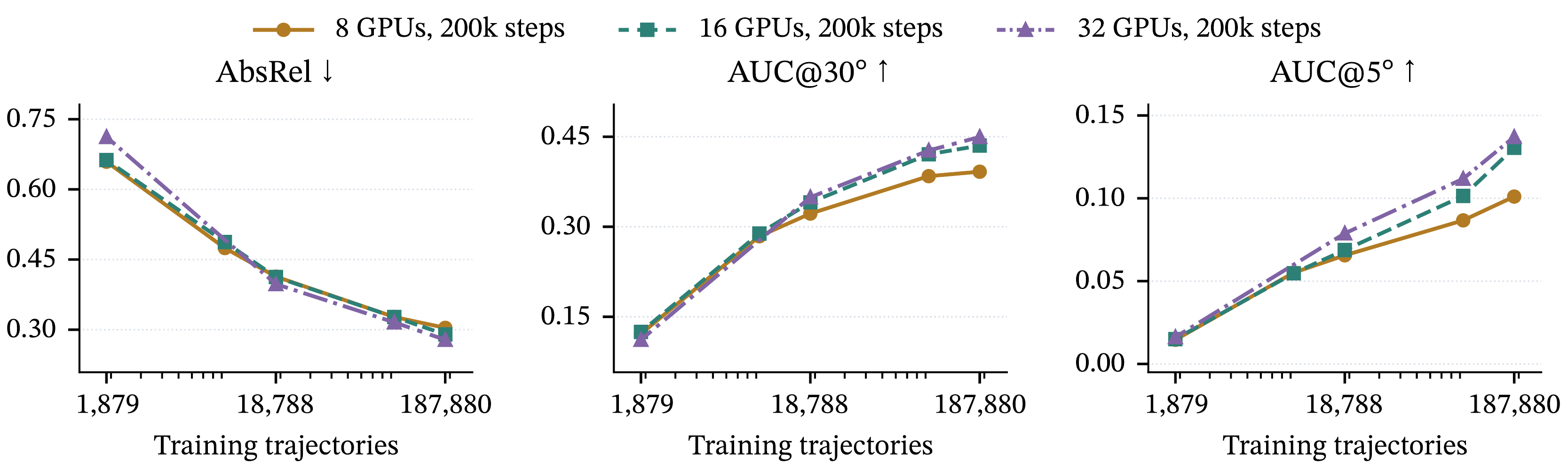}
  \caption{Data scaling for the 0.2B model. More training trajectories
  reduce depth error and improve pose accuracy at each tested budget.
  The x-axis is logarithmic. Lines connect measured observations only.}
  \label{fig:scaling-data}
\end{figure}

We select 26 publicly available datasets and construct training sets of different
sizes by sampling 1\% to 100\% of their trajectories, ranging from 1,879
to 187,880 trajectories. For a 0.2B model, more training data improves depth
estimation and camera-pose accuracy at each of the three tested budgets.
Figure~\ref{fig:scaling-data} shows the 8-, 16-, and 32-GPU series at 200k
training steps, and the trend holds across dataset sizes within each
series.

\input{tables/scaling-data-values}

\begingroup
\setlength{\intextsep}{3pt}
\subsection{Model Scale}
\label{sec:scaling-model}
Model-scale experiments use the full training set of 187,880 trajectories.
For models trained on eight GPUs for 200,000 steps, increasing model scale from 0.6B to 3.2B
consistently improves depth and camera-pose accuracy
(Figure~\ref{fig:scaling-model}). The 2.5B and 3.2B variants use the recurrent-depth construction described in Section~\ref{sec:scaling-encoder}. This comparison therefore includes both changes in independent model parameters and increases in recurrent computational depth.

\begin{figure}[!htbp]
  \centering
  \includegraphics[width=\textwidth]{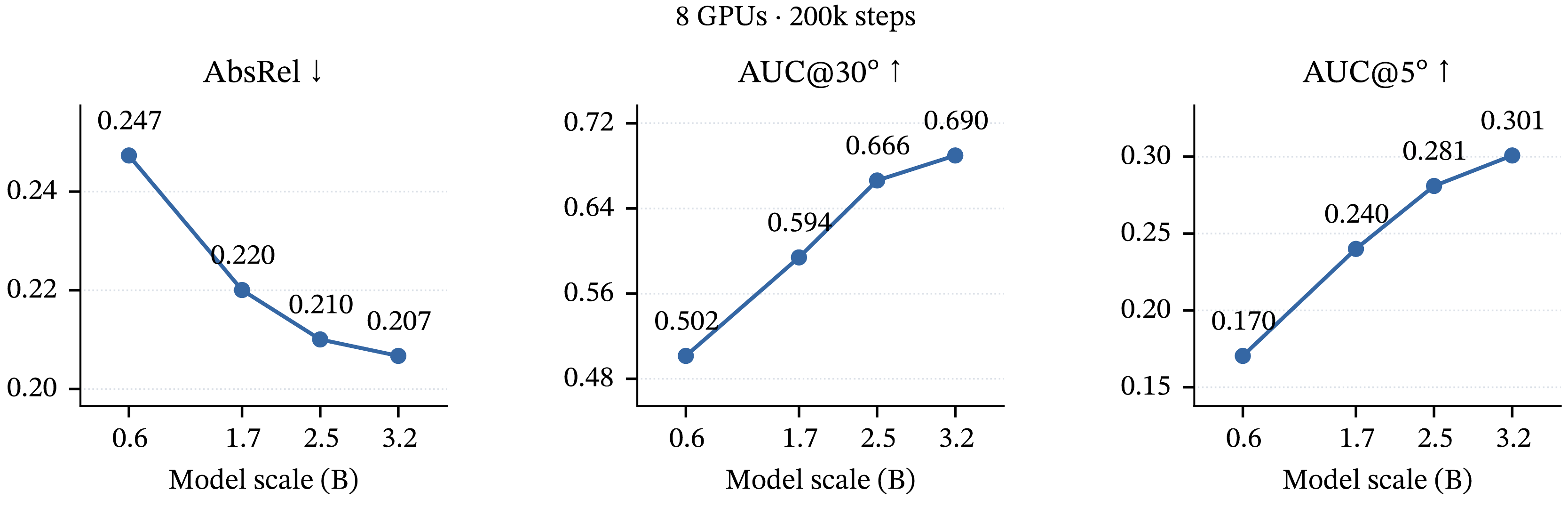}
  \caption{Model scaling using 8 GPUs for 200k steps. Depth and pose accuracy
  improve as model scale increases.}
  \label{fig:scaling-model}
\end{figure}

\subsection{Training Compute}
We evaluate training-compute scaling for four model configurations, fixing the training length at 200,000 steps and increasing the GPU count from 8 to 16 and 32. Figure~\ref{fig:scaling-compute} uses GPU count multiplied by training steps as a budget proxy, corresponding to 1.6M, 3.2M, and 6.4M GPU-steps. We use this proxy to organize the budget sweep within each model configuration; the 3.2B configuration uses recurrent depth as described in Section~\ref{sec:scaling-encoder}.

\begin{figure}[!htbp]
  \centering
  \includegraphics[width=\textwidth]{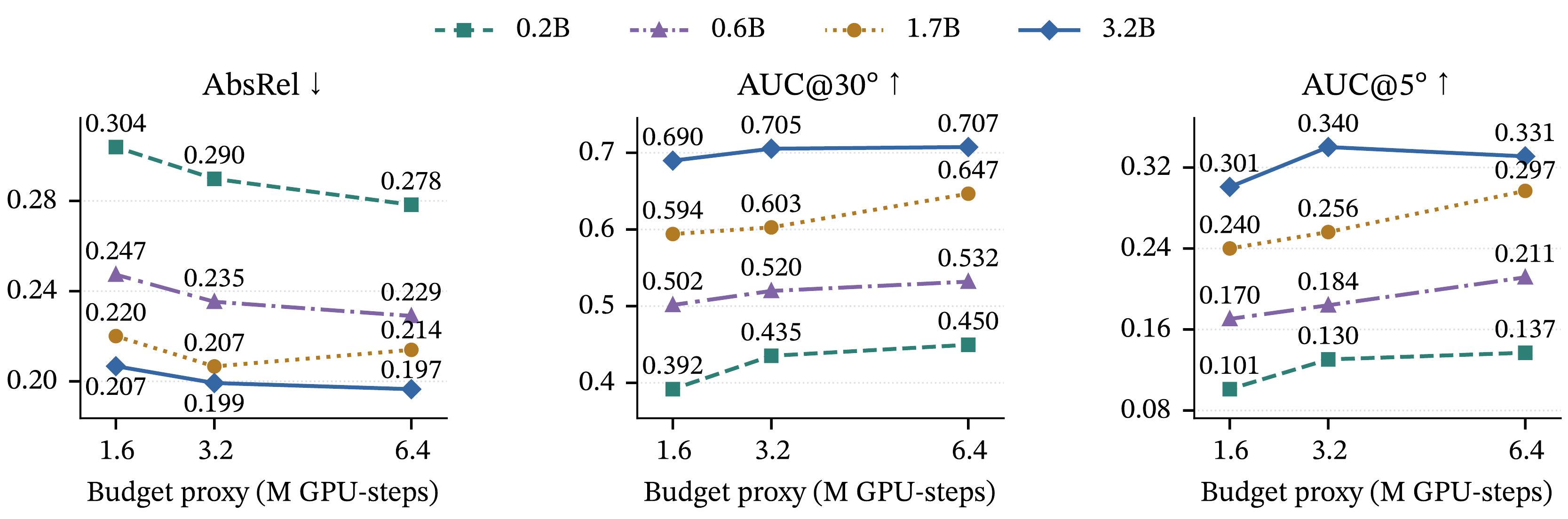}
  \caption{Training-compute scaling across four model scales. Each series uses 200k training steps with 8, 16, or 32 GPUs; the horizontal axis reports the corresponding GPU-step budget.}
  \label{fig:scaling-compute}
\end{figure}

Increasing training compute generally reduces depth error and improves camera-pose accuracy, although individual metrics fluctuate. For the 3.2B variant, however, gains become smaller at higher budgets, suggesting a trend toward saturation.

\Needspace{80pt}

\smallskip\noindent\textbf{Two hypotheses.}
We consider two possible explanations for this trend. First, the current training set of 187,880 trajectories may be insufficient in size or diversity to support further improvements at this model scale. Second, repeated applications of shared Transformer weights may yield progressively smaller representational updates as recurrent depth increases, reducing the gains from additional passes. We refer to this possible effect as recurrent ``inertia.''

\Needspace{330pt}
\smallskip\noindent\textbf{Additional recurrent-depth experiments.}
To further examine the second hypothesis, we extend the model-scale comparison in Section~\ref{sec:scaling-model} to the 4.8B recurrent variant, alongside the 2.5B and 3.2B variants (Figure~\ref{fig:scaling-recurrent}). All variants are trained on the full set of 187,880 trajectories using 8 GPUs and 200k steps. The recurrent variants share the 1.7B model's weights and follow the scale convention defined in Section~\ref{sec:scaling-encoder}.

The results show that increasing recurrent depth continues to improve accuracy, while the gains per increment in scale become smaller, particularly for AbsRel and AUC@30. From 3.2B to 4.8B, AbsRel decreases from 0.2067 to 0.2049, AUC@30 increases from 0.6898 to 0.7007, and AUC@5 increases from 0.3008 to 0.3216. This trend is consistent with diminishing marginal returns from recurrent depth.

\begin{figure}[!htbp]
  \centering
  \includegraphics[width=\textwidth]{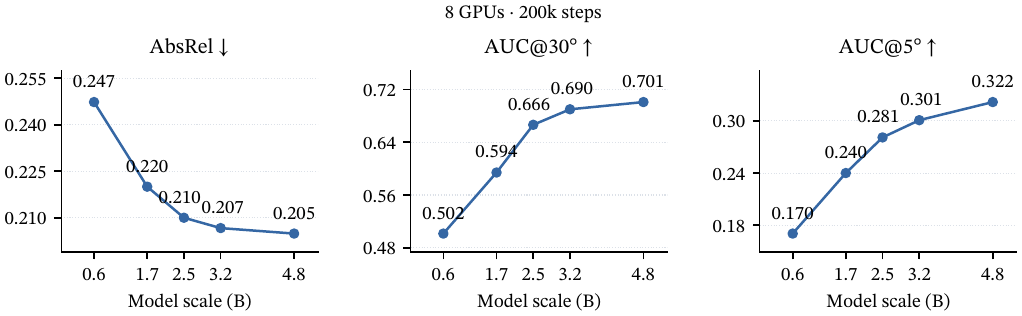}
  \caption{Recurrent-depth scaling using 8 GPUs for 200k steps. The 2.5B, 3.2B, and 4.8B variants reuse the 1.7B model's weights; their reported scales count parameter uses in the expanded computation graph, including weight reuse, rather than unique trainable parameters. The 0.6B and 1.7B models serve as references. Accuracy continues to improve at larger scales, with diminishing marginal gains.}
  \label{fig:scaling-recurrent}
\end{figure}
\FloatBarrier

\smallskip\noindent\textbf{Related work and outlook.}
Related observations have been reported in other multi-view reconstruction models. D\'ej\`a View~\citep{burzio2026dejaview} observes decreasing relative feature updates and reports a plateau beyond the trained iteration budget under a stabilization recipe. Solv3R~\citep{kwak2026solv3r} reports diminishing state updates and camera errors that stabilize with continued iteration. The favorable trends observed in this section support further exploration of data, model, and compute scaling. We remain optimistic about recurrent depth as a practical way to increase computation through weight reuse, and will continue investigating larger training sets and improved recurrence designs. Establishing a quantitative scaling law remains a goal for experiments spanning a wider range of data and compute budgets.

\endgroup

\input{sections/reconstruction}

\section{Discussion and Limitations}
\label{sec:discussion}
Kosmo-GFM combines learned geometric priors with scene-level optimization to support both static and dynamic reconstruction. We discuss limitations in static 3D geometry and dynamic 4D modeling separately, then consider applications and representation-learning opportunities enabled by 3D point tracking.

\subsection{Static 3D Reconstruction}
\label{sec:discussion-static-3d}

\noindent\textbf{Textureless scenes.} Large blank surfaces offer few distinctive features or reliable cross-view correspondences. The resulting lack of visual constraints makes camera recovery unreliable, as illustrated by the stairwell example in Figure~\ref{fig:limitation-examples}(a).

\noindent\textbf{Portrait-oriented inputs.} Our training primarily targets landscape-oriented images and videos. Portrait-oriented inputs are less well covered during training, resulting in weaker reconstruction performance, as illustrated by the indoor video in Figure~\ref{fig:limitation-examples}(b).

\noindent\textbf{Unstructured photo collections.} Training data emphasize walkthrough trajectories, whereas independently captured Internet photos vary substantially in viewpoint, framing, appearance, and view overlap. Limited training coverage of these capture patterns reduces robustness on datasets such as MegaDepth~\citep{li2018megadepth}, MegaDepth-X~\citep{li2026megadepthx}, and ArchWorld~\citep{kriplani2026archworld}. Figure~\ref{fig:limitation-examples}(c) illustrates the resulting errors in recovered camera poses on 7 West 54th Street.

\input{sections/limitation-examples}

\subsection{Dynamic 4D Reconstruction}
\label{sec:discussion-dynamic-4d}
\noindent\textbf{Static--dynamic training balance.}
Our current GFM training emphasizes dynamic sequences, with static
scenes accounting for a smaller share of the training mixture.
This allocation may leave some static environments underrepresented
and contribute to errors in depth estimation and cross-view scale
consistency. Establishing the extent of this trade-off requires
controlled ablations of the training mixture.

We are evaluating different mixing ratios of high-quality static
data and dynamic sequences through controlled experiments.
These experiments aim to improve depth accuracy and cross-view
scale consistency while preserving dynamic reconstruction
performance.

\noindent\textbf{Decoder context and temporal conditioning.}
Our dynamic-point decoder adopts the source-to-target formulation of
UniQuery4R~\citep{chen2026uniquery4r}: a query sampled from the source-frame
features cross-attends to the selected target-frame features.
D4RT~\citep{zhang2025d4rt} instead lets a source-point query, augmented
with local appearance, attend to a global representation of all input
frames, using a learned target-time embedding to specify the requested
temporal state. In our formulation, global attention in the encoder
already exchanges information across all input frames. The selected
source and target representations therefore provide a compact interface
to this multi-frame context, reducing the information explicitly accessed
by the decoder. Our design hypothesis is that this encoder-level
aggregation can make additional attention to other frames unnecessary
at decoding time; this remains to be established through controlled
ablations. Selecting the target features also identifies the requested
frame directly, removing dependence on a learned target-time embedding
indexed by a predefined frame set.

\noindent\textbf{Occlusion and temporal stability.}
We nevertheless observe temporal jitter in predicted 3D trajectories
when the queried point is occluded in the target frame. Such cases lack
direct target-view evidence and test whether the encoded context
preserves sufficient information to infer hidden motion. We will compare
decoder variants under matched training conditions and vary the amount
and coverage of occlusion supervision to assess the respective
contributions of decoder modeling and training-data limitations.

\begingroup
\makeatletter
\patchcmd{\subsubsection}{-3.25ex\@plus -1ex \@minus -.2ex}{-2ex\@plus -.5ex \@minus -.2ex}{}{\PackageError{layout}{Tracking heading before-spacing patch failed}{}}
\patchcmd{\subsubsection}{1.5ex \@plus .2ex}{1ex \@plus .2ex}{}{\PackageError{layout}{Tracking heading after-spacing patch failed}{}}
\patchcmd{\paragraph}{1.75ex \@plus.5ex \@minus.2ex}{1.25ex \@plus.3ex \@minus.2ex}{}{\PackageError{layout}{Tracking group spacing patch failed}{}}
\makeatother
\newenvironment{trackingpoints}
  {\begin{list}{}{%
    \setlength{\leftmargin}{0pt}%
    \setlength{\labelwidth}{0pt}%
    \setlength{\labelsep}{0pt}%
    \setlength{\rightmargin}{0pt}%
    \setlength{\topsep}{2pt}%
    \setlength{\partopsep}{0pt}%
    \setlength{\itemsep}{2.5pt}%
    \setlength{\parsep}{0pt}%
  }}
  {\end{list}}

\subsubsection{Applications of 3D Point Tracking}
\label{sec:tracking-applications}
3D point tracking can support downstream applications and inform representation learning.

\paragraph{Direct applications.}
We consider three uses of predicted geometry and motion in downstream systems.
\begin{trackingpoints}
\item \textit{4DGS initialization and motion priors.}
Dense dynamic point maps can initialize 4DGS geometry, while time-normalized
scene flow offers velocity priors as an alternative to FreeTimeGS's
RoMa/KNN pipeline~\citep{wang2025freetimegs}.
MoRSe~\citep{jin2026morse} uses a UniQuery4R-initialized
encoder to predict moving-region masks for pixel-level static--dynamic
separation. These masks can guide primitive allocation, and
confidence-weighted trajectories can supervise Gaussian motion.

\item \textit{Frame interpolation.}
Estimated trajectories can support frame interpolation by interpolating
3D states between observations and rendering or warping them to
intermediate times, with visibility handling and image completion for
newly exposed regions.

\item \textit{Camera reshooting.}
Holo4D~\citep{lan2026holo4d}, for example, projects persistent 3D
trajectories into requested camera views to condition video diffusion.
Following the same scene points across time supplies a geometric basis
for preserving motion and identity during video reshooting, beyond the
spatial information in independent point clouds.
\end{trackingpoints}

\paragraph{Representation learning.}
Joint supervision of 3D tracks, scene flow, and visibility may also
benefit the shared representation.
\begin{trackingpoints}
\item \textit{Temporal correspondence and point identity.}
Tracking supervision can encourage the model to preserve point identity across appearance changes and occlusions. TrackMAE~\citep{vandeghen2026trackmae} supports this direction
through 2D track reconstruction in masked video modeling; benefits from
extending such supervision to joint 3D geometry and motion learning in
Kosmo-GFM remain to be evaluated.

\item \textit{Separating camera and scene motion.}
Motion expressed in a shared reference frame can help distinguish
camera-induced image motion from scene dynamics, while relative motion
provides cues to articulated parts and coherent motion regions.

\item \textit{Geometric and motion consistency.}
Consistency among depth, camera geometry, 2D correspondences, and 3D
trajectories offers additional constraints on the representation.

\item \textit{Embodied future prediction.}
FlowDreamer~\citep{guo2025flowdreamer} predicts 3D scene flow from
observations and robot actions to condition future RGB-D synthesis.
GFM could provide confidence-filtered trajectories or geometric features
for such action-conditioned predictors, whose inputs must be restricted
to observed history; full-clip reconstruction alone does not establish
forecasting ability. These potential benefits require ablations under
matched data and compute budgets, together with downstream reconstruction
or control evaluations.
\end{trackingpoints}
\endgroup

Overall, Kosmo-GFM combines multi-task geometric prediction, scene-level camera recovery, and real-world supervision to support static and dynamic reconstruction. The reported accuracy, efficiency, and scaling results motivate extending this framework to a wider range of real-world capture conditions.

\clearpage
\phantomsection
\addcontentsline{toc}{section}{References}
\bibliographystyle{plainnat}
\bibliography{references,references-related,references-extra,references-datasets,references-additions}

\Needspace{110pt}
\section*{Acknowledgments}
\addcontentsline{toc}{section}{Acknowledgments}
We thank the following authors and interns for their work on this project.

\begingroup
\small
\setlength{\parskip}{4pt}
\noindent \textbf{Authors.}\enspace \mbox{Sheng Tang}, \mbox{Shengzhe Wei}, \mbox{Tiancheng Chen}, \mbox{Wenhua Jin}, \mbox{Yu Zhang}, \mbox{Xingshuo Chang}, \mbox{Jituo Tian}, \mbox{Hui Liang}, \mbox{Chaoqun Tang}, \mbox{Feng Tian}, \mbox{Xiaolin Chai}, \mbox{Kaojin Zhu}, \mbox{Wentao Ouyang}, \mbox{Peihuang Fu}, \mbox{Miao Zheng}, \mbox{Zesong Li}, \mbox{Xiao Lei} and \mbox{Xiang Gao} (former team members).\par
\noindent \textbf{Interns.}\enspace \mbox{Tinghao Jiang}, \mbox{Xinyang Lu}, \mbox{Feixue Cai}, \mbox{Zizhen Zou}, \mbox{Weigang Wang}, \mbox{Yexuan Yang}, \mbox{Jiahao Li}, \mbox{Haoming Liu}, \mbox{Yupeng Liu}, \mbox{Shuoxuan Feng}, \mbox{Haoyu Feng}, \mbox{Keren Zhou}, \mbox{Yuhan Xiong}, \mbox{Yunbo Hu}, \mbox{Weiqi Zhang}, \mbox{Juntong Fang}.\par
\endgroup

\end{document}

%% file: sections/related-work.tex
Learned geometric prediction, cross-view matching, and scene-level optimization address complementary requirements of 3D and dynamic reconstruction. Learned models provide geometry and motion priors, matching connects observations across views, and scene-level optimization strengthens geometric consistency. This section reviews 3D reconstruction, 4D reconstruction, matching, and pose estimation systems from this perspective. Kosmo-GFM combines a shared geometry model with PoseEngine and reconstruction-derived training supervision from KosmoWorld, bringing learned prediction and geometric constraints together within a single reconstruction system.

\subsection{3D Reconstruction}
\label{sec:related-3d}
Static reconstruction recovers scene geometry and camera parameters from
images or videos. Beyond prediction accuracy, its practical scope depends
on how observations are processed and which camera models are supported.

\paragraph{\textbf{End-to-end reconstruction.}}
DUSt3R formulates reconstruction from uncalibrated image pairs as point-map
regression~\citep{wang2023dust3r}, and MASt3R couples this geometry with local
descriptors for matching~\citep{leroy2024mast3r}. Fast3R and MUSt3R extend
pairwise prediction to multiple views in a common coordinate frame,
reducing the need to align separate pairwise reconstructions
\citep{rw_fast3r,rw_must3r}. VGGT predicts multiple geometric attributes
jointly~\citep{wang2025vggt}, whereas $\pi^3$ removes dependence on a designated
reference view through a permutation-equivariant architecture~\citep{rw_pi3}.
SAIL-Recon scales scene regression by localizing additional images against
an anchor-based neural scene representation~\citep{rw_sailrecon}, while
FastVGGT reduces inference cost through training-free token
merging~\citep{rw_fastvggt}.

Another direction broadens the available inputs: Depth Anything 3 supports
geometry recovery with or without known camera poses~\citep{lin2025da3}, while
MapAnything and WorldMirror incorporate optional geometric observations or
priors into feed-forward reconstruction~\citep{keetha2025mapanything,rw_worldmirror}.
OmniVGGT further studies the injection and fusion of auxiliary geometric
modalities~\citep{rw_omnivggt}. AMB3R uses a compact volumetric backend for
metric-scale geometric reasoning~\citep{rw_amb3r}, and VGGT-$\Omega$ examines
scaling model capacity and training data for static and dynamic
reconstruction~\citep{rw_vggtomega}. These approaches extend direct
prediction through richer conditioning, scene representations, and training
scale, rather than treating image count as the only dimension of scalability.

\paragraph{\textbf{Online reconstruction.}}
Streaming methods update a scene as observations arrive, balancing the
retention of historical geometry against memory use and latency. Spann3R
maintains external spatial memory~\citep{rw_spann3r}, while CUT3R carries a
persistent recurrent state across observations~\citep{wang2025cut3r}.
Point3R associates explicit spatial pointers with scene
locations~\citep{rw_point3r}. StreamVGGT uses causal attention and cached
features~\citep{rw_streamvggt}, and STream3R formulates sequential point-map
prediction with a causal decoder-only transformer~\citep{rw_stream3r}.
WinT3R combines windowed inference with a global camera-token
pool~\citep{rw_wint3r}.

Extending inference beyond the training sequence length also requires
controlling drift and forgetting. TTT3R
reinterprets memory updates as online learning~\citep{rw_ttt3r}, LoGeR
combines parametric memory with sliding-window context~\citep{rw_loger},
and LingBot-Map combines an anchor context, a pose-reference window, and a
trajectory memory~\citep{chen2026lingbotmap}. InfiniteVGGT uses a bounded rolling
cache~\citep{rw_infinitevggt}. LongStream combines keyframe-relative pose
prediction with scale modeling and cache management~\citep{rw_longstream}, whereas
ABot-Recon instead emphasizes local temporal context and adjacent-frame
relative predictions~\citep{rw_abotrecon}. Test-time training offers
another memory mechanism: VGG-T$^3$ distills scene information into a
fixed-size network for scalable offline reconstruction~\citep{rw_vggt3},
while Scal3R adapts a neural global-context representation at test time for
long sequences~\citep{rw_scal3r}. VGG-T$^3$ is related to streaming methods through
its memory formulation, but does not require strictly causal inference.

Video reconstruction must also account for scene dynamics. MonST3R adapts
point-map prediction to recover geometry at individual
timesteps~\citep{rw_monst3r}, while PAGE-4D separates the treatment of
dynamic information for camera estimation and geometry
reconstruction~\citep{rw_page4d}, and VGGT4D mines motion cues from a
pretrained geometry transformer's attention to suppress dynamic interference
in camera and geometry estimation without additional
training~\citep{rw_vggt4d}. These methods predict cameras and geometry
rather than scene motion. They address motion-induced
inconsistency, complementing the memory and sequence-length concerns of
streaming reconstruction.

\paragraph{\textbf{Panoramic reconstruction.}}
Panoramas offer wide spatial coverage, but spherical distortion and
projection discontinuities limit direct transfer from perspective-image
models. PanoSplatt3R adapts perspective pretraining to unposed, wide-baseline
panoramas~\citep{ren2025panosplatt3r}. PanoVGGT instead incorporates
modeling that accounts for spherical geometry to jointly predict cameras and dense geometry
from panoramic inputs~\citep{rw_panovggt}. PaGeR adapts a perspective-pretrained
geometry transformer using mixed perspective and panoramic training, jointly
estimating depth, surface normals, and sky masks~\citep{rw_pager}.
Data coverage is also important:
Holo360D provides real-world continuous panoramic trajectories with aligned
geometry and camera poses, supporting both training and evaluation
\citep{rw_holo360d}.

Argus targets metric indoor reconstruction, using learned co-visibility
to choose a reference view and geometric constraints to couple prediction
branches~\citep{rw_argus}. CasaMaestro reconstructs house-scale geometry
and camera poses from sparse multi-view panoramas~\citep{rw_casamaestro}.
Together, these works extend panoramic reconstruction from adapting
perspective priors to collecting suitable supervision and maintaining
metric consistency across large indoor spaces.

\paragraph{\textbf{AnyCamera reconstruction.}}
General camera models extend reconstruction beyond a fixed pinhole
assumption. Wid3R conditions reconstruction on the camera
model~\citep{jung2026wid3r}. CAM3R predicts ray directions and scene geometry
for uncalibrated wide-angle images~\citep{rw_cam3r}, whereas X-Lens estimates metric
depth from calibrated, heterogeneous fisheye and pinhole views~\citep{rw_xlens}.
Fisheye3R adapts pretrained multi-view models to fisheye projection while
retaining perspective-image capability~\citep{rw_fisheye3r}. RayTun3R instead
adapts lightweight position- and camera-related components on a short segment
of the test sequence, leaving the pretrained backbone fixed~\citep{rw_raytun3r}.

Other approaches address camera heterogeneity through training data or
ray-based representations. MEOW learns mixed-camera reconstruction from
procedurally rendered training tuples without requiring camera-type labels
or calibration at inference~\citep{rw_meow}. G-ray parameterizes rotary
position encodings using camera-local ray angles~\citep{rw_gray}.
Any3R uses a supplied dense ray field to condition angular positional
encoding and solid-angle-weighted supervision, supporting heterogeneous
central cameras in one reconstruction~\citep{rw_any3r}. At the single-image
level, OmniPoint decouples ray direction from distance to predict metric
point clouds across pinhole, fisheye, and panoramic projections, with
optional geometric priors~\citep{rw_omnipoint}. These methods differ in
whether calibration is supplied or inferred, whether adaptation is required,
and whether they predict single-view geometry or jointly recover cameras
and structure from multiple views.

\subsection{4D Reconstruction}
\label{sec:related-4d}
Dynamic reconstruction must recover time-varying geometry while separating
scene motion from camera motion. Occlusion, reappearance, and long-range
association make independently predicted depth maps insufficient for
describing how scene points move. We therefore use 4D reconstruction for
methods that explicitly model scene flow, dynamic points, or point
trajectories over time. Methods that estimate only cameras and per-frame
geometry in dynamic scenes are discussed under 3D reconstruction.
Dynamic Point Maps extend point-map regression to spatial and temporal
references, supporting geometry, scene flow, and correspondence prediction
from image pairs~\citep{rw_dpm}. St4RTrack jointly reconstructs and tracks
dynamic content through paired point maps in a world coordinate
frame~\citep{rw_st4rtrack}. SpatialTrackerV2 (SpaTrackerV2) combines geometry, camera motion, and point motion in a
feed-forward tracker~\citep{rw_spatialtrackerv2}. Any4D jointly predicts
metric geometry and motion across multiple frames~\citep{karhade2025any4d},
while V-DPM extends dynamic point maps to video to represent scene structure
and per-point 3D motion~\citep{rw_vdpm}. Flow4R uses relative scene flow to
connect geometry and motion in two-view prediction~\citep{rw_flow4r}.
Track4World uses a shared video representation and 3D correlation to estimate
dense flow between selected frame pairs and recover world-centric
trajectories~\citep{rw_track4world}.

Trajectory representations make temporal continuity explicit. Shape of
Motion fits shared motion bases per scene using depth and long-range 2D
tracks as priors~\citep{rw_shapeofmotion}.
Trace Anything predicts continuous per-pixel 3D trajectories parameterized
by control points~\citep{rw_traceanything}. OmniX represents dense motion
with trajectory fields~\citep{rw_omnix}. SM4RT uses structured motion bases
to model coordinated point motion~\citep{rw_sm4rt}, while Point4D extends
query-based trajectory prediction to long video sequences~\citep{rw_point4d}.
MoVieS links geometry and motion to rendering through moving, pixel-aligned
Gaussian primitives, supporting dynamic view synthesis and point
tracking~\citep{rw_movies}.

Query-based models offer a complementary way to access dynamic geometry
without decoding every point at every time. D4RT exposes point-level
spatiotemporal queries~\citep{zhang2025d4rt}.
4RC conditionally decodes geometry and motion from an encoded
clip~\citep{luo2026fourrc}.
UniQuery4R selects the source view, target view, and a continuous
source-pixel coordinate at decoding time~\citep{chen2026uniquery4r}. Fast Spatial Tracking combines global and
frame-level query features to predict trajectories from geometry-grounded
video features~\citep{rw_fastspatial}. L4P shares a video encoder across
depth, optical flow, and point tracking, while retaining lightweight
task-specific heads~\citep{rw_l4p}.

ARROW broadens the input setting from a single ordered video to arbitrary
image sets, using order-invariant queries to associate observations across
viewpoints and capture times~\citep{fradlin2026arrow}. Across these methods,
the main distinctions concern whether motion is represented by pairwise
displacement, continuous trajectories, structured motion, or conditional
queries, and how each representation handles visibility and temporal extent.

\subsection{Matching}
\label{sec:related-matching}
Visual correspondences connect image observations to geometric constraints.
Dense matching emphasizes coverage, sparse matching limits computational
cost, and multi-view matching or tracking seeks consistency across more
than two observations.

\paragraph{\textbf{Dense matching.}}
RoMa combines frozen DINOv2~\citep{oquab2023dinov2} coarse features with fine convolutional
features and a transformer match decoder for robust dense
matching~\citep{edstedt2023roma}. RoMa v2 further studies architecture,
training distribution, and efficient refinement for difficult image
pairs~\citep{rw_romav2}. UFM unifies optical flow and wide-baseline matching
as dense displacement prediction~\citep{rw_ufm}. These methods provide
constraints across a broader image area than sparse keypoints, although
their use in reconstruction still requires handling visibility and
unreliable correspondences.

\paragraph{\textbf{Sparse matching.}}
Sparse pipelines separate the selection and description of keypoints from
their association across images. DISK learns local features through a
matching-oriented reinforcement-learning objective~\citep{rw_disk}. ALIKED
uses sparse deformable descriptors for efficient extraction~\citep{rw_aliked}, while
DaD studies descriptor-free keypoint learning and detector
distillation~\citep{rw_dad}. At the matching stage, LightGlue adapts
computation to the difficulty of an image pair~\citep{lindenberger2023lightglue},
and LoMa revisits local matching through model and data
scaling~\citep{nordstrom2026loma}. Detector quality, descriptor
discriminability, and assignment accuracy thus contribute separately to
the reliability and cost of sparse geometric constraints.

\paragraph{\textbf{Multi-view matching and tracking.}}
Reliable pairwise matches do not automatically form consistent multi-view
tracks. VGGSfM uses point tracks in a differentiable SfM
pipeline~\citep{rw_vggsfm}. CoTracker3 learns 2D tracking from real-video
pseudo-labels~\citep{rw_cotracker3}, while SpatialTrackerV2 jointly models scene
geometry, camera motion, and 3D point motion~\citep{rw_spatialtrackerv2}.
MV-RoMa jointly matches a source image to multiple targets and constructs
tracks for reconstruction~\citep{rw_mvroma}. MVLoMa instead combines
features from geometry foundation models with sparse center--neighbor matching on a
star graph~\citep{lu2026mvloma}. These approaches differ in whether they
associate dense pixels, sparse keypoints, or tracked points, while sharing
the aim of using information from multiple observations jointly.

Geometry-assisted trackers explicitly use external geometric estimates.
The earlier SpatialTracker, the predecessor of SpatialTrackerV2, lifts
pixels to 3D with monocular depth and represents each
frame with a triplane~\citep{rw_spatialtracker}. DELTA combines global--local
attention with trajectory upsampling for dense 3D tracking~\citep{rw_delta}, whereas
DELTAv2 reduces its cost through coarse-to-fine trajectory expansion and
more efficient correlation computation~\citep{rw_deltav2}. TAPIP3D uses
depth and camera motion to construct stabilized 3D feature clouds, in which
local attention refines point trajectories~\citep{rw_tapip3d}.

With synchronized, calibrated cameras, MVTracker fuses multi-view features
and depth into a common 3D cloud~\citep{rw_mvtracker}. MV-TAP uses camera
geometry and cross-view attention to track 2D points across video
streams~\citep{rw_mvtap}, whereas LAPA combines appearance and geometric
constraints through attention to recover multi-camera 3D
trajectories~\citep{rw_lapa}. These methods exploit complementary views to reduce ambiguity and occlusion. They assume supplied camera calibration, whereas joint reconstruction and tracking from unposed RGB also require estimating camera parameters.

\subsection{Pose Estimation Systems}
\label{sec:related-pose}
Pose estimation systems recover cameras in a shared coordinate frame and
usually estimate scene structure alongside them. These systems fall into
three families according to whether they use no learned geometric or pose priors, geometric priors only, or both geometric and camera-pose priors.

\paragraph{\textbf{Traditional SfM.}}
Classical SfM combines visual correspondences with multi-view geometry to
estimate cameras and sparse structure. COLMAP provides an incremental
reconstruction pipeline~\citep{schoenberger2016sfm}, while GLOMAP revisits
global SfM: after rotation averaging, it jointly estimates camera positions
and 3D points in a single global positioning step instead of separate
translation averaging and triangulation, bringing global SfM close to
incremental accuracy and robustness~\citep{pan2024glomap}.
Their explicit geometric constraints remain valuable for scene-level
consistency, although reliable correspondences and adequate observation
geometry are important for successful reconstruction.

\paragraph{\textbf{Systems with geometric priors.}}
This family retains geometric estimation while introducing learned depth
or surface information. MP-SfM combines monocular depth and normal priors
with multi-view constraints~\citep{rw_mpsfm}. MegaSaM targets camera and
depth recovery from casual dynamic videos~\citep{li2024megasam}, while ViPE
estimates intrinsics, camera motion, and dense depth across diverse video
and camera settings~\citep{rw_vipe}. VidMap exploits video temporal
structure and depth priors~\citep{pataki2026vidmap}. Marginalized bundle
adjustment mitigates the high error variance of dense monocular depth when
estimating multi-view cameras~\citep{rw_mba}. Their common motivation is to strengthen
geometric evidence when appearance-based constraints alone are unreliable,
rather than replace scene-level estimation with independent predictions.

\paragraph{\textbf{Systems with geometric and pose priors.}}
Foundation models can also provide initial cameras together with local
geometry. MASt3R-SfM integrates model-derived reconstruction and
matching~\citep{duisterhof2024mast3rsfm}. VGGT-SLAM and VGGT-Long combine
local predictions across longer sequences~\citep{rw_vggtslam,rw_vggtlong}.
VGGT-SLAM++ adds DEM-based submap association and frequent local bundle
adjustment for large-scale mapping with bounded memory~\citep{rw_vggtslampp}.
GLUEMAP runs feed-forward inference on local star graphs filtered by
Doppelgangers++~\citep{xiangli2025doppelgangers} and fuses multiple track types in bundle adjustment, gaining
robustness to symmetry, appearance change, and low
overlap~\citep{pan2026gluemap}. Glob3R also integrates feed-forward
predictions into global reconstruction~\citep{rw_glob3r}, while GeoWeaver studies
hierarchical assembly and refinement of geometric
priors~\citep{rw_geoweaver}. HiP3R first aligns local reconstructions with a
chunk-level $\mathrm{Sim}(3)$ pose graph, then refines a shared
$\mathrm{SE}(3)$ camera for each frame under the fixed global layout. This
hierarchy reconciles overlapping predictions without retraining the
underlying reconstruction model~\citep{rw_hip3r}.
These systems combine learned initialization
with cross-view constraints to improve consistency beyond individual
prediction windows, with different choices for organizing observations
and controlling optimization cost.

%% file: sections/scene-scale.tex
\subsubsection{Scene Scale}
\label{sec:scene-scale}
We distinguish three scale conventions: a global geometry scale for RGB-only supervision, a camera-pose scale for camera-conditioned prediction, and an internal scene scale for the 3DGS decoder.

\paragraph{\textbf{Global geometry scale.}}
Scene extent and observation distance vary substantially across scenes. For the RGB-only model, we define a global scale as the mean distance of all valid 3D points across the input views to the first camera center. Depth, 3D points, and camera translations are divided by this shared scale. This preserves relative geometry within each scene while bringing supervision across scenes into comparable numerical ranges.

\paragraph{\textbf{Camera-pose scale.}}
For camera-conditioned models, predicted geometry should follow the input camera poses, and predictions from different chunks should be mapped back to a common scale. We therefore define the pose scale using the maximum distance from an input camera center to the first camera center:
\begin{equation}
 s_{\mathrm{pose}}=\max_i\|C_i-C_1\|_2.
 \label{eq:scene-pose-scale}
\end{equation}
Here $C_i$ denotes an input camera center. The implementation applies a lower-bound safeguard to very small normalization factors. With this scale as a shared reference, we trained two models, each using one of the following schemes.

\noindent\textit{Scheme 1: direct pose-scale normalization.}
Camera translations, depth, and 3D points are divided by $s_{\mathrm{pose}}$, and the model directly predicts geometry in pose-normalized units. Each chunk may use a different normalization factor. As long as the input poses share the same coordinate units, reversing each chunk's normalization and reference-frame transform restores its geometry to the common input-pose coordinate system. This provides a consistent scale reference for cross-chunk reconstruction.

\noindent\textit{Scheme 2: decoupled pose and per-frame geometry scales.}
When camera motion is small relative to scene distance, dividing depth directly by the pose scale can produce large values, widening the prediction range and making optimization harder. We retain the pose scale as the shared reference, but normalize each frame's geometry by its own scale $s_i$, the mean distance of valid local 3D points to that frame's camera center. A separate scale branch predicts the ratio
\begin{equation}
 r_i=\frac{s_i}{s_{\mathrm{pose}}}.
 \label{eq:scene-scale-ratio}
\end{equation}
Depth is supervised in units normalized by the per-frame geometry scale, camera translation remains normalized by the shared pose scale, and the scale ratio receives separate supervision. At inference, the predicted ratio converts normalized per-frame depth back to pose-normalized units:
\begin{equation}
 \hat d_i^{\mathrm{pose}}=\hat{\tilde d}_i\,\hat r_i.
 \label{eq:scene-depth-restoration}
\end{equation}
This keeps depth targets in a more suitable numerical range for learning while retaining the means to align them with the input poses and other chunks. Per-frame depth regression is thus decoupled from the magnitude of camera motion without losing the shared pose-scale reference.

\paragraph{\textbf{3DGS scale.}}
As an experimental extension, we build a 3DGS decoder. When training this feed-forward 3DGS model, we recompute the scene scale from the predicted geometry and apply a shared normalization. This gives 3DGS parameters a more consistent relative spatial meaning and keeps properties such as Gaussian positions and spatial sizes in more stable numerical ranges.

Specifically, we back-project predicted depths into the shared reference frame, compute the 95th percentile of the distances of valid 3D points to the origin, and scale that radius to a fixed reference value of 2.0:
\begin{equation}
 a_{\mathrm{GS}}=
 \frac{2.0}{Q_{0.95}\!\left(\{\|\hat X_{i,p}^{\mathrm{pose}}\|_2\}_{\mathrm{valid}}\right)}.
 \label{eq:gaussian-scene-scale}
\end{equation}
Here $\hat X_{i,p}^{\mathrm{pose}}$ is the back-projected point for pixel $p$ in view $i$, expressed in the shared pose-normalized reference frame. One factor $a_{\mathrm{GS}}$ uniformly scales geometry and camera translation across all input views, so the 3DGS decoder operates at a consistent relative scene scale.

%% file: sections/kosmoworld-pipeline.tex
\subsubsection{Data Generation Pipeline}
\label{sec:kosmoworld-pipeline}
KosmoWorld builds on WorldNav, the trajectory-planning module introduced in Tencent Hunyuan's HY-World 2.0~\citep{hyworld2026}. Our data-generation pipeline applies this planning framework to reconstructed real-world scenes, combining camera-location screening, trajectory planning, and RGB and depth rendering under configurable camera models. Reconstruction quality varies spatially, so the pipeline combines geometric feasibility with observation quality before generating data.

\paragraph{\textbf{Geometric feasibility and location selection.}}
Our pipeline first aligns the scene mesh, 3D Gaussians, and capture poses in a shared coordinate frame. A heightfield and navigation mesh built with Recast~\citep{recastnavigation2026} identify candidate regions using slope, step height, vertical clearance, and connectivity. When capture poses are available, probing starts near the captured area and expands to unevaluated locations. An optional constraint limits expansion to the local convex hull of capture positions at compatible heights.

\paragraph{\textbf{Panoramic observation confidence.}}
At each probe position, six cube faces cover the viewing sphere and provide RGB, alpha, and depth. Sharpness and local image structure characterize appearance quality, alpha measures rendering coverage, and the fraction of pixels with valid depth measures geometric completeness. Local checks examine the spatial distribution of blur and missing depth to distinguish low-texture surfaces from reconstruction defects. A clarity score combines normalized sharpness $s_n$, Laplacian variance $l_n$, mean alpha $a_n$, and depth completeness $h_n$:
\begin{equation}
 q_{\mathrm{clarity}}=0.4s_n+0.2l_n+0.2a_n+0.2h_n.
 \label{eq:worldnav-clarity}
\end{equation}
When both mesh quality and 3DGS rendering quality are evaluated, the joint score is $q_{\mathrm{joint}}=\min(q_{\mathrm{mesh}},q_{\mathrm{GS}})$. This prevents a strong result in one representation from compensating for a weak result in the other. Under the strict panoramic policy, every required direction must be sampled and pass the quality checks. These scores assess observation quality rather than calibrated probabilities. The usable region contains locations passing both geometric and panoramic checks, with capture support enforced when enabled. Unevaluated locations are excluded from planning.

\paragraph{\textbf{Region partitioning and route selection.}}
The trajectory planner connects usable locations by reachability and plans disconnected components independently. In the mode that prioritizes coverage, a distance transform of free space identifies cores with high clearance, and graph-geodesic assignment partitions the remaining nodes along boundaries such as walls and passages. The planner selects representative targets, adds targets where coverage is sparse, and connects them through graph search. When quality weighting is enabled, an edge cost balances travel distance, clearance, and observation quality:
\begin{equation}
 w_{ij}=d_{ij}\,\phi_{\mathrm{clearance}}(i,j)
 \left[1+\lambda_q\left(1-\frac{q_i+q_j}{2}\right)\right].
 \label{eq:worldnav-route}
\end{equation}
Here $d_{ij}$ is the distance between adjacent nodes, $\phi_{\mathrm{clearance}}$ accounts for boundaries and narrow passages, $q_i$ and $q_j$ are node-quality scores, and $\lambda_q$ controls the quality weight. All route nodes remain in the usable region. Multiple routes are selected to extend spatial coverage while reducing repeated edges and overlapping segments.

\paragraph{\textbf{Trajectory diversity and validation.}}
Four motion profiles provide complementary viewpoint changes. Regular traversal connects the origin to geodesically distant targets in different angular sectors, producing continuous translational parallax. Wandering connects successive goals while favoring unvisited locations and uncovered directions. Landmark orbit connects reachable positions around a supplied 3D target while keeping the camera directed toward it, producing a full orbit or partial arc as space permits. Elevated and variable-height trajectories introduce vertical offsets or connect validated positions at different heights, with geometry and panoramic quality checked again at the new positions.

B-spline smoothing adjusts camera positions, and interpolation along the shortest quaternion arc smooths orientations subject to turn and angular-speed limits. Because smoothing changes the visited positions, the pipeline rechecks geometric sweeps and region constraints along each segment and reassesses panoramic quality at the final sample positions. Failed samples split a route into valid continuous segments. Rendering uses only the trajectories that pass this final validation.

\paragraph{\textbf{Camera configuration and rendering.}}
Camera projection is configured separately from trajectory motion, allowing the same scene to produce different viewpoints, resolutions, and camera models. Pinhole cameras use output resolution, focal lengths, and principal point, with intrinsics optionally derived from physical focal length and sensor aperture. Fisheye cameras use calibrated intrinsics, OpenCV-fisheye / KB4 distortion coefficients, and an effective angular range. For incident angle $\theta$, the distorted angle used in radial projection is
\begin{equation}
 \theta_d=\theta\left(1+k_1\theta^2+k_2\theta^4+k_3\theta^6+k_4\theta^8\right).
 \label{eq:worldnav-fisheye}
\end{equation}
The coefficients $k_1$--$k_4$, calibration range, image canvas, and rendering backend constrain the effective field of view. Equirectangular cameras use cube-face reprojection to produce a $360^\circ\times180^\circ$ spherical view at the chosen resolution and orientation. Intrinsics are transformed consistently when output resolution changes. Camera height, motion speed, and sample spacing determine positions and timestamps along the trajectory.

%% file: tables/ablation-full.tex
\begin{table}[!htbp]
\centering
\caption{KosmoWorld data ablation on SpatialBench. For each of the Sparse and Medium subsets, we compare models trained with and without KosmoWorld data. Accuracies are fractions, and lower errors are better. Bold indicates the better value within each subset.}
\label{tab:ablation-full}
\fontsize{8}{10}\selectfont
\setlength{\tabcolsep}{2pt}\renewcommand{\arraystretch}{1.18}
\begin{tabular}{@{}l*{6}{r}@{\hspace{6pt}}*{7}{r}@{}}
\toprule
 & \multicolumn{6}{c}{Depth} & \multicolumn{7}{c}{Camera pose} \\
\cmidrule(lr){2-7}\cmidrule(lr){8-14}
\multirow{2}{*}{Setting / data} & AbsRel & SqRel & RMSE & $\delta_{1.03}$ & $\delta_{1.05}$ & $\delta_{1.10}$ & RAcc3 & RAcc5 & TAcc3 & TAcc5 & AUC@5 & AUC@15 & AUC@30 \\
 & $\downarrow$ & $\downarrow$ & $\downarrow$ & $\uparrow$ & $\uparrow$ & $\uparrow$ & $\uparrow$ & $\uparrow$ & $\uparrow$ & $\uparrow$ & $\uparrow$ & $\uparrow$ & $\uparrow$ \\
\midrule
Sparse / Without & 0.0854 & 0.1209 & 0.6328 & 0.6396 & 0.7515 & 0.8642 & 0.8169 & 0.8872 & 0.6294 & 0.7121 & 0.5422 & 0.7197 & 0.8104 \\
Sparse / With & \textbf{0.0630} & \textbf{0.1093} & \textbf{0.6207} & \textbf{0.6696} & \textbf{0.7810} & \textbf{0.8816} & \textbf{0.8358} & \textbf{0.9055} & \textbf{0.6457} & \textbf{0.7394} & \textbf{0.5726} & \textbf{0.7412} & \textbf{0.8226} \\
\addlinespace[3pt]
Medium / Without & 0.0740 & 0.0956 & 0.5939 & 0.6417 & 0.7689 & 0.8815 & 0.7907 & 0.8672 & 0.6070 & 0.6984 & 0.5374 & 0.7107 & 0.8021 \\
Medium / With & \textbf{0.0594} & \textbf{0.0884} & \textbf{0.5822} & \textbf{0.6895} & \textbf{0.8077} & \textbf{0.9025} & \textbf{0.8310} & \textbf{0.8938} & \textbf{0.6468} & \textbf{0.7322} & \textbf{0.5820} & \textbf{0.7450} & \textbf{0.8281} \\
\bottomrule
\end{tabular}
\end{table}

%% file: tables/sparse-reconstruction.tex
\begin{table}[!htbp]
\centering
\caption{SpatialBench Sparse full results. Baseline metrics are taken from the official SpatialBench release. Depth and camera-pose metrics are grouped by column. Accuracy values are fractions, and parameter counts describe model size. Best, second-best, and third-best values are \textcolor{red!75!black}{red}, \textcolor{blue!75!black}{blue}, and \textcolor{green!45!black}{green}, with ties sharing a rank. Ranks use unrounded values. Baseline reference for $\pi^3$-X: \citealp{rw_pi3,pi3x2025implementation}.}
\label{tab:sparse-reconstruction}
\fontsize{8}{10}\selectfont
\setlength{\tabcolsep}{2pt}\renewcommand{\arraystretch}{1.18}
\begin{tabular}{@{}lr*{6}{r}@{\hspace{6pt}}*{7}{r}@{}}
\toprule
 & & \multicolumn{6}{c}{Depth} & \multicolumn{7}{c}{Camera pose} \\
\cmidrule(lr){3-8}\cmidrule(lr){9-15}
\multirow{2}{*}{Method} & Params & AbsRel & SqRel & RMSE & $\delta_{1.03}$ & $\delta_{1.05}$ & $\delta_{1.10}$ & RAcc3 & RAcc5 & TAcc3 & TAcc5 & AUC@5 & AUC@15 & AUC@30 \\
 & (M) & $\downarrow$ & $\downarrow$ & $\downarrow$ & $\uparrow$ & $\uparrow$ & $\uparrow$ & $\uparrow$ & $\uparrow$ & $\uparrow$ & $\uparrow$ & $\uparrow$ & $\uparrow$ & $\uparrow$ \\
\midrule
VGGT & 1256.54 & 0.105 & 0.105 & 0.673 & 0.498 & 0.627 & 0.778 & 0.734 & 0.833 & 0.454 & 0.550 & 0.405 & 0.585 & 0.700 \\
Fast3R & 647.55 & 0.260 & 0.760 & 1.576 & 0.228 & 0.340 & 0.508 & 0.347 & 0.486 & 0.156 & 0.235 & 0.109 & 0.264 & 0.392 \\
FastVGGT & 1157.94 & 0.113 & \textcolor{green!45!black}{0.098} & 0.666 & 0.457 & 0.590 & 0.752 & 0.642 & 0.773 & 0.384 & 0.472 & 0.314 & 0.503 & 0.631 \\
MUSt3R & 423.43 & 0.165 & 0.276 & 0.999 & 0.366 & 0.502 & 0.686 & 0.622 & 0.726 & 0.343 & 0.435 & 0.273 & 0.482 & 0.614 \\
MapAnything & 1228.49 & 0.153 & 1.079 & 1.337 & 0.361 & 0.512 & 0.701 & 0.608 & 0.762 & 0.300 & 0.423 & 0.244 & 0.446 & 0.579 \\
OmniVGGT & 1217.49 & 0.117 & 0.119 & 0.669 & 0.534 & 0.671 & 0.810 & 0.620 & 0.746 & 0.416 & 0.507 & 0.332 & 0.537 & 0.665 \\
$\pi^3$ & 958.70 & 0.092 & 0.258 & 0.855 & 0.547 & 0.686 & 0.826 & 0.769 & 0.854 & 0.474 & 0.586 & 0.412 & 0.619 & 0.742 \\
$\pi^3$-X & 1360.03 & \textcolor{green!45!black}{0.084} & \textcolor{blue!75!black}{0.084} & \textcolor{red!75!black}{\textbf{0.599}} & \textcolor{blue!75!black}{0.576} & \textcolor{blue!75!black}{0.710} & \textcolor{green!45!black}{0.833} & 0.756 & 0.837 & 0.491 & 0.595 & 0.427 & 0.627 & 0.741 \\
AMB3R & 1563.12 & 0.088 & \textcolor{red!75!black}{\textbf{0.083}} & \textcolor{green!45!black}{0.617} & 0.534 & 0.668 & 0.812 & 0.743 & 0.844 & 0.475 & 0.591 & 0.412 & 0.619 & 0.739 \\
DA3-Small & 34.30 & 0.191 & 0.270 & 1.057 & 0.308 & 0.440 & 0.633 & 0.392 & 0.568 & 0.172 & 0.300 & 0.127 & 0.336 & 0.476 \\
DA3-Base & 135.37 & 0.159 & 0.210 & 0.937 & 0.386 & 0.520 & 0.690 & 0.528 & 0.678 & 0.292 & 0.399 & 0.222 & 0.427 & 0.566 \\
DA3-Large & 410.94 & 0.128 & 0.198 & 0.839 & 0.470 & 0.601 & 0.753 & 0.662 & 0.776 & 0.443 & 0.528 & 0.361 & 0.559 & 0.688 \\
DA3-Giant & 1355.67 & 0.095 & 0.107 & \textcolor{blue!75!black}{0.608} & 0.563 & 0.689 & 0.821 & \textcolor{green!45!black}{0.792} & \textcolor{green!45!black}{0.871} & \textcolor{blue!75!black}{0.586} & \textcolor{blue!75!black}{0.682} & \textcolor{blue!75!black}{0.525} & \textcolor{blue!75!black}{0.699} & \textcolor{green!45!black}{0.785} \\
DA3-Nested & 1689.85 & 0.106 & 0.199 & 0.805 & \textcolor{green!45!black}{0.564} & 0.692 & 0.814 & 0.784 & 0.855 & 0.577 & \textcolor{blue!75!black}{0.682} & \textcolor{green!45!black}{0.519} & 0.691 & 0.779 \\
WorldMirror & 1263.34 & 0.139 & 0.162 & 0.798 & 0.440 & 0.583 & 0.748 & 0.697 & 0.793 & 0.394 & 0.506 & 0.320 & 0.532 & 0.660 \\
VGGT-$\Omega$ & 1143.81 & \textcolor{blue!75!black}{0.077} & 0.362 & 1.089 & 0.556 & \textcolor{green!45!black}{0.702} & \textcolor{blue!75!black}{0.843} & \textcolor{blue!75!black}{0.808} & \textcolor{blue!75!black}{0.892} & \textcolor{green!45!black}{0.585} & \textcolor{green!45!black}{0.678} & 0.504 & \textcolor{green!45!black}{0.695} & \textcolor{blue!75!black}{0.803} \\
\midrule
\textbf{Kosmo-GFM} & 1655.29 & \textcolor{red!75!black}{\textbf{0.063}} & 0.109 & 0.621 & \textcolor{red!75!black}{\textbf{0.670}} & \textcolor{red!75!black}{\textbf{0.781}} & \textcolor{red!75!black}{\textbf{0.882}} & \textcolor{red!75!black}{\textbf{0.836}} & \textcolor{red!75!black}{\textbf{0.906}} & \textcolor{red!75!black}{\textbf{0.646}} & \textcolor{red!75!black}{\textbf{0.739}} & \textcolor{red!75!black}{\textbf{0.573}} & \textcolor{red!75!black}{\textbf{0.741}} & \textcolor{red!75!black}{\textbf{0.823}} \\
\bottomrule
\end{tabular}
\end{table}

%% file: tables/medium-reconstruction.tex
\begin{table}[!htbp]
\centering
\caption{SpatialBench Medium full results. Baseline metrics are taken from the official SpatialBench release. Depth, camera-pose, and trajectory metrics are grouped by column. ATE and RPE follow the benchmark alignment and units. Accuracy values are fractions, and parameter counts describe model size. Best, second-best, and third-best values are \textcolor{red!75!black}{red}, \textcolor{blue!75!black}{blue}, and \textcolor{green!45!black}{green}, with ties sharing a rank. Ranks use unrounded values.}
\label{tab:medium-reconstruction}
\fontsize{8}{10}\selectfont
\setlength{\tabcolsep}{2pt}\renewcommand{\arraystretch}{1.18}
\begin{tabular}{@{}lr*{6}{r}@{\hspace{6pt}}*{7}{r}@{\hspace{6pt}}*{3}{r}@{}}
\toprule
 & & \multicolumn{6}{c}{Depth} & \multicolumn{7}{c}{Camera pose} & \multicolumn{3}{c}{Trajectory} \\
\cmidrule(lr){3-8}\cmidrule(lr){9-15}\cmidrule(lr){16-18}
\multirow{2}{*}{Method} & Params & AbsRel & SqRel & RMSE & $\delta_{1.03}$ & $\delta_{1.05}$ & $\delta_{1.10}$ & RAcc3 & RAcc5 & TAcc3 & TAcc5 & AUC@5 & AUC@15 & AUC@30 & ATE & RPE$_t$ & RPE$_r$ \\
 & (M) & $\downarrow$ & $\downarrow$ & $\downarrow$ & $\uparrow$ & $\uparrow$ & $\uparrow$ & $\uparrow$ & $\uparrow$ & $\uparrow$ & $\uparrow$ & $\uparrow$ & $\uparrow$ & $\uparrow$ & $\downarrow$ & $\downarrow$ & $\downarrow$ \\
\midrule
VGGT & 1256.54 & 0.125 & 0.602 & 0.688 & 0.499 & 0.613 & 0.747 & 0.695 & 0.782 & 0.484 & 0.562 & 0.432 & 0.586 & 0.687 & 0.727 & 0.216 & 2.202 \\
Fast3R & 647.55 & 0.255 & 0.567 & 1.539 & 0.241 & 0.344 & 0.496 & 0.296 & 0.422 & 0.164 & 0.248 & 0.117 & 0.272 & 0.386 & 6.582 & 2.055 & 13.650 \\
FastVGGT & 1157.94 & 0.105 & \textcolor{green!45!black}{0.086} & 0.611 & 0.479 & 0.597 & 0.737 & 0.678 & 0.771 & 0.448 & 0.542 & 0.379 & 0.554 & 0.662 & 0.738 & 0.259 & 3.260 \\
MUSt3R & 423.43 & 0.162 & 0.327 & 0.966 & 0.388 & 0.522 & 0.696 & 0.618 & 0.734 & 0.361 & 0.475 & 0.296 & 0.507 & 0.643 & 3.097 & 0.613 & 2.686 \\
MapAnything & 1228.49 & 0.146 & 0.490 & 1.052 & 0.347 & 0.491 & 0.681 & 0.563 & 0.702 & 0.312 & 0.419 & 0.254 & 0.451 & 0.579 & 1.737 & 0.533 & 2.852 \\
OmniVGGT & 1217.49 & 0.111 & 0.096 & 0.649 & 0.518 & 0.645 & 0.780 & 0.609 & 0.726 & 0.426 & 0.527 & 0.340 & 0.542 & 0.665 & 1.491 & 0.355 & 2.768 \\
$\pi^3$ & 958.70 & 0.082 & 0.229 & 0.822 & 0.563 & 0.689 & 0.814 & 0.742 & 0.830 & 0.501 & 0.613 & 0.433 & 0.636 & 0.749 & \textcolor{green!45!black}{0.565} & \textcolor{blue!75!black}{0.128} & \textcolor{blue!75!black}{1.240} \\
$\pi^3$-X & 1360.03 & \textcolor{green!45!black}{0.078} & \textcolor{red!75!black}{\textbf{0.061}} & \textcolor{red!75!black}{\textbf{0.538}} & \textcolor{green!45!black}{0.582} & \textcolor{green!45!black}{0.712} & \textcolor{green!45!black}{0.831} & 0.741 & 0.827 & 0.536 & 0.628 & 0.463 & 0.644 & 0.744 & \textcolor{red!75!black}{\textbf{0.369}} & \textcolor{red!75!black}{\textbf{0.108}} & 1.459 \\
AMB3R & 1563.12 & 0.085 & \textcolor{blue!75!black}{0.074} & \textcolor{green!45!black}{0.580} & 0.539 & 0.661 & 0.795 & 0.704 & 0.799 & 0.496 & 0.588 & 0.429 & 0.613 & 0.727 & 0.645 & 0.223 & 1.881 \\
DA3-Small & 34.30 & 0.176 & 0.226 & 1.008 & 0.312 & 0.435 & 0.617 & 0.370 & 0.535 & 0.192 & 0.301 & 0.136 & 0.338 & 0.479 & 4.850 & 1.356 & 5.534 \\
DA3-Base & 135.37 & 0.142 & 0.171 & 0.884 & 0.395 & 0.521 & 0.683 & 0.494 & 0.626 & 0.289 & 0.407 & 0.227 & 0.430 & 0.562 & 3.865 & 0.976 & 3.928 \\
DA3-Large & 410.94 & 0.105 & 0.155 & 0.779 & 0.498 & 0.618 & 0.767 & 0.655 & 0.758 & 0.487 & 0.575 & 0.411 & 0.588 & 0.701 & 2.722 & 0.460 & 2.800 \\
DA3-Giant & 1355.67 & 0.086 & 0.088 & \textcolor{blue!75!black}{0.578} & 0.572 & 0.686 & 0.812 & \textcolor{green!45!black}{0.749} & \textcolor{green!45!black}{0.834} & \textcolor{green!45!black}{0.587} & \textcolor{green!45!black}{0.663} & \textcolor{blue!75!black}{0.532} & \textcolor{green!45!black}{0.684} & \textcolor{green!45!black}{0.776} & 1.161 & 0.285 & 2.253 \\
DA3-Nested & 1689.85 & 0.086 & 0.182 & 0.801 & 0.577 & 0.690 & 0.810 & 0.742 & 0.833 & 0.569 & 0.658 & 0.510 & 0.676 & 0.770 & 1.980 & 0.394 & 2.708 \\
WorldMirror & 1263.34 & 0.118 & 0.126 & 0.741 & 0.445 & 0.573 & 0.721 & 0.674 & 0.775 & 0.417 & 0.533 & 0.352 & 0.556 & 0.674 & 1.357 & 0.347 & 2.056 \\
VGGT-$\Omega$ & 1143.81 & \textcolor{blue!75!black}{0.067} & 0.304 & 1.049 & \textcolor{blue!75!black}{0.589} & \textcolor{blue!75!black}{0.721} & \textcolor{blue!75!black}{0.845} & \textcolor{blue!75!black}{0.787} & \textcolor{blue!75!black}{0.870} & \textcolor{blue!75!black}{0.589} & \textcolor{blue!75!black}{0.692} & \textcolor{green!45!black}{0.512} & \textcolor{blue!75!black}{0.700} & \textcolor{blue!75!black}{0.795} & 0.659 & 0.160 & \textcolor{green!45!black}{1.369} \\
\midrule
\textbf{Kosmo-GFM} & 1655.29 & \textcolor{red!75!black}{\textbf{0.059}} & 0.088 & 0.582 & \textcolor{red!75!black}{\textbf{0.690}} & \textcolor{red!75!black}{\textbf{0.808}} & \textcolor{red!75!black}{\textbf{0.903}} & \textcolor{red!75!black}{\textbf{0.831}} & \textcolor{red!75!black}{\textbf{0.894}} & \textcolor{red!75!black}{\textbf{0.647}} & \textcolor{red!75!black}{\textbf{0.732}} & \textcolor{red!75!black}{\textbf{0.582}} & \textcolor{red!75!black}{\textbf{0.745}} & \textcolor{red!75!black}{\textbf{0.828}} & \textcolor{blue!75!black}{0.508} & \textcolor{green!45!black}{0.145} & \textcolor{red!75!black}{\textbf{1.088}} \\
\bottomrule
\end{tabular}
\end{table}

%% file: tables/prompt.tex
\begin{table}[!htbp]
\centering
\caption{Kosmo-GFM on SpatialBench Sparse: RGB-only GFM and Camera-conditioned GFM, with and without per-frame scale prediction. Columns and precision follow Table~\ref{tab:sparse-reconstruction}. Camera inputs provide additional information, so the pose scores are not directly comparable to uncalibrated pose estimation.}
\label{tab:prompt}
\fontsize{8}{10}\selectfont
\setlength{\tabcolsep}{1.5pt}\renewcommand{\arraystretch}{1.18}
\begin{tabular}{@{}lr*{6}{r}@{\hspace{6pt}}*{7}{r}@{}}
\toprule
 & & \multicolumn{6}{c}{Depth} & \multicolumn{7}{c}{Camera pose} \\
\cmidrule(lr){3-8}\cmidrule(lr){9-15}
\multirow{2}{*}{Model setting} & Params & AbsRel & SqRel & RMSE & $\delta_{1.03}$ & $\delta_{1.05}$ & $\delta_{1.10}$ & RAcc3 & RAcc5 & TAcc3 & TAcc5 & AUC@5 & AUC@15 & AUC@30 \\
 & (M) & $\downarrow$ & $\downarrow$ & $\downarrow$ & $\uparrow$ & $\uparrow$ & $\uparrow$ & $\uparrow$ & $\uparrow$ & $\uparrow$ & $\uparrow$ & $\uparrow$ & $\uparrow$ & $\uparrow$ \\
\midrule

RGB-only & 1655.29 & 0.063 & 0.109 & 0.621 & 0.670 & 0.781 & 0.882 & 0.836 & 0.906 & 0.646 & 0.739 & 0.573 & 0.741 & 0.823 \\
Camera-conditioned & 1769.82 & 0.082 & 0.167 & 0.757 & 0.577 & 0.707 & 0.842 & 1.000 & 1.000 & 0.997 & 0.998 & 0.996 & 0.998 & 0.999 \\
\shortstack[l]{Camera-conditioned\\+ per-frame scale} & 1788.74 & 0.067 & 0.131 & 0.689 & 0.630 & 0.753 & 0.870 & 1.000 & 1.000 & 0.996 & 0.998 & 0.995 & 0.998 & 0.999 \\
\bottomrule
\end{tabular}
\end{table}

%% file: tables/sparse-datasets.tex
\begin{table}[!htbp]
\centering
\caption{Sparse per-dataset breakdown. Each metric group is ordered DA3-Giant / VGGT-$\Omega$ / Kosmo-GFM. DA3-Giant and VGGT-$\Omega$ results are from our local evaluations, separate from the official SpatialBench results in Table~\ref{tab:sparse-reconstruction}.}
\label{tab:sparse-datasets}
\footnotesize
\setlength{\tabcolsep}{4pt}\renewcommand{\arraystretch}{1.12}
\begin{tabular}{p{0.34\textwidth}rrrrrr}
\toprule
Dataset & \multicolumn{3}{c}{AbsRel $\downarrow$} & \multicolumn{3}{c}{AUC@30 $\uparrow$} \\
\cmidrule(lr){2-4}\cmidrule(lr){5-7}
 & DA3-Giant & VGGT-$\Omega$ & \textbf{Kosmo-GFM} & DA3-Giant & VGGT-$\Omega$ & \textbf{Kosmo-GFM} \\
\midrule
7-Scenes~\citep{shotton2013sevenscenes} & 0.0656 & \textbf{0.0557} & 0.0637 & 0.7848 & \textbf{0.8067} & 0.7638 \\
ADT~\citep{pan2023adt} & 0.0627 & 0.0638 & \textbf{0.0408} & 0.7551 & 0.8506 & \textbf{0.9574} \\
DROID~\citep{khazatsky2024droid} & 0.1249 & 0.0907 & \textbf{0.0709} & 0.6510 & 0.6194 & \textbf{0.7046} \\
DTU~\citep{jensen2014dtu} & 0.0142 & 0.0135 & \textbf{0.0126} & 0.9828 & \textbf{0.9885} & 0.9836 \\
ETH3D~\citep{schoeps2017eth3d} & 0.0346 & \textbf{0.0231} & 0.0349 & 0.7858 & \textbf{0.8646} & 0.7992 \\
HiRoom~\citep{lin2025da3} & 0.0117 & 0.0134 & \textbf{0.0086} & 0.9167 & \textbf{0.9787} & 0.9731 \\
NRGBD~\citep{azinovic2022nrgbd} & \textbf{0.0092} & 0.0101 & 0.0099 & \textbf{0.9451} & 0.9353 & 0.9404 \\
OmniWorld~\citep{zhou2025omniworld} & 0.0505 & 0.0501 & \textbf{0.0473} & \textbf{0.9843} & 0.9696 & 0.9825 \\
RLBench~\citep{james2020rlbench,pumacay2024colosseum} & 0.2755 & 0.2055 & \textbf{0.1439} & 0.5757 & 0.5787 & \textbf{0.6507} \\
RoboLab~\citep{yang2026robolab} & 0.1712 & 0.1469 & \textbf{0.0996} & 0.7114 & 0.7356 & \textbf{0.8116} \\
RoboTwin~\citep{chen2025robotwin} & 0.3247 & 0.2202 & \textbf{0.1995} & 0.4429 & 0.5371 & \textbf{0.5483} \\
Ropedia~\citep{ropedia2026xperience} & 0.0494 & \textbf{0.0395} & 0.0451 & 0.2959 & \textbf{0.4797} & 0.3396 \\
ScanNet++~\citep{yeshwanth2023scannetpp} & 0.0316 & 0.0418 & \textbf{0.0259} & 0.9623 & 0.9463 & \textbf{0.9665} \\
Tanks \& Temples~\citep{knapitsch2017tanks} & 0.0405 & 0.0414 & \textbf{0.0273} & 0.6203 & 0.5954 & \textbf{0.6992} \\
TUM~\citep{sturm2012tum} & 0.0879 & \textbf{0.0419} & 0.0460 & 0.7689 & \textbf{0.8067} & 0.7928 \\
vKITTI~\citep{cabon2020vkitti2} & 0.0635 & 0.0996 & \textbf{0.0627} & 0.7833 & 0.7934 & \textbf{0.8320} \\
Waymo~\citep{sun2020waymo} & \textbf{0.0778} & 0.0881 & 0.0838 & \textbf{0.9832} & 0.9711 & 0.9746 \\
\midrule
\textbf{Average} & 0.0958 & 0.0777 & \textbf{0.0630} & 0.7831 & 0.8002 & \textbf{0.8226} \\
\bottomrule
\end{tabular}
\end{table}

%% file: tables/medium-datasets.tex
\begin{table}[!htbp]
\centering
\caption{Medium per-dataset breakdown. Each metric group is ordered DA3-Giant / VGGT-$\Omega$ / Kosmo-GFM. DA3-Giant and VGGT-$\Omega$ results are from our local evaluations, separate from the official SpatialBench results in Table~\ref{tab:medium-reconstruction}.}
\label{tab:medium-datasets}
\footnotesize
\setlength{\tabcolsep}{4pt}\renewcommand{\arraystretch}{1.12}
\begin{tabular}{lrrrrrr}
\toprule
Dataset & \multicolumn{3}{c}{AbsRel $\downarrow$} & \multicolumn{3}{c}{AUC@30 $\uparrow$} \\
\cmidrule(lr){2-4}\cmidrule(lr){5-7}
 & DA3-Giant & VGGT-$\Omega$ & \textbf{Kosmo-GFM} & DA3-Giant & VGGT-$\Omega$ & \textbf{Kosmo-GFM} \\
\midrule
7-Scenes & 0.0577 & \textbf{0.0491} & 0.0586 & 0.7878 & \textbf{0.8585} & 0.8130 \\
ADT & 0.0564 & 0.0423 & \textbf{0.0304} & 0.8473 & 0.9365 & \textbf{0.9746} \\
DROID & 0.0915 & 0.0584 & \textbf{0.0506} & 0.5952 & 0.5324 & \textbf{0.6064} \\
DTU & \textbf{0.0108} & \textbf{0.0108} & 0.0148 & \textbf{0.9844} & 0.9730 & 0.9602 \\
ETH3D & 0.0282 & \textbf{0.0209} & 0.0263 & 0.8723 & \textbf{0.9008} & 0.8857 \\
HiRoom & \textbf{0.0056} & 0.0089 & \textbf{0.0056} & \textbf{0.9974} & 0.9795 & 0.9957 \\
NRGBD & 0.0088 & \textbf{0.0069} & 0.0088 & \textbf{0.9711} & 0.9609 & 0.9695 \\
OmniWorld & 0.0463 & 0.0468 & \textbf{0.0395} & 0.9791 & 0.9604 & \textbf{0.9795} \\
RLBench & 0.2435 & 0.1821 & \textbf{0.1236} & 0.4719 & 0.5952 & \textbf{0.6489} \\
RoboLab & 0.1704 & 0.1486 & \textbf{0.0725} & 0.6424 & 0.8115 & \textbf{0.8953} \\
RoboTwin & 0.3124 & \textbf{0.1929} & 0.2810 & 0.2593 & 0.3845 & \textbf{0.4538} \\
Ropedia & \textbf{0.0568} & 0.0691 & 0.0824 & 0.2298 & 0.2687 & \textbf{0.3079} \\
ScanNet++ & 0.0309 & 0.0408 & \textbf{0.0260} & \textbf{0.9787} & 0.9697 & 0.9737 \\
Tanks \& Temples & 0.0341 & 0.0370 & \textbf{0.0246} & 0.7468 & 0.6365 & \textbf{0.7486} \\
TUM & 0.0795 & \textbf{0.0396} & 0.0448 & 0.8327 & \textbf{0.9016} & 0.8868 \\
vKITTI & 0.0612 & 0.0933 & \textbf{0.0504} & 0.7566 & 0.7875 & \textbf{0.9223} \\
Waymo & 0.0666 & 0.0731 & \textbf{0.0606} & \textbf{0.9761} & 0.9499 & 0.9748 \\
\midrule
\textbf{Average} & 0.0853 & 0.0672 & \textbf{0.0594} & 0.7720 & 0.7962 & \textbf{0.8281} \\
\bottomrule
\end{tabular}
\end{table}

%% file: tables/motion.tex
\begin{table}[!htbp]
\centering
\caption{Dynamic-point and scene-flow averages over Panoptic Studio, PointOdyssey, Dynamic Replica, Aria Digital Twin, and SynthVerse. APD and flow accuracy are percentages, and EPE is in meters. Best, second-best, and third-best values are \textcolor{red!75!black}{red}, \textcolor{blue!75!black}{blue}, and \textcolor{green!45!black}{green}, with ties sharing a rank. Means are recomputed from Tables~\ref{tab:motion-point-accuracy} and \ref{tab:motion-flow-accuracy}, with rounding applied only for display. Ranks use unrounded means.}
\label{tab:motion}
\small
\setlength{\tabcolsep}{4pt}\renewcommand{\arraystretch}{1.12}
\begin{tabular}{lrrrr}
\toprule
Method & Point APD $\uparrow$ & Point EPE $\downarrow$ & Flow acc. $\uparrow$ & Flow EPE $\downarrow$ \\
\midrule
UniQuery4R & \textcolor{green!45!black}{72.7} & \textcolor{green!45!black}{0.610} & \textcolor{green!45!black}{71.0} & \textcolor{green!45!black}{0.351} \\
SpaTrackerV2 & 48.8 & 1.443 & 15.5 & 1.075 \\
St4RTrack & 60.9 & 0.993 & 41.4 & 0.796 \\
TraceAnything & 53.1 & 1.896 & 54.8 & 0.642 \\
Any4D & 63.4 & 1.287 & 59.5 & 0.525 \\
V-DPM & 69.7 & 0.703 & 69.3 & 0.418 \\
4RC & 68.3 & 0.827 & 70.9 & 0.352 \\
OpenD4RT & 62.4 & 1.159 & 64.4 & 0.608 \\
ARROW & \textcolor{red!75!black}{\textbf{77.0}} & \textcolor{blue!75!black}{0.551} & \textcolor{red!75!black}{\textbf{75.1}} & \textcolor{blue!75!black}{0.291} \\
\midrule
\textbf{Kosmo-GFM} & \textcolor{blue!75!black}{76.4} & \textcolor{red!75!black}{\textbf{0.342}} & \textcolor{blue!75!black}{71.1} & \textcolor{red!75!black}{\textbf{0.188}} \\
\bottomrule
\end{tabular}
\end{table}

%% file: tables/motion-point-accuracy.tex
\begin{table}[!htbp]
\centering
\caption{Dynamic-point APD (percent) and EPE (meters). All ten methods are shown on all five datasets. Mean is recomputed as the arithmetic mean of the five displayed dataset values. Best, second-best, and third-best values are \textcolor{red!75!black}{red}, \textcolor{blue!75!black}{blue}, and \textcolor{green!45!black}{green}, with ties sharing a rank.}
\label{tab:motion-point-accuracy}
\fontsize{8}{10}\selectfont
\setlength{\tabcolsep}{2.7pt}\renewcommand{\arraystretch}{1.18}
\begin{tabular}{@{}l*{12}{r}@{}}
\toprule
 & \multicolumn{2}{c}{\shortstack{Panoptic\\Studio}} & \multicolumn{2}{c}{PointOdyssey} & \multicolumn{2}{c}{\shortstack{Dynamic\\Replica}} & \multicolumn{2}{c}{\shortstack{Aria Digital\\Twin}} & \multicolumn{2}{c}{SynthVerse} & \multicolumn{2}{c}{Mean} \\
\cmidrule(lr){2-3}\cmidrule(lr){4-5}\cmidrule(lr){6-7}\cmidrule(lr){8-9}\cmidrule(lr){10-11}\cmidrule(lr){12-13}
Method & APD $\uparrow$ & EPE $\downarrow$ & APD $\uparrow$ & EPE $\downarrow$ & APD $\uparrow$ & EPE $\downarrow$ & APD $\uparrow$ & EPE $\downarrow$ & APD $\uparrow$ & EPE $\downarrow$ & APD $\uparrow$ & EPE $\downarrow$ \\
\midrule
SpaTrackerV2 & 49.93 & 0.4639 & 53.45 & 0.5173 & 54.34 & 0.4368 & 69.39 & 0.2764 & 16.67 & 5.5187 & 48.76 & 1.4426 \\
St4RTrack & 70.80 & 0.2479 & 67.38 & 0.3140 & 74.24 & 0.2594 & 72.90 & 0.2958 & 19.24 & 3.8485 & 60.91 & 0.9931 \\
TraceAnything & 70.86 & 0.2780 & 39.37 & 1.0462 & 60.89 & 0.5642 & 75.69 & 0.2515 & 18.78 & 7.3385 & 53.12 & 1.8957 \\
Any4D & 74.72 & 0.2361 & 65.99 & 0.3783 & 76.72 & 0.3463 & 78.09 & 0.2378 & 21.67 & 5.2362 & 63.44 & 1.2869 \\
V-DPM & 78.95 & 0.1757 & 80.82 & 0.1857 & 75.29 & 0.2438 & 85.47 & 0.1725 & \textcolor{green!45!black}{28.01} & 2.7367 & 69.71 & 0.7029 \\
4RC & 69.08 & 0.2590 & 79.81 & 0.2353 & \textcolor{blue!75!black}{84.65} & \textcolor{blue!75!black}{0.1667} & 84.16 & 0.1770 & 24.04 & 3.2957 & 68.35 & 0.8267 \\
OpenD4RT & 78.63 & 0.1811 & 66.03 & 0.3398 & 76.78 & 0.2407 & 69.91 & 0.2966 & 20.86 & 4.7392 & 62.44 & 1.1595 \\
UniQuery4R & \textcolor{blue!75!black}{84.64} & \textcolor{red!75!black}{\textbf{0.1333}} & \textcolor{green!45!black}{83.74} & \textcolor{green!45!black}{0.1604} & 80.00 & 0.2078 & \textcolor{green!45!black}{87.17} & \textcolor{green!45!black}{0.1387} & 27.74 & \textcolor{green!45!black}{2.4105} & \textcolor{green!45!black}{72.66} & \textcolor{green!45!black}{0.6101} \\
ARROW & \textcolor{red!75!black}{\textbf{85.70}} & \textcolor{blue!75!black}{0.1343} & \textcolor{blue!75!black}{84.87} & \textcolor{blue!75!black}{0.1544} & \textcolor{red!75!black}{\textbf{87.91}} & \textcolor{red!75!black}{\textbf{0.1433}} & \textcolor{red!75!black}{\textbf{93.57}} & \textcolor{red!75!black}{\textbf{0.0823}} & \textcolor{blue!75!black}{32.73} & \textcolor{blue!75!black}{2.2402} & \textcolor{red!75!black}{\textbf{76.96}} & \textcolor{blue!75!black}{0.5509} \\
\midrule
\textbf{Kosmo-GFM} & \textcolor{green!45!black}{83.41} & \textcolor{green!45!black}{0.1411} & \textcolor{red!75!black}{\textbf{85.73}} & \textcolor{red!75!black}{\textbf{0.1474}} & \textcolor{green!45!black}{80.76} & \textcolor{green!45!black}{0.2017} & \textcolor{blue!75!black}{89.78} & \textcolor{blue!75!black}{0.1102} & \textcolor{red!75!black}{\textbf{42.49}} & \textcolor{red!75!black}{\textbf{1.1105}} & \textcolor{blue!75!black}{76.43} & \textcolor{red!75!black}{\textbf{0.3422}} \\
\bottomrule
\end{tabular}
\end{table}

%% file: tables/motion-flow-accuracy.tex
\begin{table}[!htbp]
\centering
\caption{Scene-flow accuracy (percent) and EPE (meters). All ten methods are shown on all five datasets. Mean is recomputed as the arithmetic mean of the five displayed dataset values. Best, second-best, and third-best values are \textcolor{red!75!black}{red}, \textcolor{blue!75!black}{blue}, and \textcolor{green!45!black}{green}, with ties sharing a rank.}
\label{tab:motion-flow-accuracy}
\fontsize{8}{10}\selectfont
\setlength{\tabcolsep}{2.7pt}\renewcommand{\arraystretch}{1.18}
\begin{tabular}{@{}l*{12}{r}@{}}
\toprule
 & \multicolumn{2}{c}{\shortstack{Panoptic\\Studio}} & \multicolumn{2}{c}{PointOdyssey} & \multicolumn{2}{c}{\shortstack{Dynamic\\Replica}} & \multicolumn{2}{c}{\shortstack{Aria Digital\\Twin}} & \multicolumn{2}{c}{SynthVerse} & \multicolumn{2}{c}{Mean} \\
\cmidrule(lr){2-3}\cmidrule(lr){4-5}\cmidrule(lr){6-7}\cmidrule(lr){8-9}\cmidrule(lr){10-11}\cmidrule(lr){12-13}
Method & Acc. $\uparrow$ & EPE $\downarrow$ & Acc. $\uparrow$ & EPE $\downarrow$ & Acc. $\uparrow$ & EPE $\downarrow$ & Acc. $\uparrow$ & EPE $\downarrow$ & Acc. $\uparrow$ & EPE $\downarrow$ & Acc. $\uparrow$ & EPE $\downarrow$ \\
\midrule
SpaTrackerV2 & 8.78 & 0.4095 & 15.08 & 0.3089 & 13.09 & 0.2707 & 34.40 & 0.2054 & 6.10 & 4.1792 & 15.49 & 1.0747 \\
St4RTrack & 27.45 & 0.2259 & 37.24 & 0.2190 & 75.69 & 0.0858 & 55.76 & 0.1481 & 10.77 & 3.3004 & 41.38 & 0.7958 \\
TraceAnything & 23.94 & 0.2456 & 64.33 & 0.1173 & 83.83 & 0.0495 & 79.98 & 0.0839 & 22.16 & 2.7136 & 54.85 & 0.6420 \\
Any4D & 25.08 & 0.2406 & 70.43 & 0.0940 & 85.94 & 0.0451 & 88.25 & 0.0562 & 27.93 & 2.1907 & 59.53 & 0.5253 \\
V-DPM & \textcolor{blue!75!black}{58.99} & \textcolor{green!45!black}{0.1152} & 82.91 & 0.0563 & 88.98 & 0.0381 & 88.75 & 0.0559 & 27.09 & 1.8237 & 69.34 & 0.4178 \\
4RC & 53.88 & 0.1255 & \textcolor{red!75!black}{\textbf{86.66}} & \textcolor{red!75!black}{\textbf{0.0466}} & \textcolor{red!75!black}{\textbf{94.13}} & \textcolor{red!75!black}{\textbf{0.0213}} & 88.14 & 0.0633 & \textcolor{blue!75!black}{31.59} & \textcolor{green!45!black}{1.5045} & 70.88 & 0.3522 \\
OpenD4RT & 38.25 & 0.1772 & 76.62 & 0.0784 & 90.45 & 0.0289 & 88.23 & 0.0627 & 28.36 & 2.6931 & 64.38 & 0.6081 \\
UniQuery4R & \textcolor{green!45!black}{56.75} & \textcolor{blue!75!black}{0.1137} & 82.52 & \textcolor{green!45!black}{0.0541} & \textcolor{green!45!black}{91.54} & \textcolor{blue!75!black}{0.0275} & \textcolor{green!45!black}{95.10} & \textcolor{blue!75!black}{0.0167} & \textcolor{green!45!black}{29.25} & 1.5411 & \textcolor{green!45!black}{71.03} & \textcolor{green!45!black}{0.3506} \\
ARROW & \textcolor{red!75!black}{\textbf{72.89}} & \textcolor{red!75!black}{\textbf{0.0814}} & \textcolor{blue!75!black}{84.85} & 0.0628 & \textcolor{blue!75!black}{93.37} & 0.0319 & \textcolor{blue!75!black}{96.01} & \textcolor{green!45!black}{0.0174} & 28.36 & \textcolor{blue!75!black}{1.2604} & \textcolor{red!75!black}{\textbf{75.10}} & \textcolor{blue!75!black}{0.2908} \\
\midrule
\textbf{Kosmo-GFM} & 50.15 & 0.1257 & \textcolor{green!45!black}{83.02} & \textcolor{blue!75!black}{0.0521} & 91.31 & \textcolor{green!45!black}{0.0279} & \textcolor{red!75!black}{\textbf{96.34}} & \textcolor{red!75!black}{\textbf{0.0134}} & \textcolor{red!75!black}{\textbf{34.56}} & \textcolor{red!75!black}{\textbf{0.7213}} & \textcolor{blue!75!black}{71.08} & \textcolor{red!75!black}{\textbf{0.1881}} \\
\bottomrule
\end{tabular}
\end{table}

%% file: sections/pose-scenes.tex
\paragraph{\textbf{Scene selection.}}
The common-scene evaluation uses the following scenes. Names follow the
dataset directories. The additional ETH3D scenes listed below were selected
initially but are excluded from the common-scene aggregates because VidMap
failed on their inconsistent image resolutions.

\begingroup
\small
\setlength{\tabcolsep}{5pt}
\renewcommand{\arraystretch}{1.18}
\noindent\begin{tabularx}{\textwidth}{@{}p{.19\textwidth}X@{}}
\toprule
Dataset & Scenes in the common-scene evaluation \\
\midrule
ETH3D (15) & botanical\_garden, boulders, bridge, door, exhibition\_hall,
kicker, lecture\_room, living\_room, lounge, meadow, office, old\_computer,
pipes, statue, terrace\_2. \\
\addlinespace[4pt]
MipNeRF360 (9) & bicycle, bonsai, counter, flowers, garden, kitchen, room,
stump, treehill. \\
\addlinespace[4pt]
Tanks \& Temples (6) & Barn, Caterpillar, Courthouse, Ignatius, Meetingroom, Truck. \\
\addlinespace[4pt]
SMERF (4) & alameda, berlin, london, nyc. These are the Zip-NeRF scenes
released through SMERF~\citep{barron2023zipnerf,duckworth2024smerf}. \\
\bottomrule
\end{tabularx}
\par\smallskip
\noindent\textbf{Additional selected ETH3D scenes (10), excluded from the
common-scene aggregates:} courtyard, delivery\_area, electro, facade,
observatory, playground, relief, relief\_2, terrace, terrains.
\par
\endgroup

%% file: tables/pose-summary.tex
\begin{table}[!htbp]
\centering
\caption{Reconstruction, camera calibration, and trajectory consistency on 34 completed scenes with 11,336 images. Metrics are averaged equally across scenes. AUC, R@1, and T@1 are percentages. ATE is Sim(3)-aligned camera-center nRMSE, normalized by the GT center RMS radius. Runtime is cumulative hours. Bold indicates the best value.}
\label{tab:pose}
\small
\setlength{\tabcolsep}{2.5pt}\renewcommand{\arraystretch}{1.12}
\begin{tabular}{@{}lrrrrrrrrr@{}}
\toprule
Method & AUC@1 $\uparrow$ & AUC@3 $\uparrow$ & AUC@5 $\uparrow$ & R@1 $\uparrow$ & T@1 $\uparrow$ & \shortstack{Focal\\AbsRel $\downarrow$} & \shortstack{ATE\\nRMSE $\downarrow$} & Registered & Time (h) $\downarrow$ \\
\midrule
COLMAP & 75.73 & 86.33 & 89.28 & 93.58 & 88.45 & 0.00389 & 0.05805 & 11,315 & 47.08 \\
GLUEMAP & 69.13 & 86.18 & 90.66 & 96.24 & 89.61 & 0.00230 & 0.03176 & 11,336 & 69.67 \\
VidMap & 57.23 & 71.10 & 74.29 & 78.23 & 76.61 & 0.00328 & 0.33359 & 9,017 & 20.32 \\
\midrule
PoseEngine & \textbf{78.45} & \textbf{91.28} & \textbf{94.26} & \textbf{98.74} & \textbf{96.00} & \textbf{0.00158} & \textbf{0.02640} & 11,336 & 14.07 \\
PoseEngine Flash & 62.50 & 83.64 & 89.28 & 95.95 & 89.87 & 0.00316 & 0.02823 & 11,336 & \textbf{2.62} \\
\bottomrule
\end{tabular}
\end{table}

%% file: tables/matching.tex
\begin{table}[!htbp]
\centering
\caption{MVLoMa versus LoMa Large on 48 scenes for each STAR policy. AUC@1 is a percentage, and $\Delta$ is MVLoMa minus LoMa Large: AUC@1 differences are percentage points, and ATE differences use the displayed means in GT coordinate units. ATE is raw Sim(3)-aligned camera-position RMSE in GT coordinate units, averaged equally over scenes within each dataset. Raw ATE is not aggregated across datasets; dashes mark omitted Overall ATE and $\Delta$ATE. Bold marks the better result in each method pair.}
\label{tab:matching}
\small
\setlength{\tabcolsep}{4pt}\renewcommand{\arraystretch}{1.12}
\begin{tabular}{lrrrrrrr}
\toprule
\multirow{2}{*}{Dataset} & \multirow{2}{*}{$N$} & \multicolumn{2}{c}{MVLoMa} & \multicolumn{2}{c}{LoMa Large} & \multicolumn{2}{c}{$\Delta$} \\
\cmidrule(lr){3-4}\cmidrule(lr){5-6}\cmidrule(lr){7-8}
 & & AUC@1 $\uparrow$ & ATE $\downarrow$ & AUC@1 $\uparrow$ & ATE $\downarrow$ & AUC@1 $\uparrow$ & ATE $\downarrow$ \\
\midrule
\multicolumn{8}{l}{\textbf{Unordered STAR}} \\
ETH3D & 25 & \textbf{80.29} & \textbf{0.00879} & 78.00 & 0.01996 & +2.30 & -0.01117 \\
MipNeRF360 & 9 & \textbf{78.81} & \textbf{0.00903} & 74.35 & 0.01433 & +4.46 & -0.00530 \\
Tanks \& Temples & 6 & \textbf{84.13} & \textbf{0.43660} & 81.41 & 0.43793 & +2.72 & -0.00133 \\
Oxford Spires & 4 & \textbf{11.39} & \textbf{2.57710} & 6.71 & 2.68618 & +4.69 & -0.10908 \\
SMERF & 4 & \textbf{88.09} & \textbf{0.00363} & 81.92 & 0.00809 & +6.17 & -0.00446 \\
\midrule
Overall & 48 & \textbf{75.40} & -- & 72.12 & -- & +3.28 & -- \\
\addlinespace[5pt]
\multicolumn{8}{l}{\textbf{Ordered STAR}} \\
ETH3D & 25 & \textbf{78.89} & \textbf{0.01094} & 75.82 & 0.01811 & +3.08 & -0.00717 \\
MipNeRF360 & 9 & \textbf{71.58} & \textbf{0.05099} & 66.33 & 0.05711 & +5.25 & -0.00612 \\
Tanks \& Temples & 6 & \textbf{85.25} & \textbf{0.43623} & 78.41 & 0.56239 & +6.84 & -0.12616 \\
Oxford Spires & 4 & \textbf{16.47} & \textbf{2.28675} & 9.37 & 3.30433 & +7.10 & -1.01758 \\
SMERF & 4 & \textbf{83.31} & \textbf{0.00595} & 76.50 & 0.01234 & +6.81 & -0.00639 \\
\midrule
Overall & 48 & \textbf{73.48} & -- & 68.88 & -- & +4.60 & -- \\
\addlinespace[5pt]
\bottomrule
\end{tabular}

\end{table}

%% file: tables/rendering.tex
\begingroup
\setlength{\intextsep}{4pt}
\begin{table}[H]
\captionsetup{skip=3pt}
\centering
\caption{3DGS rendering after camera refinement. All runs use LiteGS, 30,000 iterations and the same image split. Non-reference methods optimize training cameras and refine each test pose against its test image for 20 steps. The fixed-reference baseline uses neither training-pose optimization nor test-pose refinement. Dataset means weight scenes equally.}
\label{tab:rendering}
\small
\setlength{\tabcolsep}{4pt}\renewcommand{\arraystretch}{1.0}
\begin{tabular}{llrrrr}
\toprule
Dataset & Initialization & $N$ & PSNR (dB) $\uparrow$ & SSIM $\uparrow$ & LPIPS $\downarrow$ \\
\midrule
MipNeRF360 & Reference poses & 9 & 28.19 & 0.8224 & 0.2570 \\
 & GLUEMAP & 9 & 28.40 & 0.8306 & 0.2510 \\
 & PoseEngine & 9 & 26.69 & 0.7714 & 0.2739 \\
 & PoseEngine Flash & 9 & 26.34 & 0.7439 & 0.2802 \\
\midrule
Tanks \& Temples & Reference poses & 6 & 25.90 & 0.8472 & 0.2543 \\
 & GLUEMAP & 6 & 25.80 & 0.8486 & 0.2525 \\
 & PoseEngine & 6 & 25.53 & 0.8392 & 0.2605 \\
 & PoseEngine Flash & 6 & 24.18 & 0.7923 & 0.2774 \\
\bottomrule
\end{tabular}
\par\vspace{3pt}
{\footnotesize Higher PSNR/SSIM and lower VGG LPIPS are better. This is image-conditioned pose-refined evaluation, not a fixed-pose held-out test.}
\end{table}
\endgroup

%% file: tables/scaling-encoder.tex
\begin{table}[H]
\centering
\caption{Encoder initialization for the scaling model. Base-size reconstruction models are initialized with the Base variant of each pretrained encoder and trained for 100k iterations on Hypersim with 2--24 input views. The geometry-pretrained Depth Anything 3-Base encoder is listed as a reference. Bold marks the best value among the three general-purpose encoders.}
\label{tab:scaling-encoder}
\small
\setlength{\tabcolsep}{4pt}\renewcommand{\arraystretch}{1.12}
\begin{tabular}{lrrrr}
\toprule
Encoder initialization & Local AbsRel $\downarrow$ & Global AbsRel $\downarrow$ & AUC@30 $\uparrow$ & AUC@5 $\uparrow$ \\
\midrule
Depth Anything 3-Base (reference) & 0.0793 & 0.4819 & 0.9606 & 0.7742 \\
\midrule
DINOv2-Base & 0.1333 & 1.0499 & 0.7823 & 0.3527 \\
DINOv3-Base & 0.1218 & 0.5506 & 0.8241 & 0.3883 \\
LingBot-Vision-Base & \textbf{0.1175} & \textbf{0.5284} & \textbf{0.8441} & \textbf{0.4064} \\
\bottomrule
\end{tabular}
\end{table}

%% file: tables/scaling-data-values.tex
\begin{table}[!htbp]
\centering
\caption{Data-scaling observations for the 0.2B model at 200k steps. AUC values are fractions, and dashes denote an unreported setting.}
\label{tab:scaling-data}
\fontsize{8}{10}\selectfont
\setlength{\tabcolsep}{4pt}\renewcommand{\arraystretch}{1.0}
\begin{tabular}{l*{9}{r}}
\toprule
\multirow{2}{*}{Trajectories} & \multicolumn{3}{c}{8 GPUs} & \multicolumn{3}{c}{16 GPUs} & \multicolumn{3}{c}{32 GPUs} \\
\cmidrule(lr){2-4}\cmidrule(lr){5-7}\cmidrule(lr){8-10}
 & AbsRel $\downarrow$ & AUC@30 $\uparrow$ & AUC@5 $\uparrow$ & AbsRel $\downarrow$ & AUC@30 $\uparrow$ & AUC@5 $\uparrow$ & AbsRel $\downarrow$ & AUC@30 $\uparrow$ & AUC@5 $\uparrow$ \\
\midrule
1,879 & 0.65917 & 0.12169 & 0.01456 & 0.66276 & 0.12468 & 0.01498 & 0.71213 & 0.11167 & 0.01621 \\
9,394 & 0.47453 & 0.28411 & 0.05513 & 0.48733 & 0.28863 & 0.05477 & -- & -- & -- \\
18,788 & 0.41348 & 0.32196 & 0.06557 & 0.41262 & 0.34058 & 0.06877 & 0.39743 & 0.34915 & 0.07881 \\
93,940 & 0.32707 & 0.38433 & 0.08666 & 0.32710 & 0.42081 & 0.10139 & 0.31562 & 0.42707 & 0.11179 \\
187,880 & 0.30391 & 0.39172 & 0.10103 & 0.28976 & 0.43517 & 0.13044 & 0.27833 & 0.44961 & 0.13702 \\
\bottomrule
\end{tabular}
\end{table}

%% file: sections/reconstruction.tex
\clearpage
\section{Qualitative Results}
\label{sec:applications}
This section illustrates the range of reconstruction outputs supported by Kosmo-GFM on public datasets and captured sequences: camera trajectories, sparse and dense geometry, feed-forward 3DGS, 3DGS, dynamic points, and 4DGS. Camera-conditioned prediction incorporates recovered camera information into dense geometry and renderable scene representations, while the dynamic examples illustrate scene structure and motion over time. These qualitative results show the workflow from camera recovery to geometry reconstruction and rendering, demonstrate support for multiple reconstruction tasks and output representations, and complement the preceding quantitative evaluations.

\subsection{Camera Pose and Sparse Geometry}
Figure~\ref{fig:recon-pose-1} compares Kosmo-GFM and COLMAP camera trajectories and sparse reconstructions on LingBot-Map demo sequences~\citep{lingbotmap2026demo} and Oxford Spires data~\citep{tao2024oxfordspires}. Oxford Spires trajectories are aligned to ground-truth poses.
\begingroup\setlength{\intextsep}{6pt}
\input{sections/recon-camera_pose}
\endgroup

\clearpage
\subsection{Camera-Conditioned Dense Points}
Figure~\ref{fig:recon-static-example-1} presents dense point clouds from LingBot-Map demo sequences~\citep{lingbotmap2026demo}, Oxford Spires, SMERF, Tanks \& Temples, MipNeRF360, and ETH3D. GFM predicts dense points conditioned on PoseEngine cameras and accumulates them across the input frames.
\input{sections/recon-static_points}

\clearpage
\subsection{Feed-Forward 3DGS}
We extend camera-conditioned GFM with a query-based 3DGS decoder that directly predicts a renderable scene representation. Figure~\ref{fig:recon-ffgs-1} shows results on Hypersim~\citep{roberts2021hypersim} alongside InfiniSplat v2~\citep{hu2026infinisplat} with ground-truth poses.
\input{sections/recon-feed_forward_gs}

\clearpage
\subsection{3DGS}
Figure~\ref{fig:recon-3dgs-insta360-1} shows 3DGS reconstructions from Insta360 captures, rendered from multiple viewpoints.
\input{sections/recon-gaussian_3d}

\clearpage
\begingroup\setlength{\intextsep}{6pt}
\subsection{Dynamic Points}
Figure~\ref{fig:recon-dynamic-1} presents dynamic-point reconstructions on DAVIS videos~\citep{perazzi2016davis,ponttuset2017davis} and studio captures, showing scene geometry and motion over time.
\input{sections/recon-dynamic_points}

\subsection{4DGS}
Figure~\ref{fig:recon-4dgs-1} shows reconstructions from our in-house 4DGS implementation on Volumetric Video Challenge (VVC 2025) sequences~\citep{vvc2025challenge}.
\input{sections/recon-gaussian_4d}
\endgroup
\clearpage

%% file: sections/recon-camera_pose.tex
\begin{figure}[H]
\centering
\makebox[\textwidth][c]{\includegraphics[width=1.06\textwidth,height=.80\textheight,keepaspectratio]{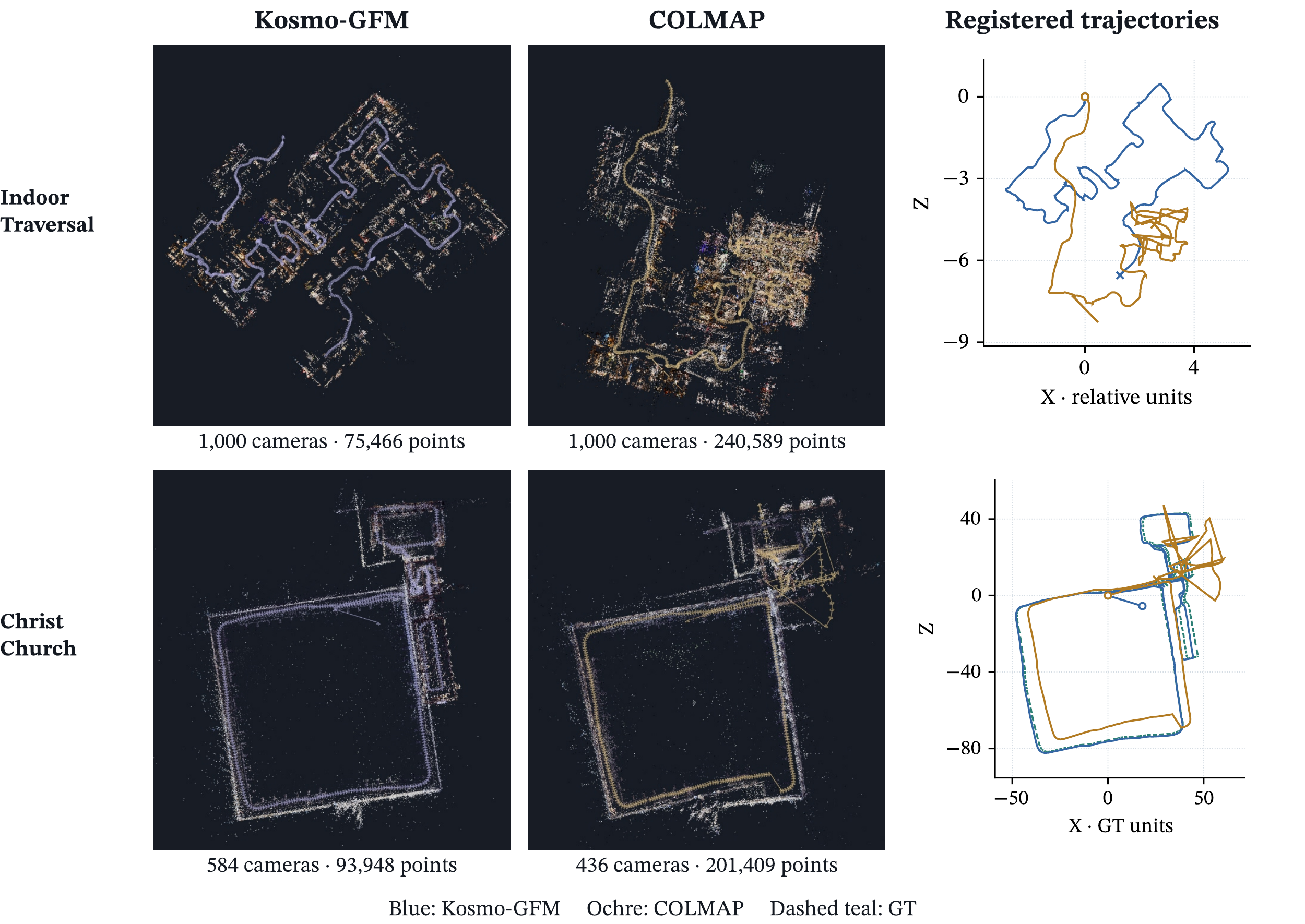}}
\caption{Camera pose and sparse geometry. Left and center: sparse points of each reconstruction, framed independently to show the main structure. Right: complete registered trajectories with circles/crosses at the start/end, retaining the recorded alignment in a common coordinate system.}
\label{fig:recon-pose-1}
\end{figure}

%% file: sections/recon-static_points.tex
\begin{figure}[H]
\centering
\includegraphics[width=\textwidth,height=\dimexpr\textheight-125pt\relax,keepaspectratio]{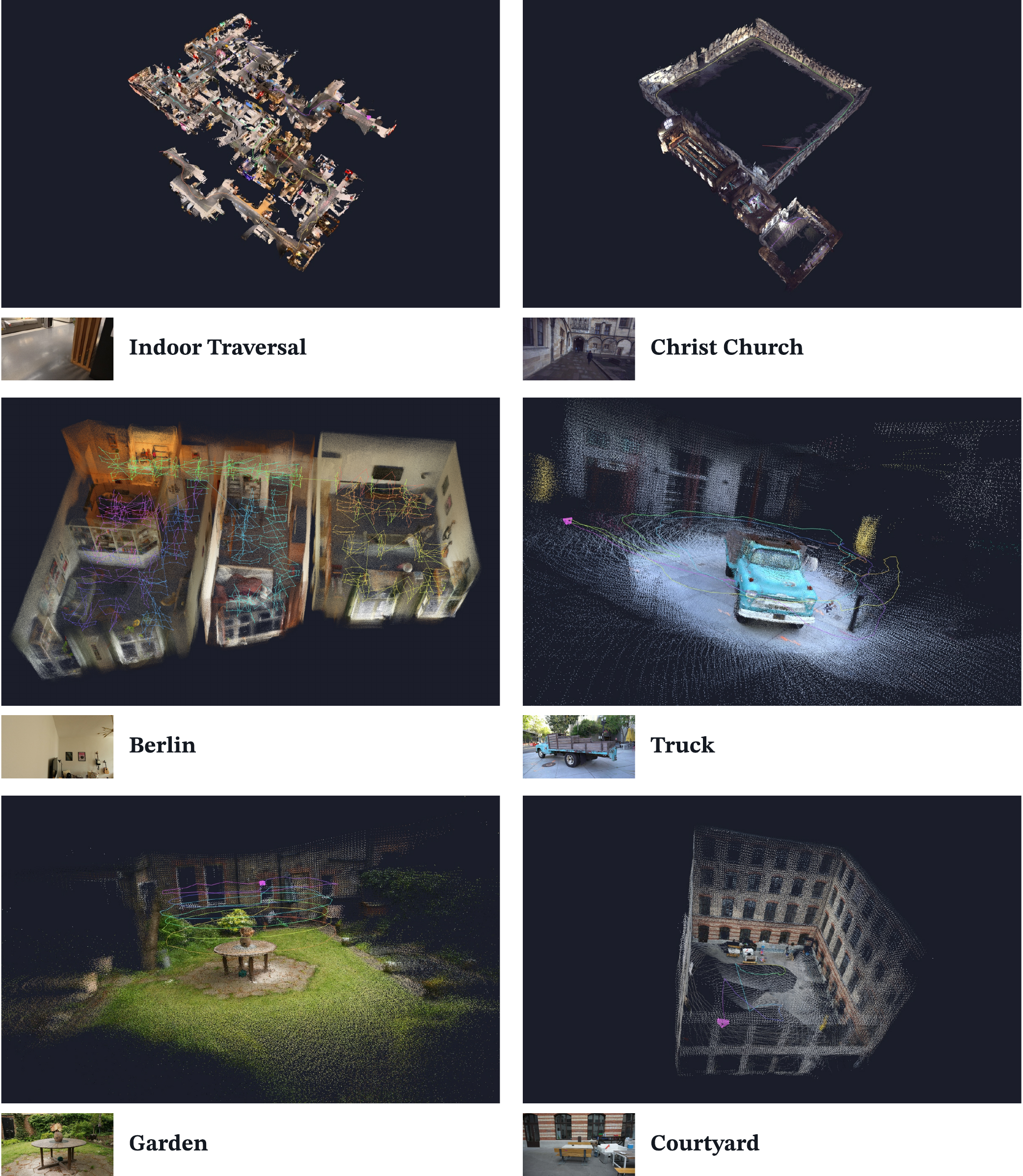}
\caption{Camera-conditioned dense points. Each scene shows a dense-point reconstruction above a small reference input image.  We compute the confidence threshold separately for each chunk as the 20th percentile of its point confidence values. For each selected scene, we accumulate point predictions from all available frames.}
\label{fig:recon-static-example-1}
\end{figure}

%% file: sections/recon-feed_forward_gs.tex
\begin{figure}[H]
\centering
\includegraphics[width=\textwidth,height=.80\textheight,keepaspectratio]{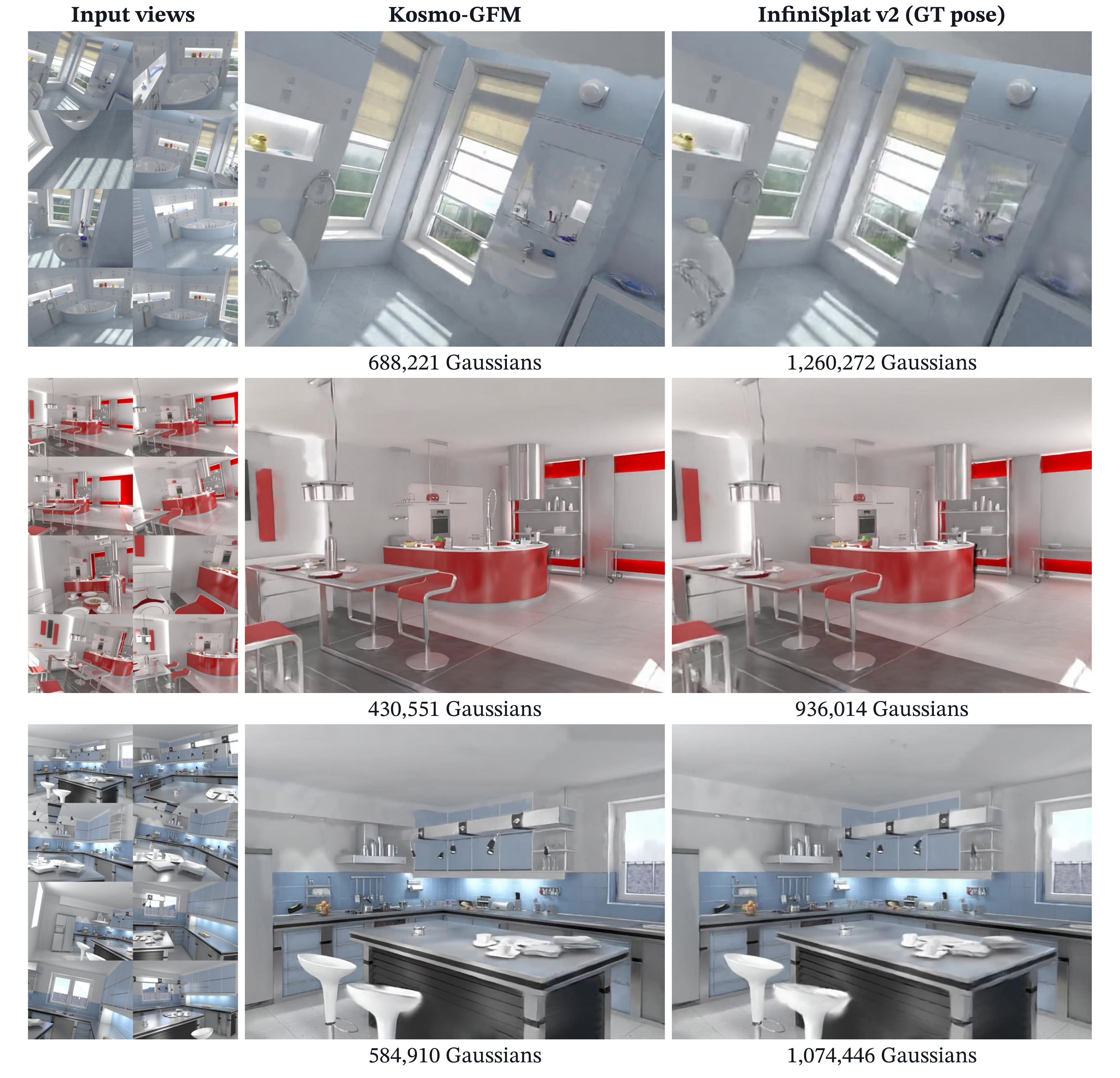}
\caption{Feed-Forward 3DGS. Columns show eight input views and rendered views for Kosmo-GFM and InfiniSplat v2 with GT poses. The rendering viewpoints may differ between methods, so these are not matched-pixel quantitative comparisons. Counts describe representation size, not reconstruction accuracy.}
\label{fig:recon-ffgs-1}
\end{figure}

%% file: sections/recon-gaussian_3d.tex
\begin{figure}[H]
\centering
\includegraphics[width=\textwidth,height=\dimexpr\textheight-75pt\relax,keepaspectratio]{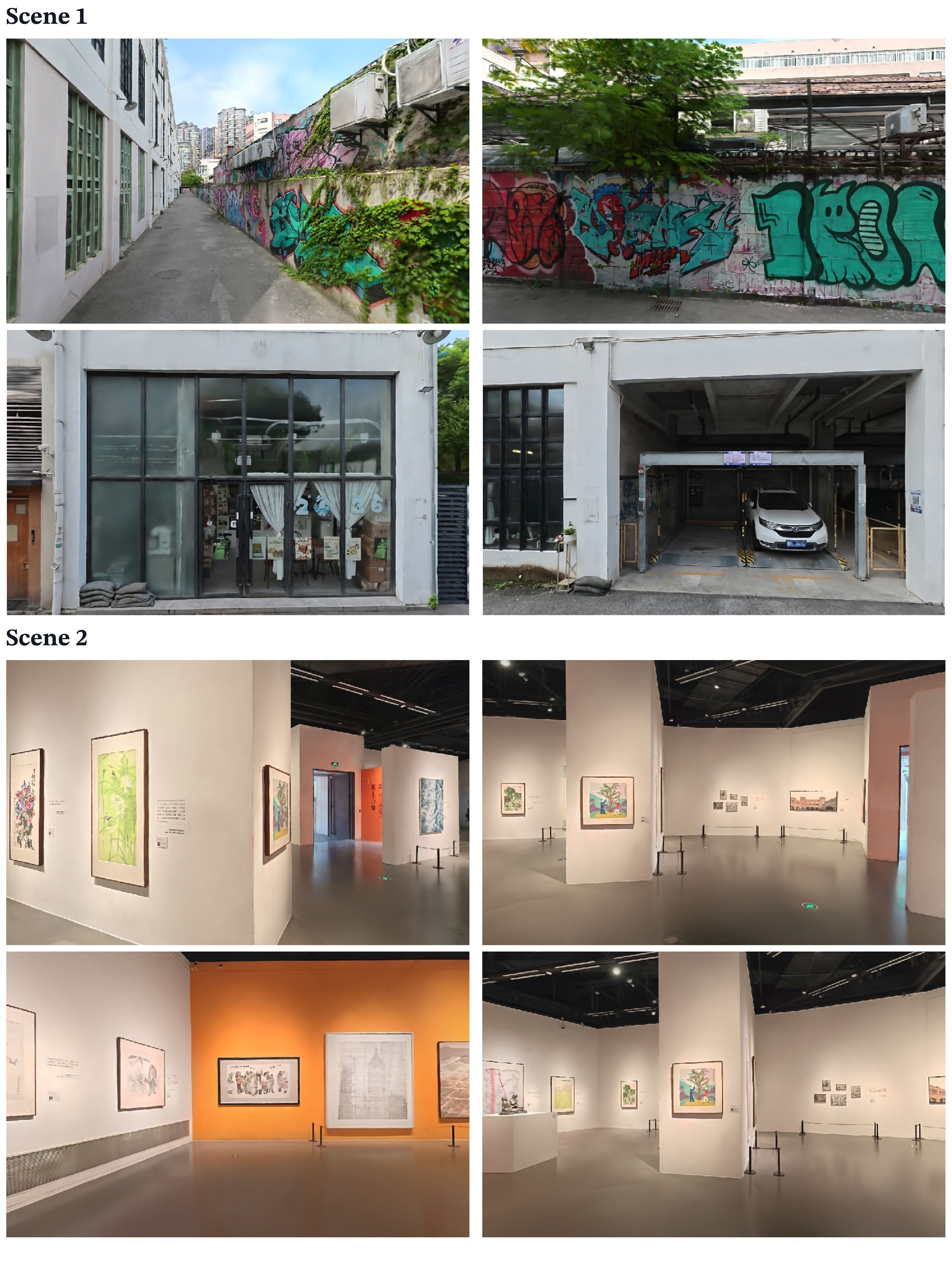}
\caption{3DGS. Four views along the recorded camera trajectory are shown for each of Scene 1 and Scene 2. All panels are rendered outputs of the reconstructed scenes.}
\label{fig:recon-3dgs-insta360-1}
\end{figure}

%% file: sections/recon-dynamic_points.tex
\begin{figure}[H]
\centering
\includegraphics[width=\textwidth,height=.80\textheight,keepaspectratio]{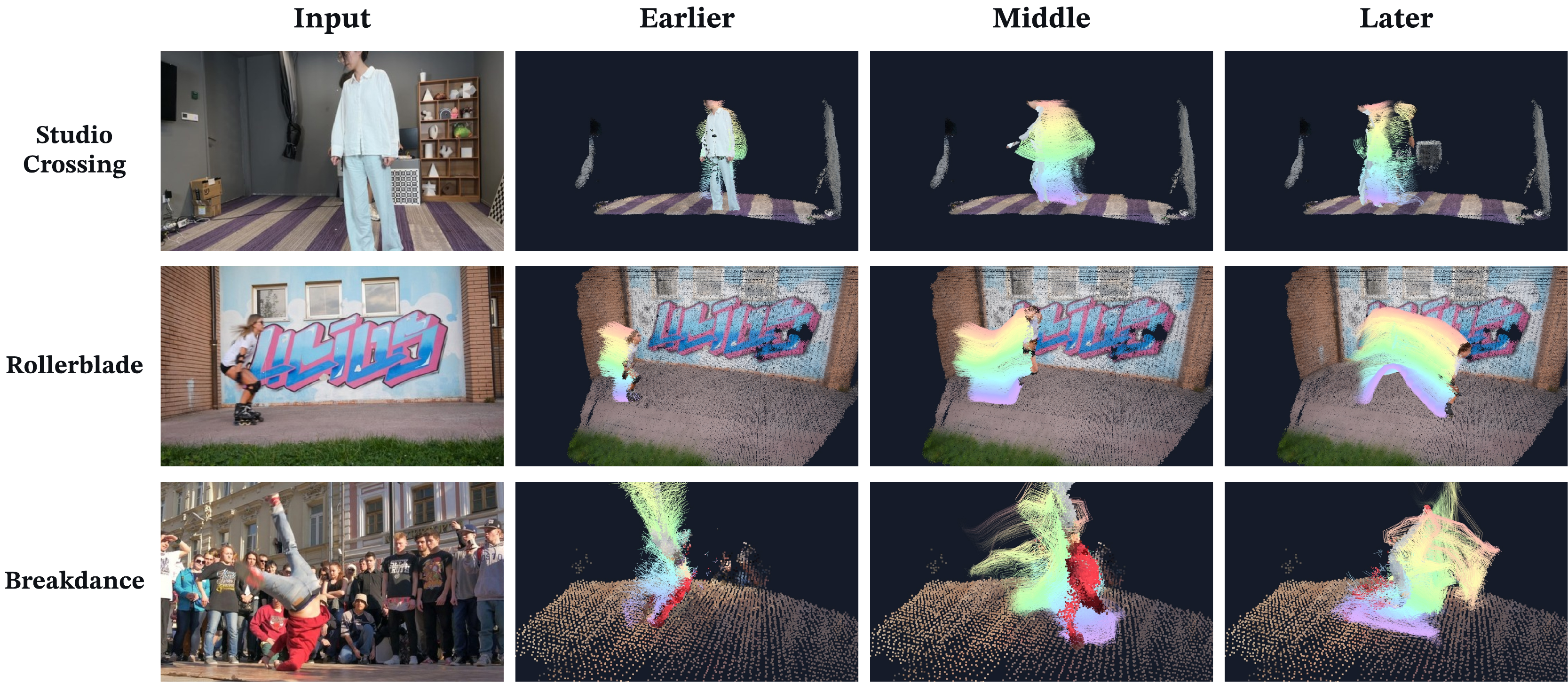}
\caption{Dynamic-point sequences. Each row shows an input frame and reconstructions at three time points from a fixed viewpoint. Backgrounds, masks, and motion trails use consistent display settings.}
\label{fig:recon-dynamic-1}
\end{figure}

%% file: sections/recon-gaussian_4d.tex
\begin{figure}[H]
\centering
\includegraphics[width=\textwidth,height=.80\textheight,keepaspectratio]{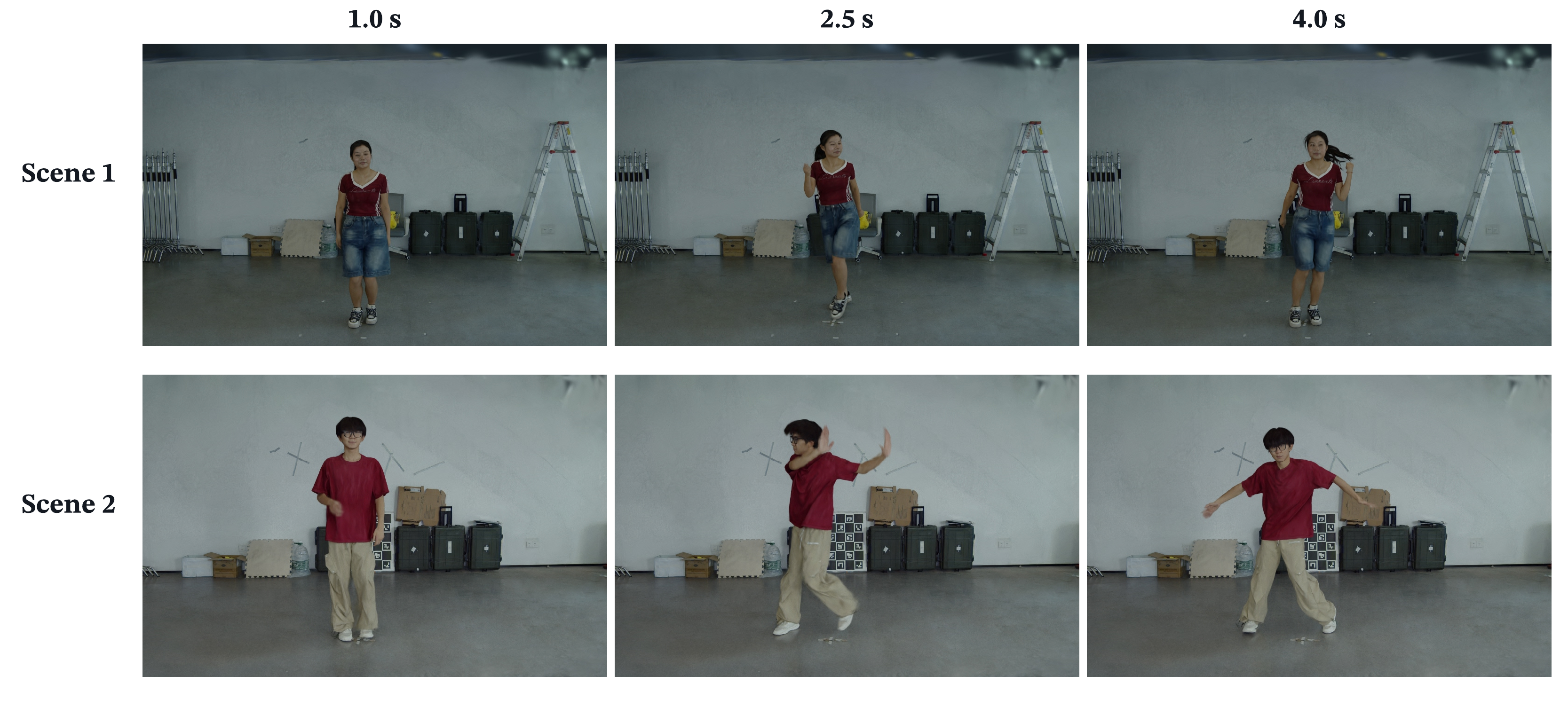}
\caption{4DGS on VVC 2025 sequences. Top: Scene 1. Bottom: Scene 2. Columns show 1.0, 2.5, and 4.0 seconds from a fixed viewpoint for each sequence.}
\label{fig:recon-4dgs-1}
\end{figure}

%% file: sections/limitation-examples.tex
\begin{figure}[H]
  \centering
  \includegraphics[width=.92\textwidth]{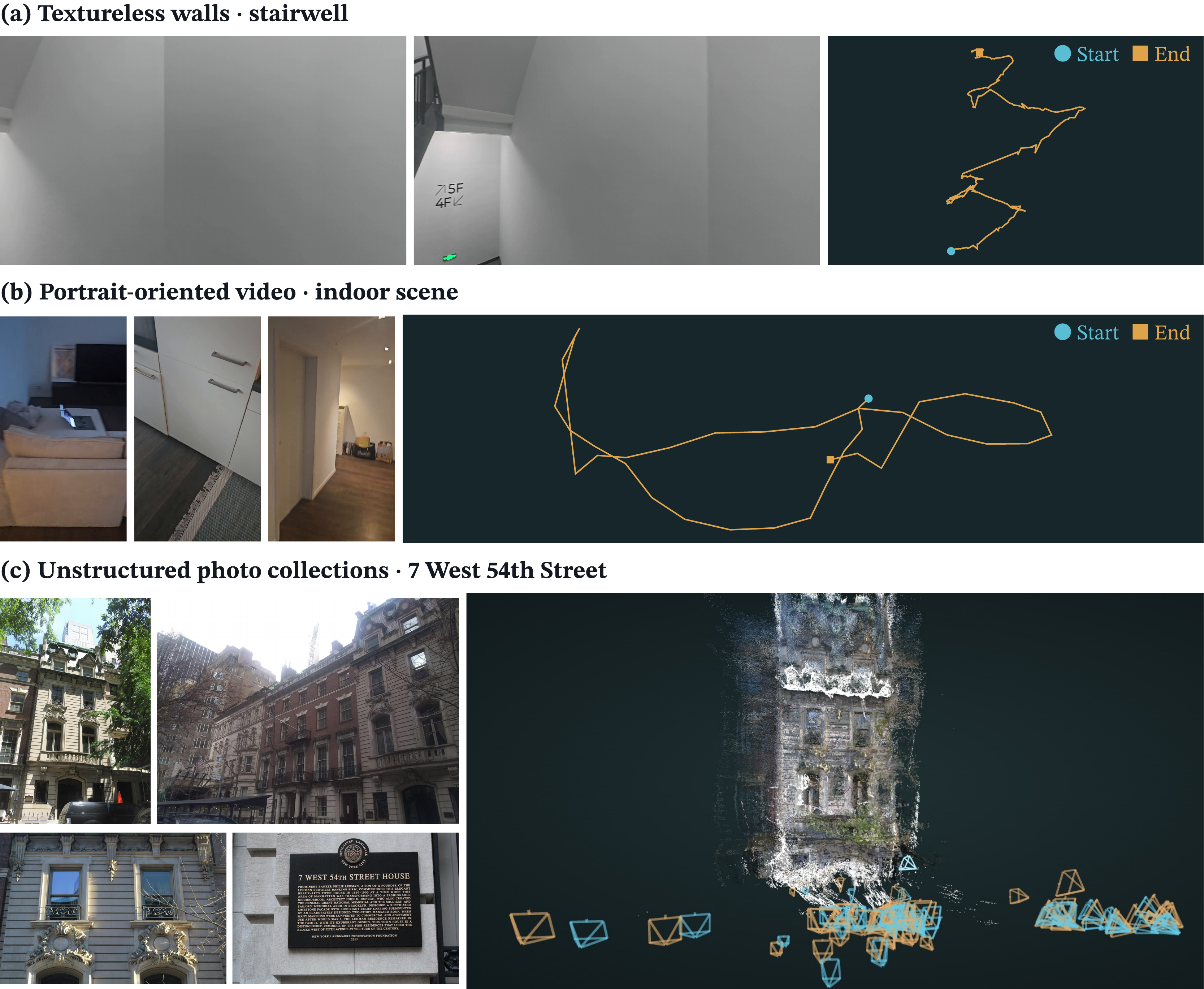}
  \caption{Qualitative examples. \textbf{(a)} Stairwell input frames and the complete camera trajectory, displayed vertically. \textbf{(b)} Portrait-oriented video frames and the complete camera trajectory. \textbf{(c)} Four input photos of 7 West 54th Street and a point cloud overlaid with GFM cameras (orange) and GT cameras (blue).}
  \label{fig:limitation-examples}
\end{figure}